\documentclass[times, 3p]{elsarticle}
\usepackage{lineno,hyperref}
\usepackage{float}
\usepackage{booktabs}
\usepackage{amsmath}
\usepackage{bm}
\usepackage{dsfont}
\usepackage{mathrsfs}
\usepackage{subcaption}
\usepackage{longtable}
\usepackage{amsfonts}
\usepackage{amssymb}
\usepackage{float}
\usepackage{graphicx}
\usepackage{adjustbox}
\usepackage{epstopdf}
\usepackage{dsfont}
\usepackage{makecell}
\usepackage{multirow}
\usepackage{adjustbox}

\usepackage[flushleft]{threeparttable}

\usepackage[ruled,vlined]{algorithm2e}

\newdefinition{defn}{Definition}
\newdefinition{asp}{Assumption}
\newdefinition{rmk}{Remark}
\newproof{proof}{Proof}

\journal{Elsevier}

\usepackage{natbib}

\hypersetup{
    colorlinks,
    linkcolor={black},
    citecolor={black},
    urlcolor={black}
}
\biboptions{authoryear}

\begin{document}
\nolinenumbers
\begin{frontmatter}

\title{Scalable Dynamic Origin-Destination Demand Estimation Enhanced by High-Resolution Satellite Imagery Data}

\author[1]{Jiachao Liu}

\author[2]{Pablo Guarda}

\author[2]{Koichiro Niinuma}

\author[1,3]{Sean Qian}
\ead{seanqian@cmu.edu}

\address[1]{Department of Civil and Environmental Engineering, Carnegie Mellon University, Pittsburgh, PA}
\address[2]{Fujitsu Research of America, Pittsburgh, PA}
\address[3]{Heinz College of Information Systems and Public Policy, Carnegie Mellon University, Pittsburgh, PA}

\begin{abstract}
This study presents a novel integrated framework for dynamic origin-destination demand estimation (DODE) in multi-class mesoscopic network models, incorporating high-resolution satellite imagery together with conventional traffic data from local sensors. Unlike sparse local detectors, satellite imagery offers consistent, city-wide road and traffic information of both parking and moving vehicles, overcoming data availability limitations. To extract information from imagery data, we design a computer vision pipeline for class-specific vehicle detection and map matching, generating link-level traffic density observations by vehicle class. Building upon this information, we formulate a computational graph-based DODE framework that calibrates dynamic network states by jointly matching observed traffic counts/speeds from local sensors with density measurements derived from satellite imagery. To assess the accuracy and robustness of the proposed framework, we conduct a series of numerical experiments using both synthetic and real-world data.  {The results demonstrate that supplementing traditional data with satellite-derived density significantly improves estimation performance, especially for links without local sensors. Real-world experiments also show the framework’s potential for practical deployment on large-scale networks.} Sensitivity analysis further evaluates the impact of data quality related to satellite imagery data.
\end{abstract}


\end{frontmatter}


\section{Introduction}
The widespread availability of spatio-temporal data has created new opportunities for advancing computational tools to model network flows, individual traveler behavior, and travel demand in dynamic transportation networks. Recent developments in sensing technologies and artificial intelligence are revolutionizing traditional models, making them more data-driven, scalable, and effective for complex, large-scale applications. Dynamic Origin-Destination Demand Estimation (DODE) is a foundational prerequisite for dynamic network models to accurately reproduce the current spatio-temporal network conditions, supporting traffic assignment \citep{pi2019general} and control strategies \citep{ye2019gating, liu2023optimal, wu2024participatory, wu2025multiday}. {Traditional DODE studies embed physics-based traffic assignment models in a bi-level model structure, considering both traffic flow propagation and travel behavioral dynamics, while emerging studies formulate the problem using purely data-driven models without traffic assignment constraints.} Recent studies have merged the advantages of both approaches, reformulating physics-based DODE within learning-based frameworks and demonstrating promise in large-scale applications \citep{wu2018hierarchical, liu2023end, liu2025end, zhou2025flow}. Despite extensive research over the past few decades, DODE remains challenging because it is inherently a highly data-demanding problem due to its high-dimensional and under-determined nature. Accurate and reliable DODE, particularly on large-scale networks, requires the integration of diverse and comprehensive data sources. 

Existing DODE studies rely on either single or multiple data sources from fixed traffic detectors (e.g., traffic counts), mobile sensors (e.g., probe vehicle traces, speeds) and activity-based data (e.g., surveys, mobile device traces). However, one major limitation shared by these data sources is that they capture only a subset of travelers within the system due to the sparsity of the geographic coverage of sensors, low penetration rates of mobile sensors, low sample rates of surveys and privacy issues in data collection. Prior studies have demonstrated that the performance of DODE is highly sensitive to sensor placement \citep{zhu2016optimal, zhu2017integrating} while optimal sensor deployment requires long-term planning for new installations and routine maintenance for deployed sensors, both of which can be costly and time-consuming \citep{zhou2010information,salari2021modeling,li2023submodularity,hu2024sensor}. 

In recent years, satellite remote sensing data have emerged as promising new data sources for city-wide transportation models, overcoming the weaknesses of localized and mobile ground-based sensors \citep{sheehan2023city, gagliardi2023satellite}. They have demonstrated effectiveness in monitoring both transportation infrastructure conditions and network-wide traffic dynamics. These data often have larger geographical coverage, and provide detailed, consistent information of all individual vehicles detected and road features in city-wide networks. Advances in computer vision have significantly enhanced the information extraction process from these satellite images, including vehicle detection \citep{leitloff2010vehicle,chen2014vehicle,cao2016vehicle} and road segmentation \citep{dai2020road,mei2021coanet, kawamura2025rn}. 
Fusing this emerging data source, which is unconstrained by spatial locations, into DODE has great potential to make it more scalable, accurate, and economic. 

This study investigates the potential of incorporating high-resolution imagery data collected by flexible sky-view sensing techniques such as satellites into a computation graph-based DODE framework to enhance model scalability and performance in replicating real-world network-wide traffic conditions.  {The main contributions are summarized as follows:
\begin{itemize}
\item We introduce high-resolution satellite imagery as a new observation source for the DODE problem. These images provide class-specific network-wide traffic information, which is not restricted to detector locations, therefore complementing the sparse coverage of localized sensors and providing a scalable way to observe traffic states, especially on links without detectors. They also serve as an additional regularization for DODE to address underdetermined issues when historical OD matrices are not available.
\item We integrate a CV pipeline into our DODE framework for density observation extraction. This CV module is compatible with the computational graph-based DODE framework and can be further enhanced to improve detection accuracy. We explicitly model parking and moving vehicles in an agent-based mesoscopic dynamic network loading (DNL) model and use separate dynamic assignment ratio (DAR) matrices for inflow and outflow of curbside parking and moving vehicles to derive gradients for densities. 
\item We test our model on toy and large-scale networks using both synthetic and real world data. Our experiments reveal the benefit of incorporating satellite-derived density data into DODE and evaluate the sensitivity to sensing errors and sampling frequency. The results show the potential of using this emerging remote sensing data in large-scale applications.
\end{itemize}}

The organization of the remaining paper is as follows: Section~\ref{sec:related} summarizes related studies in DODE and the usage of satellite imagery in transportation research. Section~\ref{sec:model} introduces the proposed DODE framework integrating multi-source data including satellite imagery information and the general DNL model for density estimation. Section~\ref{sec:experiment} conducts numerical experiments using both real-world multi-source data and synthetic data. Sensitivity analysis is also performed to test the model's robustness to the data quality of density information extracted from satellite imagery. Section~\ref{sec:conclusion} summarizes the findings and provides future research directions.

\section{Related Work}
\label{sec:related}
\subsection{Dynamic OD Demand Estimation}
 {Traditional dynamic origin-destination demand estimation (DODE) has been formulated as an inverse traffic assignment problem, where time-varying OD demand is inferred as inputs to an equilibrium or simulation model to reproduce observed traffic states. We categorize existing DODE studies as model-based, learning-based, and hybrid approaches in this section. Model-based formulations embed a dynamic traffic assignment (DTA) model within a bi-level optimization problem, where the upper level minimizes the discrepancy between observed and simulated traffic states while the lower level propagates traffic flows through a physics-based network flow model subject to behavioral constraints. Learning-based approaches do not rely on explicit traffic assignment models and infer OD flows from observations using data-driven architectures. Hybrid approaches combine both perspectives by embedding learnable components (e.g. neural networks) within DODE pipelines while retaining behavioral and physical consistency.}

 {Recent studies have proposed computational graph-based formulations, enabling end-to-end differentiation and heterogeneous data fusion within DODE pipelines. In this line of work, network loading operators, travel behavior models, and other learnable modules can be encoded as a layered computational graph, enabling efficient gradient-based calibration, which combines model-based and learning-based models within a unified framework. \cite{wu2018hierarchical} proposed a multi-layered computational graph representation to structurally model travel demand variables and individual behavior parameters. \cite{Ma2020} and \cite{liu2024modeling} formulated DODE problems on a computational graph with a mesoscopic simulation utilizing multi-source system-level data. \cite{lu2023physics} proposed a physics-informed framework for joint traffic state and queue profile estimation on layered computational graph with heterogeneous data, and \cite{kim2024computational} further integrated transport demand and supply models in the framework. \cite{guarda_estimating_2024, guarda_traffic_2024} developed a computational graph framework to estimate network flow and travel behavior using day-to-day data and further used this framework to estimate traffic conditions for unobserved network locations. \cite{liu2023end, liu2025end} embedded neural networks in implicit layers of an end-to-end learning framework to directly learn travel choice preferences, supply and demand components, and equilibrium state from multi-day observations. \cite{du2025modeling} developed an end-to-end simulation-based optimization framework to calibrate metro passenger route choice. \cite{zhou2025flow} proposed a unified tensor-based computational graph for advanced traffic assignment across multiple modeling layers. These studies demonstrate the potential of this computational-graph framework to improve demand estimation while maintaining consistency with physical and behavioral models and leveraging diverse data sources.}

 {Regardless of the modeling approaches, DODE problems are highly data-demanding. Conventional traffic data used in DODE is collected using either fixed detectors or mobile detectors. Fixed detectors have fixed locations and provide continuous localized traffic conditions at these locations. Some examples are traffic counts collected by loop detectors, automated vehicle identification (AVI) data collected through communication between vehicle tags and barrier devices, and license plate recognition (LPR) data captured by monitoring cameras. Mobile detectors are portable devices attached to a subset of travelers in the system and collect successive traffic conditions based on location changes of these travelers. Some examples are link speed data collected by probe vehicles, vehicle traces of probe vehicles, and human activity trajectories collected by mobile phone location tracking. Table~\ref{tab:dode} summarizes some representative studies utilizing various data sources within each modeling category.}

\begin{table}
\small
    \caption{Representative studies}
    \label{tab:dode}
    \centering
    \begin{tabular}{lll}
    \hline
        representative literature & data type & model type \\
    \hline
    \cite{dixon2002real} & automated vehicle identification $\&$ vehicle traces & model-free\\
    \cite{zhou2006dynamic}  & automated vehicle identification data & model-based\\
    \cite{Frederix2010_density} & traffic flow, link density & model-based\\
    \cite{toledo2012estimation} & traffic count & model-based \\
    \cite{wang2013estimating} & cell phone location data & model-based\\
    \cite{lu2013dynamic} & traffic count & model-based \\
    \cite{lu2015kalman}  & traffic count & model-based\\
    \cite{CARRESE201783} & traffic count, vehicle trajectories & model-based \\
    \cite{ou2019learn} & traffic count & model-free\\
    \cite{mo2020_LPR} & license plate recognition data & model-based\\
    \cite{krishnakumari2020data} & traffic count & model-free\\
    \cite{Ma2020} & traffic count $\&$ speed & model-based\\
    \cite{tang2021dynamic}  & automated vehicle identification data & model-free\\
    \cite{cao2021day} & automated vehicle identification $\&$ vehicle traces & model-free\\
    \cite{ros2022practical} & traffic count, link travel time & model-free\\
    \cite{kumarage2023hybrid} & traffic count & model-based\\ 
    \cite{huo2023simulation} & traffic count & model-based \\
    \cite{sun2024stochastic} & traffic count & model-based \\
    \cite{lu2024two} & location based social network data & model-based\\
    \cite{Cao02092024} & vehicle trajectories & model-based\\   
    \hline
    \end{tabular}  
\end{table}

 {One widely acknowledged challenge of DODE is that it is intrinsically an underdetermined problem. Even if links are fully observed, multiple OD patterns may generate the same network conditions, and the relative error of an estimated OD matrix can be unbounded \citep{yang1991analysis}. To tackle this challenge, existing literature follows two complementary lines: (1) imposing structural assumptions on the OD matrix and parametric models and (2) leveraging richer and more diverse data sources to add independent constraints.}

 {One representative approach in the first line is low rank based dimension reduction, where temporal or spatial OD patterns are represented using a small number of latent factors. Principal Component Analysis (PCA) \citep{djukic2012application, djukic2012efficient, prakash2017reducing, prakash2018improving, krishnakumari2020data, qurashi2022dynamic} and compressed sensing techniques \citep{sanandaji2016compressive, wang_transportation_2022} project high-dimensional OD matrices onto a low-dimensional basis, which can be estimated even when historical OD matrices are sparse or partially unavailable. Therefore, low rank-based models can alleviate underdetermination and scalability issues without requiring new data types, by exploiting low rank structure in historical OD patterns. Another category of quasi-dynamic models specifies parametric evolution over time, reduce the number of independent decision variables, and improve scalability on large-scale networks by adding the structural constraints in parametric models \citep{cascetta2013quasi}. In addition, utility-based OD estimation and choice-model-driven formulations impose behavioral structure on OD demand \citep{cantelmo2018utility}. By linking OD flows to behavioral models and other demand-side parameters, they effectively shrink the feasible OD space through assumptions on preferences and responses to various attributes.}

 {The second line of work retains a high-dimensional OD representation and addresses underdetermined nature by incorporating rich, independent observations as regularization. Prior studies have incorporated traffic counts and travel times \citep{Ma2020}, probe trajectories \citep{dixon2002real, cao2021day, CARRESE201783}, and other mobility datasets \citep{liu2024modeling, wang2013estimating, lu2024two} to better constrain optimization problem and estimate network flow patterns. These heterogeneous data sources are complementary but each has distinct characteristics and limitations. Fixed detectors can be expensive and spatially sparse, and sensitive to sensor placement \citep{hu2024sensor}. Additionally, maintaining installed sensors and planning new locations can be costly, especially in large-scale networks, limiting scalability and sustainability. Mobile detectors are more flexible but rely on penetration rates and may suffer from biased sampling, incomplete coverage, and privacy issues \citep{he2024efficient}. Moreover, both sensing categories can exhibit inconsistent data quality across time and space, potentially leading to contradictory measurements for the same link. To overcome these limitations, researchers have explored alternative observations that capture traffic states more directly. For example, \cite{Frederix2010_density} combined additional link-level density information into DODE, outperforming traditional estimation approaches using only traffic flow because density observations are effective in distinguishing congested and uncongested traffic states. In this context, emerging remote sensing (e.g., satellite imagery) provides network-wide snapshots of traffic states (density) with broad spatial coverage but limited temporal frequency, making it a promising complementary data source for DODE. This study follows this line of literature and explores the potential of incorporating satellite image-derived densities into DODE as an additional data source.}

\subsection{Satellite imagery in transportation research}
Advances in computer vision and sensing have unlocked the potential of remote sensing data for enhancing city-wide infrastructure monitoring \citep{gagliardi2023satellite} and traffic condition monitoring \citep{sheehan2023city}. High-resolution imagery data captures consistent, city-wide traffic conditions across all links at the same time, overcoming the limitations of ground-based sensors. Some pilot studies have demonstrated the effectiveness of traffic monitoring using satellite imagery data \citep{mccord2003estimating, larsen2009traffic, larsen2013automatic, kaack2019truck},  {and advanced computer vision algorithms support the extraction of fine-grained information from satellite images \citep{kawamura2025rn}.}
Additionally, Unmanned Aerial Vehicles (UAV) can be another similar but more flexible approach to collect precise imagery or video-based traffic data in small sub-regions of a city, due to their rapid speed, cost-effectiveness, and extensive field-of-view \citep{balamuralidhar2021multeye, butilua2022urban, bai2025dynamic, fonod2025advanced, espadaler2025accurate, xiong2025multi}. 

Some pilot studies have explored the utilization of these imagery data for transportation research, including road network extraction \citep{dai2020road,mei2021coanet}, vehicle detection and classification \citep{leitloff2010vehicle,chen2014vehicle,cao2016vehicle}. Some studies have focused on traffic density estimation based on imagery or video collected from satellites \citep{kopsiaftis2015vehicle, sakai2019traffic}. 
\cite{reksten_estimating_2021} proposed a deep learning-based framework to estimate annual average daily traffic (AADT) using very-high-resolution optical remote-sensing imagery with spatially limited ground-based measurements.
\cite{ganji_traffic_2022} used Google aerial images together with sparse data from traffic counts to predict traffic volumes. 
\cite{chen2021spatial} used traffic density detection data to examine changes in spatio-temporal traffic patterns during the COVID-19 pandemic.
\cite{he2021inferring} integrated satellite imagery and GPS trajectories to generate high-resolution accident risk maps.
\cite{yu2024sky} integrated high-resolution satellite imagery with geo-spatial tabular and crash data to analyze factors contributing to freight truck-related crashes.
\cite{horsler2023predicting} incorporated information from satellite imagery to predict regional road transport emissions, highlighting the potential in environmental impact assessments. 
Our initial work demonstrates the potential of incorporating satellite imagery into mesoscopic network model calibration but the information extraction from imagery data is not vehicle class specific and the parking information is not considered, leading to limited accuracy in large-scale network applications \citep{liu2024sat}.

To the best of our knowledge, none of the existing studies have incorporated detailed traffic information of both moving and on-street parking vehicles by class extracted from satellite images directly into calibrating dynamic network models and performing DODE. Incorporating satellite imagery has the potential to fill gaps in areas where ground-based sensors are not available, providing a more complete picture of consistent network conditions to enhance DODE performance. 

\begin{figure}[H]
    \centering
    \includegraphics[width=0.7\linewidth]{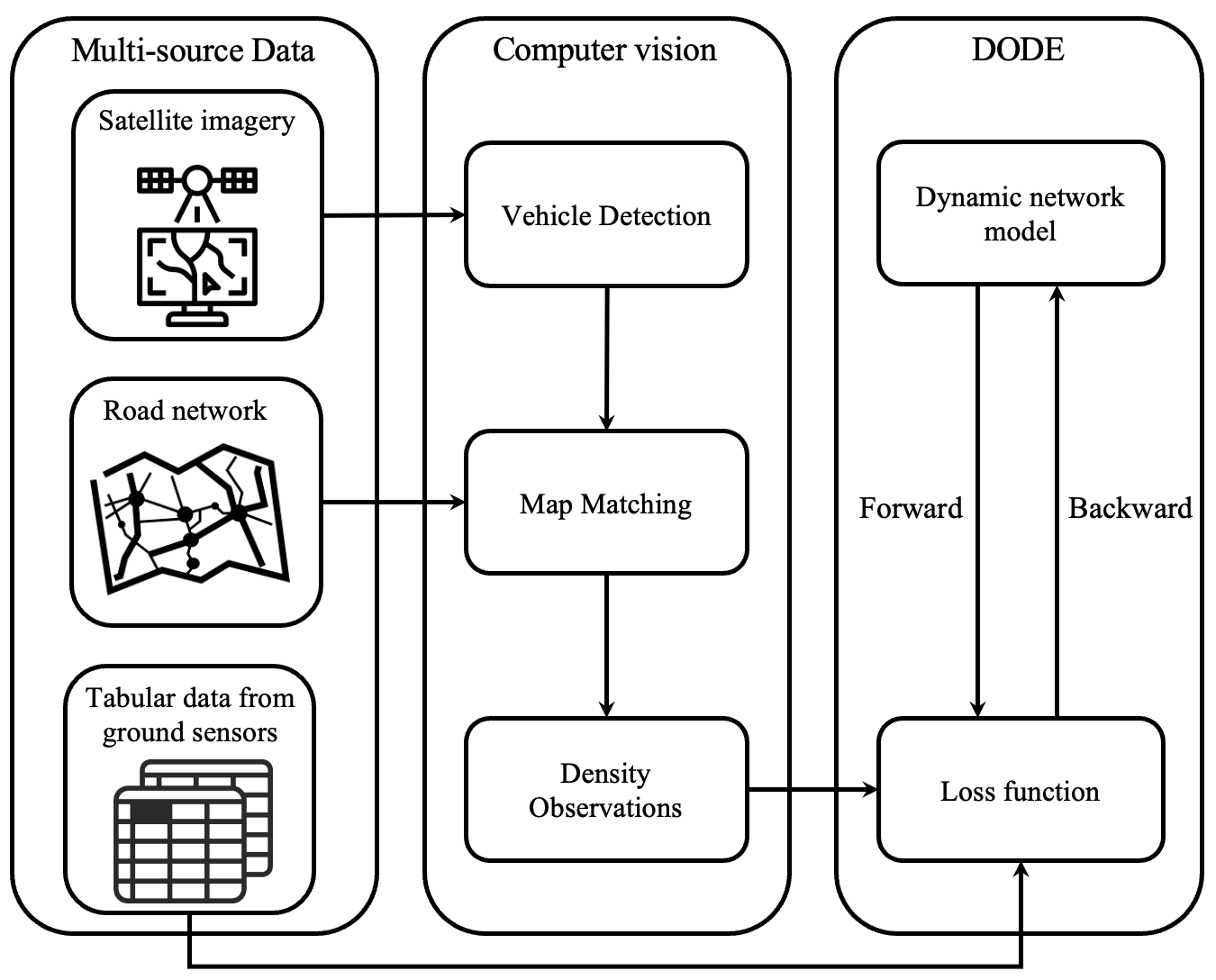}
    \caption{Model Overview}
    \label{fig:overview}
\end{figure}

\section{Model}\label{sec:model}
This section introduces the DODE framework leveraging high-resolution satellite imagery as an additional data source. Figure~\ref{fig:overview} demonstrates the overall framework, consisting of three parts: (1) multi-source data collection, including satellite imagery, road network, and tabular data from localized ground-based sensors. (2) a computer vision pipeline to extract traffic information from satellite imagery and (3) DODE model incorporating all available data sources. In the following subsections, we first introduce the computer vision pipeline for density extraction. Next, we formulate the DODE problem using traffic information from local ground-based detectors and satellite images, with an introduction to a general dynamic network loading model that explicitly models parking and on-road vehicles. Furthermore, we present the framework on a computational graph to solve DODE more efficiently using forward-backward algorithms. Lastly, we discuss data availability issues and corresponding aggregation methods.

\subsection{Computer vision pipeline for traffic information extraction from satellite images}
\subsubsection{Vehicle detection}
\label{sssec:vehicle-detection}
Vehicle detection in satellite imagery is a well-studied problem in computer vision, and various pre-trained models are available as open-source  implementations \citep{nachmany_detecting_2019}. \cite{liu2024sat} compared the performance of two model architectures provided in the library Raster Vision \citep{azaveaelement_84_robert_cheetham_raster_nodate}. The first architecture is a Single Shot Multibox Detector (SSD) model \citep{liu_ssd_2016}, which uses VGG as the backbone \citep{simonyan_very_2014} and the second one is a Faster R-CNN (FRCNN) model \citep{ren_faster_2017} that uses ResNet-50 as the backbone \citep{he_deep_2016}. \cite{liu2024sat} found that FRCNN achieves better results in terms of precision, recall, and F1-score, so this study adopts the same architecture but extends its application to a multi-class classification setting, enabling the model to distinguish between cars and trucks.

The object detection model was trained using high-resolution satellite imagery from the xView dataset \citep{lam_xview_2018}, which contains 744 annotated images at a 30-centimeter resolution. The dataset includes 60 object categories, 26 of which correspond to various types of vehicles. These 60 child categories are grouped into 7 parent classes. When labelers are unable to assign an object to a specific child category confidently, they assign it to the more general parent category \citep{lam_xview_2018}. Instead of classifying each object into its original xView category, the model was designed to perform a coarser multi-class classification task by categorizing detected objects as either \textit{car}, \textit{truck}, \textit{other vehicle}, or \textit{non-vehicle}. The \textit{car} class includes the parent class Passenger Vehicle and the child class Small Car. The \textit{truck} class includes the parent class Truck and nine truck-related child categories, such as Cargo Truck, Truck Tractor, and Haul Truck.

\subsubsection{Map matching and density calculation}
\label{sssec:map-matching}
Raster Vision provides built-in functionality to extract the geospatial coordinates of the bounding boxes associated with cars and trucks in the satellite imagery. Using standard geospatial analysis tools, the centroids of these bounding boxes were computed and subsequently matched to the nearest links in the street network. To avoid assigning vehicles parked far from road segments or on segments not included in the network, only centroid points of vehicles falling within a predefined buffer around each road centerline were considered eligible for map matching.

After cars and trucks are matched to the links of the street network, traffic density for each link can be calculated as the ratio of the number of matched vehicles to the corresponding link length. This computation is performed separately for cars and trucks, as the DODE framework is capable of incorporating multi-class traffic density data. Since the link lengths are exogenous information, we use the total detected and matched vehicle counts by class to represent link density without loss of generality.  

\subsection{Dynamic Origin-destination Demand Estimation (DODE)}
This section formulates the DODE problem that integrates multi-source mobility data, including link-level traffic counts, travel times, and additional traffic density extracted from satellite images. We begin by describing the approach to modeling link-level densities for different vehicle classes in a mesoscopic DNL model. Next, we formulate the DODE problem and present it on a computational graph, and we derive the forward and backward processes. Lastly, we discuss potential issues that may arise when deploying this framework in real-world scenarios with varying data availability and aggregation approaches.

\subsubsection{Preliminary}
Let $G = (N,A)$ represent a general road network with a node set $N$ and a link set $A$. A subset of links, denoted as $A_p\subset A$, represents links allowing on-street parking. The assignment horizon is defined as $T_d = [1,2,...,T]$. $R\subset N$ and $S\subset N$ are sets of origin and destination nodes, respectively. Between an origin-destination (OD) pair $rs$ where $r,s\in R,S$, $q^{rs}_{i,t}$ represents the total demand of vehicle class $i$ departing at time $t\in T_d$ from $r$ and heading to $s$. 
We consider two vehicle classes in this study: cars denoted by $\mathcal{C}$ and trucks denoted by $\mathcal{T}$.
$P^{rs}_i$ is the set of paths available for vehicle class $i$ to choose for the OD pair $rs$, and $f^{rs}_{k,i,t}$ represents the path flow (number of trips) of vehicle class $i$ departing at time $t$ and choosing path $k\in P^{rs}_{i}$. The corresponding path travel cost is denoted as $c^{rs}_{i,k,t}$. $p^{rs}_{k,i,t}$ represents the route choice proportion for path $k$ for vehicle class $i$ departing from $r$ at time $t$. Therefore, path flow $f^{rs}_{k,i,t}$ can be obtained by multiplying the route choice proportion $p^{rs}_{k,i,t}$ with OD demand $q^{rs}_{i,t}$.
\begin{equation}\label{eq:f=pq}
f^{rs}_{k,i,t} = p^{rs}_{k,i,t}\cdot q^{rs}_{i,t}, \ k\in P^{rs}_{i},\ rs\in RS,\ t\in T_d
\end{equation}
and in vectorized form
\begin{equation}\label{eq:vec_f=pq}
    \bm{f}_i = \bm{p}_i\bm{q}_i,\ \forall i\in \mathcal{C}
\end{equation}

The route choice proportions $p^{rs}_{k,i,t}$ can be determined using various route choice functions, such as dynamic user equilibrium \citep{pi2019general, liu2024modeling} and stochastic user equilibrium based on the Logit model \citep{guarda_estimating_2024}. The inputs depend on the specific formulation and here we use path and link level travel costs, denoted by $c^{rs}_{k,i,t}$ and $h^{t}_{a,i}$, without loss of generality.
\begin{equation}\label{eq:psi}
        p^{rs}_{k,i,t} = \Psi\left(\{c^{rs}_{k,i,t}\}_{rs,i,k,t},\{h^{t}_{a,i}\}_{i,a,t}\right)
\end{equation}
Vectorized form can be written as 
\begin{equation}\label{eq:vec_psi}
    \bm{p}_i = \Psi(\{\bm{c}_i\}_i, \{\bm{{h}}_i\}_i),\ \forall i\in \mathcal{C}
\end{equation}

In this study, the agent-based mesoscopic dynamic network loading (DNL) model, MAC-POSTS \citep{pi2018regional,Ma2020}, is used to simulate network-wide time-varying traffic conditions, including traffic counts, travel times and density. In the DNL, cumulative curves are installed at the head and tail of each observed link to extract time-varying traffic conditions. 

For links with traffic flow and/or travel time observations, we use two cumulative curves recording cumulative numbers of arriving and departing through vehicles , respectively on link $a$, denoted by $\mathcal{A}^m_{a,i}$ and $\mathcal{D}^m_{a,i}$ respectively. For links with density observations, especially links in $A_p$ with on-street parking spaces, we install two additional cumulative curves at the head and tail of link $a$, denoted by $\mathcal{A}^p_{a,i}$ and $\mathcal{D}^p_{a,i}$ respectively, to record inflows and outflows for parking vehicles. 

\begin{figure}[ht]
    \centering
    \begin{subfigure}{0.495\textwidth}
    \centering
        \includegraphics[width=\linewidth]{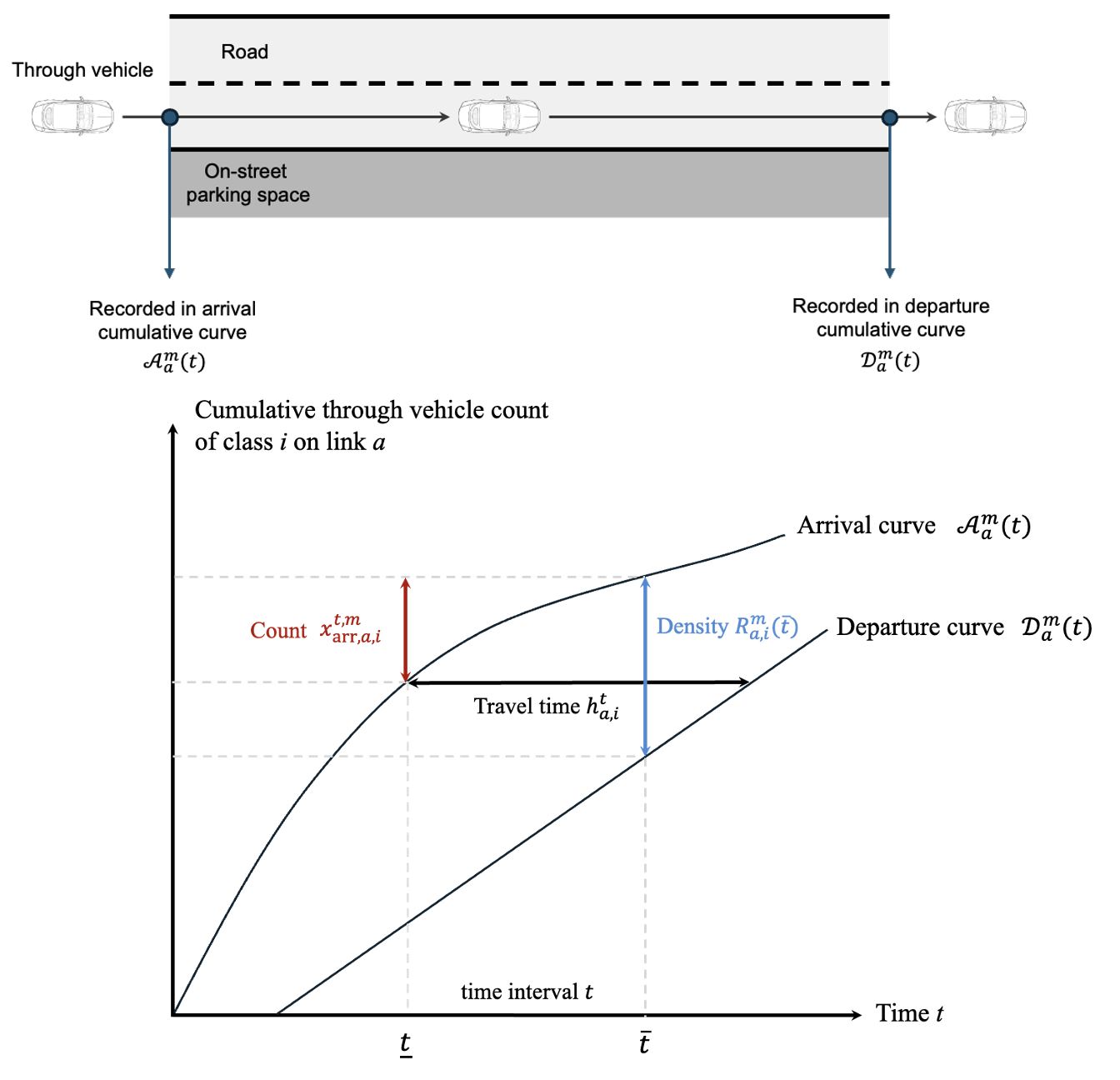}
        \caption{Cumulative curves for vehicles in through traffic}
    \end{subfigure}
    \begin{subfigure}{0.495\textwidth}
    \centering
        \includegraphics[width=\linewidth]{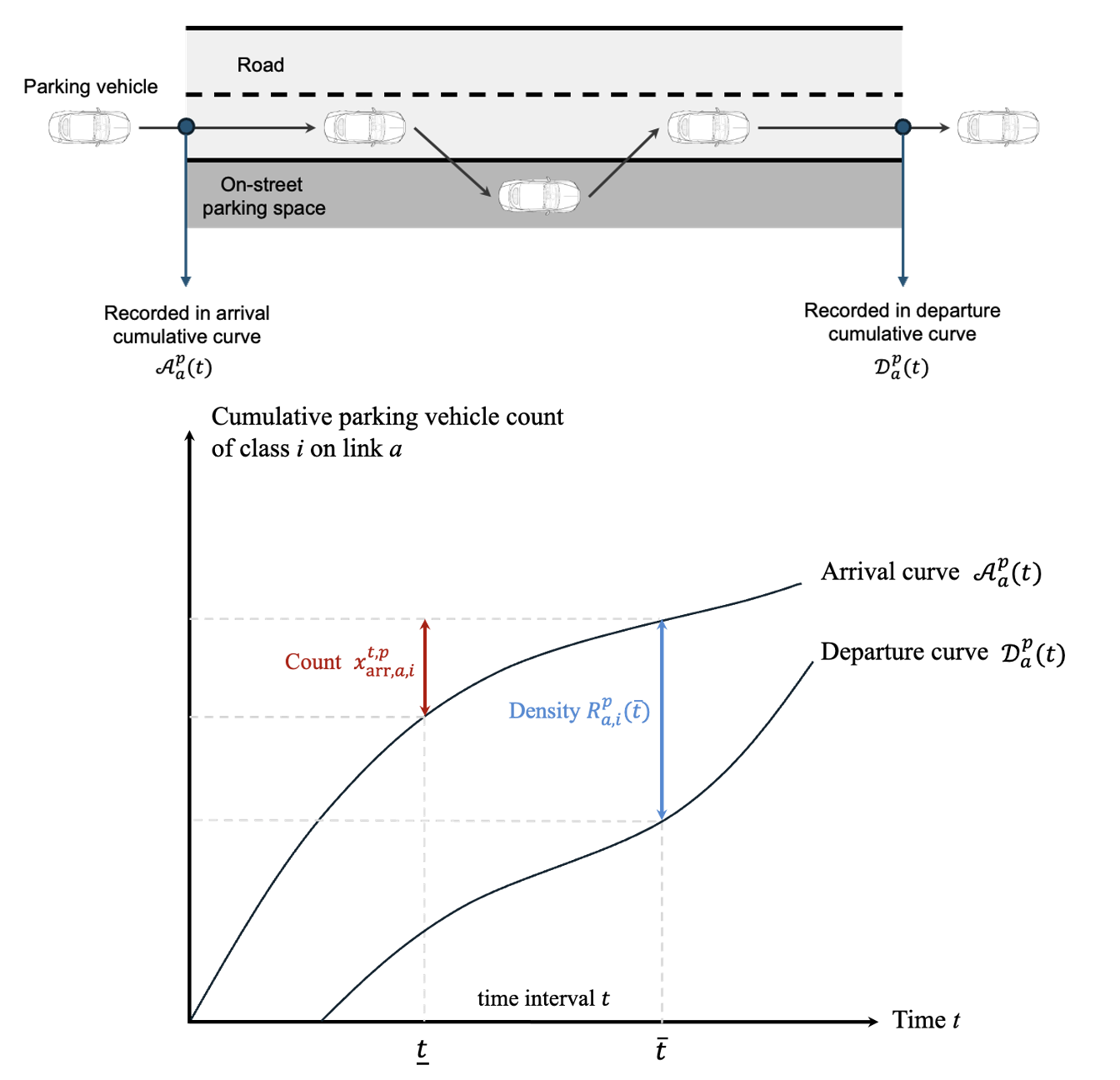}
        \caption{Cumulative curves for on-street parking vehicles} 
    \end{subfigure}
    \caption{Traffic condition extracted from cumulative curves}
    \label{fig:cumulative_curve}
\end{figure}

Figure~\ref{fig:cumulative_curve} demonstrates the setting of two pairs of cumulative curves at observed links with on-street parking spaces. To efficiently extract DAR matrices, these cumulative curves are tree-based \citep{Ma2020,liu2024modeling}. 

We incorporate curbside parking dynamics into the DNL model using the approach proposed in \cite{liu2024modeling}. Truncated fundamental diagrams with respect to curb occupancies are used to govern link dynamics, which can be generalized to any link dynamics model, such as link queue or link transmission models. For instance, for a link queue model, we can use link-level parking queues (instead of curb cells in CTM) to store parking vehicles, with the same curb parking dynamics to track curb availability and departure times. 
 {Because the DNL model runs with its queue-based link and node dynamics, First-In-First-Out (FIFO) is explicitly enforced. It uses separate cumulative curves for parking and through vehicles, ensuring that travel times are extracted from FIFO-preserving cumulative curves of through vehicles without the impact of parking-related non-FIFO behaviors.}
Using these two pairs of cumulative curves, the traffic count of class $i$ during time interval $t$ is modeled as the total arrival count of both parking and through vehicles during time interval $t$, shown as
\begin{equation}
    x^{t}_{\text{arr},a,i} = x^{t,m}_{\text{arr},a,i} + x^{t,p}_{\text{arr},a,i} = \mathcal{A}^{m}_{a,i}(\bar{t}) - \mathcal{A}^{m}_{a,i}(\underline{t}) + \mathcal{A}^{p}_{a,i}(\bar{t}) - \mathcal{A}^{p}_{a,i}(\underline{t})
\end{equation}

Link travel time at time $t$ is modeled as the traversal time for vehicles arriving at time $t$ on link $a$, extracted from cumulative curves of through vehicles, which reads
\begin{equation}\label{eq:linkdynamic}
    h^{t}_{a,i} = \mathcal{D}^{m,-1}_{a,i}(\mathcal{A}^{m}_{a,i}(t)) - t = \bar{\Lambda}\left(x^{t}_{\text{arr},a,i}\right)
\end{equation}
where $\bar{\Lambda}$ is a general link dynamic function using link inflow as input because both arrival and departure cumulative curves can be seen as a function of link inflow, and link flow dynamics governs the relations between two curves.

\subsubsection{Modeling traffic density} \label{sec:model_conditions}
Traffic density is a measure of the number of vehicles per unit length of road. Because road length is exogenous, we use the total number of vehicles remaining on link $a$ at timestamp $\bar{t}$, denoted by $R_{a,i}(\bar{t})$ to represent the link density, which can be modeled as the summation of remaining through and parking vehicle numbers at time $t$, each of them is the difference between the cumulative arrival vehicle number and the cumulative departure vehicle number before timestamp $t$. As our DNL model is discrete in time, $R_{a,i}(\bar{t})$ (the upper bound) is used to represent the density for time interval $t$.
\begin{equation}\label{eq:k}
    R_{a,i}(\bar{t}) = \mathcal{A}^{m}_{a,i}(\bar{t}) + \mathcal{A}^p_{a,i}(\bar{t})  - \mathcal{D}^{m}_{a,i}(\bar{t})- \mathcal{D}^p_{a,i}(\bar{t})
\end{equation}
and the link density at time $\bar{t}$ is 
\begin{equation}\label{eq:k_t}
    k^{{t}}_{a,i} = \frac{R_{a,i}(\bar{t})}{l_a}
\end{equation}

To obtain more stable and smooth density estimation, we average the densities across several neighboring time intervals around time ${t}$, denoted by $[{t} - \delta, {t} + \delta]$. The average density can be computed by
\begin{equation}
    k^{{t}}_{a,i}\cdot l_a = \frac{1}{2\delta + 1}\sum_{t'={t}-\delta}^{{t}+\delta}k^{t'}_{a,i}\cdot l_a
\end{equation}

 {To make the simulation-based DODE problem tractable, we linearize the DNL process using Dynamic Assignment Ratio (DAR) matrices to express the mapping between path flows and link flows in a differentiable form for the computational-graph implementation. The estimated density computation is also built upon DAR matrices.} Because both arrival and departure link flows are used in estimating remaining vehicle numbers, and both parking and non-parking vehicles are accounted for link inflows, four separate DAR matrices are used to record the corresponding mapping relationships. The traffic flow, represented as link inflow in our DNL, during time $t_2$ has the following relationship with path flows
\begin{equation}\label{eq:arr}
    x^{t_2}_{\text{arr},a,i} = \sum_{r,s\in R,S}\sum_{k\in P^{rs}_i}\sum_{t_1\in T_d}\rho_{\text{arr},k,a,i}^{rs,m}(t_1,t_2)\cdot f^{rs}_{k,i,t_1} + \sum_{r,s\in R,S}\sum_{k\in P^{rs}_i}\sum_{t_1\in T_d}\rho_{\text{arr},k,a,i}^{rs,p}(t_1,t_2)\cdot f^{rs}_{k,i,t_1}
\end{equation}
where $\rho_{\text{arr},k,a,i}^{rs,m/p}(t_1,t_2)$ represents the time-dependent incidence matrix, which is the proportion of the path flow $f^{rs}_{k,i,t_1}$ contributing to the link arrival flow $x^{t_2,n/p}_{\text{arr},a,i}$. Here we use $t_1$ and $t_2$ to differentiate the time index for path and link flows respectively. The corresponding vectorized form can be written as
\begin{equation}\label{eq:vec_flow}
    \bm{x}_i = \bm{\rho}^{m}_{a,i}\bm{f}_i + \bm{\rho}^{p}_{a,i}\bm{f}_i
\end{equation}
Similarly, the departure flow of link $a$ at time $t_2$ for class $i$ can be modeled using the departure DAR matrix as
\begin{equation}\label{eq:dep}
    x^{t_2}_{\text{dep},a,i} = \sum_{r,s\in R,S}\sum_{k\in P^{rs}_i}\sum_{t_1\in T_d}\rho_{\text{dep},k,a,i}^{rs,m}(t_1,t_2)\cdot f^{rs}_{k,i,t_1} + \sum_{r,s\in R,S}\sum_{k\in P^{rs}_i}\sum_{t_1\in T_d}\rho_{\text{dep},k,a,i}^{rs,p}(t_1,t_2)\cdot f^{rs}_{k,i,t_1}
\end{equation}
Then the cumulative arrival and departure flows on link $a$ at time ${t}$ for vehicle class $i$ can be computed as the summation of all arrival or departure flows before time ${t}$, that are arrival or departure flows of intervals from $1$ to $t$,
\begin{equation}\label{eq:AD}
    A_{a,i}(\bar{t}) = \sum_{t_2=1}^{t}x^{t_2}_{\text{arr},a,i}\quad \text{and}\quad
    D_{a,i}(\bar{t}) = \sum_{t_2=1}^{t}x^{t_2}_{\text{dep},a,i}
\end{equation}
Combining Equation~\ref{eq:arr},\ref{eq:dep} and \ref{eq:AD}, link density and path flow have the following linear relationship
\begin{equation}\label{eq:kl}
\begin{aligned}
    k^{t}_{a,i}\cdot l_a &= R_{a,i}(\bar{t}) = A_{a,i}(\bar{t}) - D_{a,i}(\bar{t})\\
    & = \sum_{t_2=1}^{t}\sum_{r,s\in R,S}\sum_{k\in P^{rs}_i}\sum_{t_1\in T_d}\rho_{\text{dep},k,a,i}^{rs,m}(t_1,t_2)\cdot f^{rs}_{k,i,t_1} + \sum_{t_2=1}^{t}\sum_{r,s\in R,S}\sum_{k\in P^{rs}_i}\sum_{t_1\in T_d}\rho_{\text{dep},k,a,i}^{rs,p}(t_1,t_2)\cdot f^{rs}_{k,i,t_1}\\
    &- \sum_{t_2=1}^{t}\sum_{r,s\in R,S}\sum_{k\in P^{rs}_i}\sum_{t_1\in T_d}\rho_{\text{dep},k,a,i}^{rs,m}(t_1,t_2)\cdot f^{rs}_{k,i,t_1} - \sum_{t_2=1}^{t}\sum_{r,s\in R,S}\sum_{k\in P^{rs}_i}\sum_{t_1\in T_d}\rho_{\text{dep},k,a,i}^{rs,p}(t_1,t_2)\cdot f^{rs}_{k,i,t_1}\\
\end{aligned}
\end{equation}
The corresponding vectorized form is
\begin{equation}\label{eq:vec_k}
    \bm{k}_i\odot \bm{l} = \bm{H}_i(\bm{\rho}^{m}_{a,i}\bm{f}_i + \bm{\rho}^{p}_{a,i}\bm{f}_i) - \bm{H}_i(\bm{\rho}^{m}_{d,i}\bm{f}_i + \bm{\rho}^{p}_{d,i}\bm{f}_i)
\end{equation}
where $\bm{H}_i$ is used to aggregate total arrival and departure flows from time interval $1$ to $t_2$ in Equation~\ref{eq:kl}.

To summarize, the DNL model, denoted by $\Lambda$, uses time-varying path flows as inputs and outputs link-level traffic conditions and DAR matrices that record the mapping relationships between path flows and link flows, shown in Equation~\eqref{dnl}. 
\begin{equation}
\label{dnl}
\begin{aligned}
    &\left\{
    x^{t_2}_{\text{arr},a,i},
    h^{t_2}_{a,i}, 
    k^{t_2}_{a,i},
    \rho_{\text{arr},k,a,i}^{rs,m}(t_1,t_2),
    \rho_{\text{arr},k,a,i}^{rs,p}(t_1,t_2),
    \rho_{\text{dep},k,a,i}^{rs,m}(t_1,t_2),
    \rho_{\text{dep},k,a,i}^{rs,p}(t_1,t_2)
    \right\}_{rs,i,k,a,t_1,t_2} \\
    & = \Lambda\left(\{f^{rs}_{k,i,t}\}_{rs,k,i,t}\right)\\
\end{aligned}
\end{equation}
In vectorized form, the DNL function can be written as 
\begin{equation}\label{eq:vec_lambda}
    \{\bm{x}_i, \bm{{h}}_i, \bm{k}_i, \bm{\rho}^{m}_{a,i}, \bm{\rho}^{p}_{a,i}, \bm{\rho}^{m}_{d,i}, \bm{\rho}^{p}_{d,i}\}_i = \Lambda(\{\bm{f}_i\}_i),\ \forall i\in \mathcal{C}
\end{equation}

\subsubsection{DODE formulation} 
The DODE can be formulated as a mathematical programming problem shown in Equation~\ref{eq:DODE}. The decision variables are OD demand $\{\bm{q}_i\}_i$. The objective function minimizes the differences between modeled and observed traffic conditions, including link traffic counts, link travel times, and link density. The constraints include route choice, DNL, the path-demand relationship and non-negativity constraint of demand.

\begin{equation}
\label{eq:DODE}
\begin{aligned}
    \min_{\{\bm{q}_i\}_i}\quad  &\mathscr{L}
    = \ w_{1}{\left(\left\Vert \bm{x}^o - \sum_{i} \bm{L}_i \bm{{x}}_i\right\Vert^2_2\right)} + w_{2}{\left(\left\Vert \bm{h}^o - \sum_{i} \bm{M}_i\bm{{h}}_i\right\Vert^2_2\right)} + w_{3}{\left(\left\Vert \bm{k}^o\odot \bm{l} - \sum_{i} \bm{I}_i \bm{{k}}_i\odot \bm{l}  \right\Vert^2_2\right)}\\
    \text{s.t.}\quad &\text{Equations~\eqref{eq:vec_f=pq}, \eqref{eq:vec_psi}}, \eqref{eq:vec_flow}, \eqref{eq:vec_k}, \eqref{eq:vec_lambda}\\
    & \bm{q}_i \geq 0,\ \forall i\in \mathcal{C}\\
\end{aligned}
\end{equation}
where $\bm{x}^o$, $\bm{h}^o$ and $\bm{k}^o\odot \bm{l}$ are observed counts, travel times, and remaining vehicles (density multiplied by link length). $\bm{L}_i$, $\bm{M}_i$ and $\bm{I}_i$ are aggregation matrices used to map simulated conditions to available observations, depending on data availability which will be discussed in section \ref{sec:data_availability}.

\subsection{Computational graph-based solution algorithm}
\begin{figure}[H]
    \centering
    \begin{subfigure}{\textwidth}
        \centering
        \includegraphics[width=\linewidth]{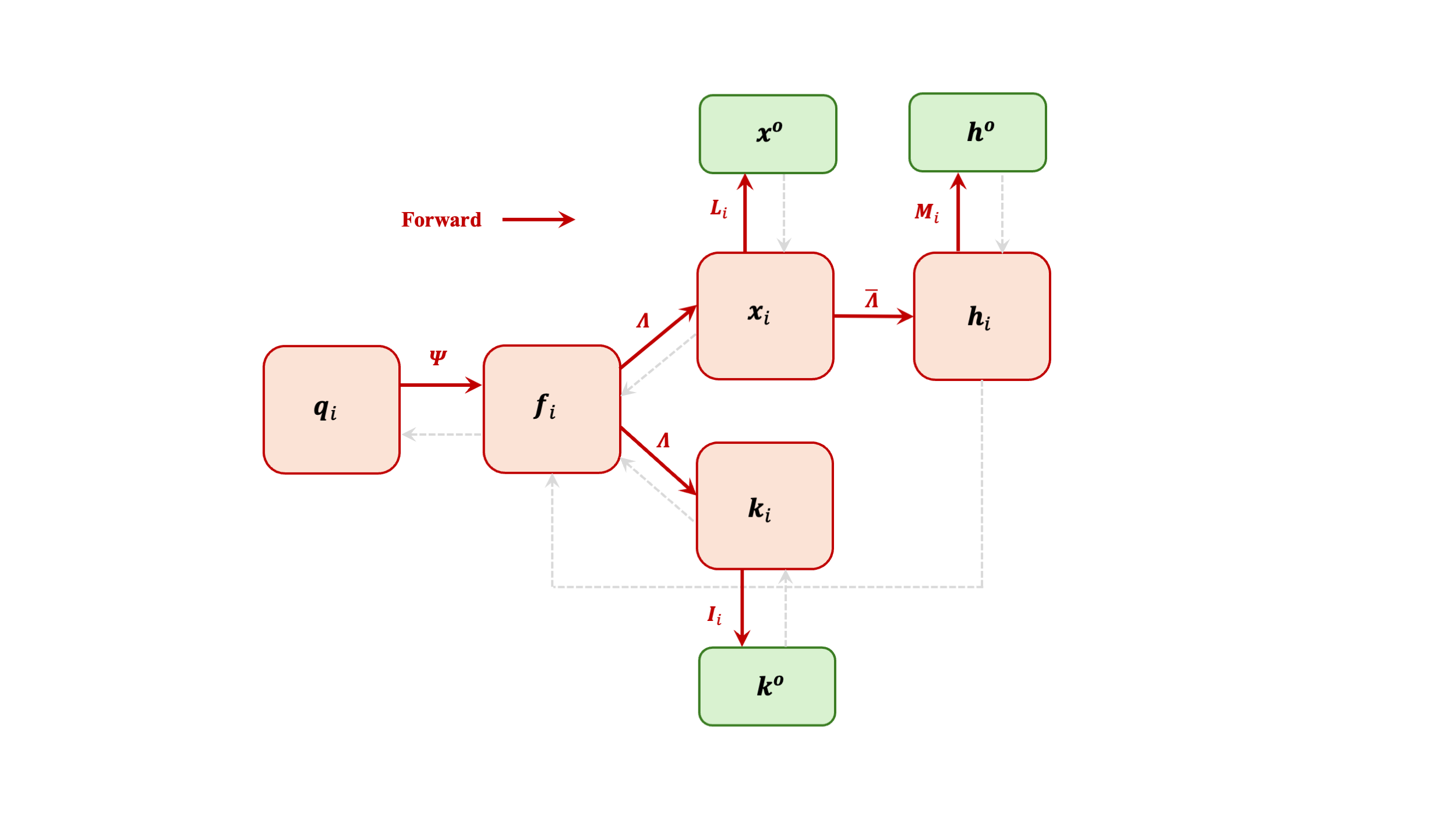}
        \caption{Forward process}
    \end{subfigure}
    \begin{subfigure}{\textwidth}
        \centering
        \includegraphics[width=\textwidth]{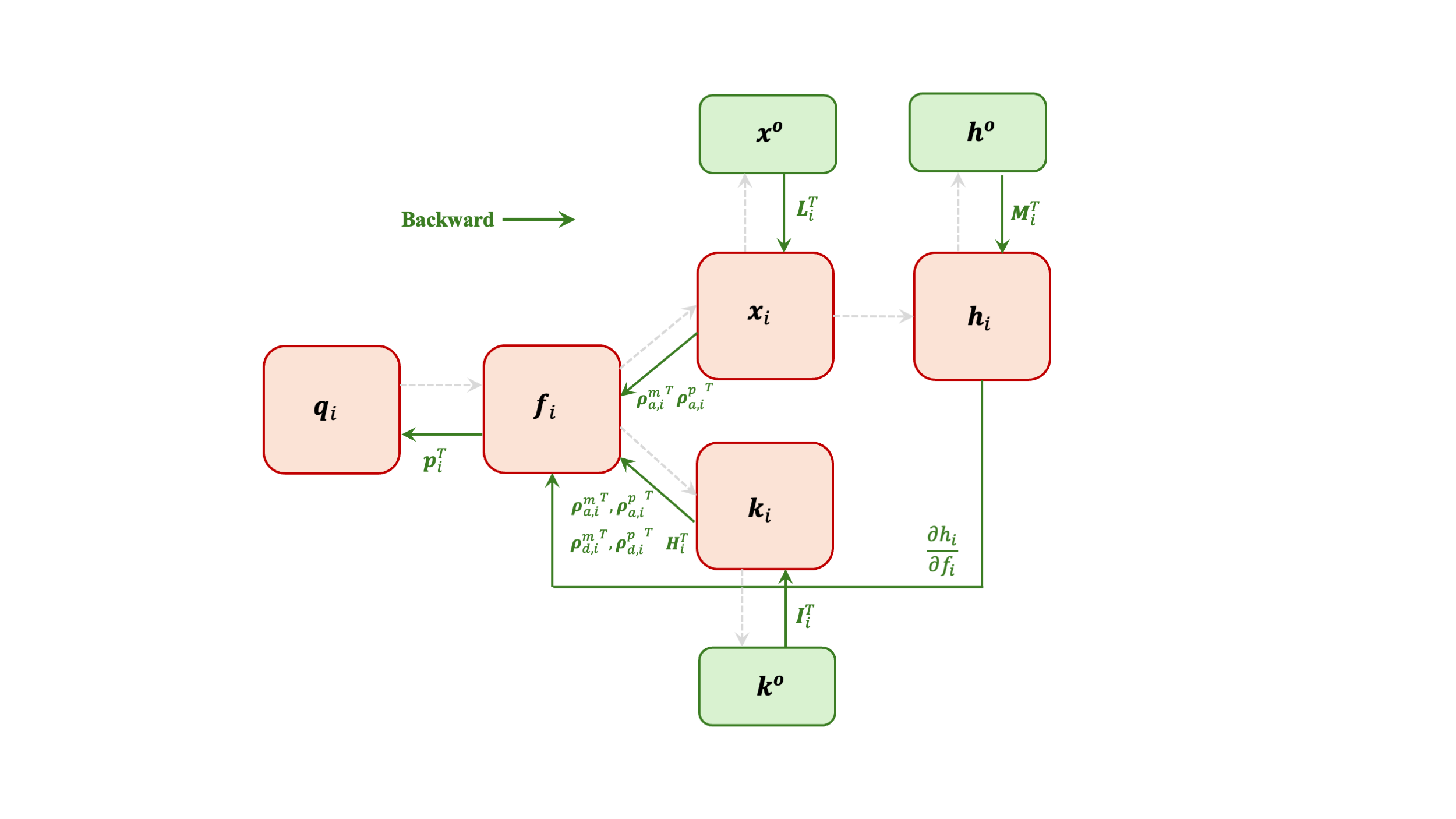}
        \caption{Backward process}
    \end{subfigure}
    \caption{Computational graph of DODE}
    \label{fig:CG}
\end{figure}

The formulation can be presented on a computational graph and solved by forward-backward algorithms efficiently. In the forward pass shown in Figure~\ref{fig:CG}, a dynamic traffic assignment problem is solved using the behavior function $\Psi$ and DNL model $\Lambda$, given fixed time-varying multi-class demand $\{\bm{q}_i\}_i$ (initialized demand or estimated demand from the previous epoch). Route choices $\{\bm{p}_{i}\}_i$, link-level traffic states and DAR matrices $\bm{\rho}^{m}_{a,i}, \bm{\rho}^{p}_{a,i}, \bm{\rho}^{m}_{d,i}, \bm{\rho}^{p}_{d,i}$ are obtained from Equations~\eqref{eq:vec_lambda}, \eqref{eq:vec_psi} and \eqref{eq:vec_f=pq}. To improve computation efficiency in DNL on large-scale networks, we can use the hybrid route choice models which is embedded in the DNL model to retrieve link traffic measurements and DAR matrices in one DNL run. In the hybrid route choice model, a fraction of travelers follow a pre-trip route choice model and the other travelers adaptively explore time-varying shortest paths following an en-route route choice model \citep{qian2013hybrid, liu2024modeling}.

In the backward pass, shown in Figure~\ref{fig:CG}, the gradient of the loss function with respect to demand is derived using the Chain Rule in Equation~\ref{eq:back_eq}. The gradient of the objective function with respect to the decision variable $\bm{q}_i$ is shown in Equation~\ref{eq:grad}. Any gradient-based solution algorithm can be used to solve the DODE based on this gradient on the computational graph.

\begin{equation}
\label{eq:back_eq}
    \begin{aligned}
        \frac{\partial \mathscr{L}}{\partial \mathscr{L}_1} &= w_1\\
        \frac{\partial \mathscr{L}_1}{\partial \bm{x}_i} &= -2\bm{L}_i^T\left(\bm{x}^o - \sum_{i\in C}\bm{L}_{i} \bm{x}_i\right)\\
        \frac{\partial \mathscr{L}_1}{\partial \bm{f}_i} &= ({\bm{\rho}^{m}_{a,i}}^\intercal + {\bm{\rho}^{p}_{a,i}}^\intercal)\frac{\partial \mathscr{L}_1}{\partial \bm{x}_i}\\
        \frac{\partial \mathscr{L}_1}{\partial \bm{q}_i} &= \bm{p}^T_i\frac{\partial \mathscr{L}_1}{\partial \bm{f}_i}\\
        & \\
        \frac{\partial \mathscr{L}}{\partial \mathscr{L}_{2}} &= w_2\\
        \frac{\partial \mathscr{L}_{2}}{\partial \bm{h}_i} &= -2\bm{M}_i^T\left(\bm{h}^o - \sum_{i\in C}\bm{M}_{i}\bm{h}_{i}\right)\\
        \frac{\partial \mathscr{L}_{2}}{\partial \bm{f}_i} &= \frac{\partial \bm{h}_i}{\partial \bm{f}_i}\frac{\partial \mathscr{L}_{2}}{\partial \bm{h}_i}\\
        \frac{\partial \mathscr{L}_{2}}{\partial \bm{q}_i} &= \bm{p}^T_i\frac{\partial \mathscr{L}_{2}}{\partial \bm{f}_i}\\
        & \\
        \frac{\partial \mathscr{L}}{\partial \mathscr{L}_{3}} &= w_3\\
        \frac{\partial \mathscr{L}_{3}}{\partial \bm{r}_i} &= -2\bm{I}^T_{i}\left(\bm{k}^o\odot \bm{l} - \sum_{i} \bm{I}_i\bm{k}_i\odot \bm{l}\right)\\
        \frac{\partial \mathscr{L}_{3}}{\partial \bm{f}_i} &= \left({\bm{\rho}^{m}_{a,i}}^\intercal + {\bm{\rho}^{p}_{a,i}}^\intercal - {\bm{\rho}^{m}_{d,i}}^\intercal - {\bm{\rho}^{p}_{d,i}}^\intercal\right)\bm{H}_i^\intercal\frac{\partial \mathscr{L}_{3}}{\partial \bm{r}_i}\\
        \frac{\partial \mathscr{L}_{3}}{\partial \bm{q}_i} &= \bm{p}^T_i\frac{\partial \mathscr{L}_{3}}{\partial \bm{f}_i}\\
    \end{aligned}
\end{equation}

\begin{equation}
\label{eq:grad}
\begin{aligned}
    \frac{\partial \mathscr{L}}{\partial \bm{q}_i} 
    = &-2 w_{1} \bm{p}_i^\intercal ({\bm{\rho}^{m}_{a,i}}^\intercal + {\bm{\rho}^{p}_{a,i}}^\intercal) \bm{L}_i^\intercal \left(\bm{x}^o - \sum_{i}\bm{L}_{i} \bm{x}_i\right)-2w_{2}\bm{p}^\intercal_i\frac{\partial {\bm{{h}}_{i}}}{\partial \bm{f}_i}\bm{M}_i^\intercal\left(\bm{h}^o - \sum_{i}\bm{M}_{i}\bm{{h}}_{i}\right)\\
    &-2w_{3} \bm{p}^\intercal_i \cdot \left({\bm{\rho}^{m}_{a,i}}^\intercal + {\bm{\rho}^{p}_{a,i}}^\intercal - {\bm{\rho}^{m}_{d,i}}^\intercal - {\bm{\rho}^{p}_{d,i}}^\intercal\right)\bm{H}_i^\intercal\bm{I}_i^\intercal\left(\bm{k}^o\odot \bm{l} - \sum_{i} \bm{I}_i\bm{k}_i\odot \bm{l}\right)\\
\end{aligned}
\end{equation}

\subsection{Discussions}\label{sec:data_availability}
This section first addresses issues related to data availability and aggregation methods. The DNL model simulates network-wide traffic conditions at various spatio-temporal resolutions of interest, but observed data may not be collected at the same frequency. Moreover, data quality and pre-processing errors may not yield precise spatio-temporal observations. To address this, we use aggregation matrices to map modeled conditions to observations, and aggregation can occur in both spatial and temporal domains. Temporal aggregation refers to aggregating traffic conditions into larger time intervals. For example, summing four 15-min traffic counts to match hourly traffic flow observation \citep{Ma2020,liu2024modeling}.  {In this study, satellite imagery provides snapshot-like density observations with a known capture time window, and each density snapshot is assigned to the corresponding 15-min interval in the DNL model. For consistency, ground-sensor observations are also aggregated to the same 15-minute resolution. We use spatio-temporal aggregation operators to map simulated link states to the exact time intervals where measurements exist. This allows the estimation objective to incorporate heterogeneous data streams in a unified way.}

Spatial aggregation primarily arises from limitations in satellite image quality and the computer vision pipeline's performance. We discuss three potential scenarios that require spatial aggregation. The first scenario is about lane-level vehicle detection. When CV pipeline does not have precise road segmentation capabilities, it cannot conduct lane-level vehicle detection, failing to differentiate between parking and through vehicles. In such cases, observations provide link density for all vehicles on the link and we aggregate parking and through vehicles together to model density, which is already described in Equation~\eqref{eq:k} and \eqref{eq:k_t}. In this study, we integrate the aggregation into density estimation in the DNL model. Another general way is to estimate separate parking density and through traffic density in the DNL and using the matrix $\bm{I}_i$ to aggregate them. Another reason we aggregate them in the DNL process is that one through vehicle captured by a satellite image only represent the status at the timestamp of the image, and we are not sure if this vehicle will perform on-street parking after this timestamp or just left a parking spot before this timestamp. Aggregating through and parking vehicles into total density can overcome this information gap.
The second scenario is about vehicle moving directions. Similar to scenario 1, if the CV pipeline cannot perform road segmentation and detect vehicle orientations, detected vehicles cannot be correctly matched to directed road segments, potentially leading to flow inconsistency on successive links. To address this, we aggregate two (even more) moving directions together to match regional total density. Another scenario is about vehicle class detection. When the CV pipeline fails to detect vehicle classes with high precision, we can simply aggregate density across vehicle classes. 

 {This study extends the work of \cite{liu2024sat} in both modeling and image data processing. This study incorporates curbside parking in dynamic network loading to explicitly model densities of both parking and moving vehicles, consistent with what satellite imagery captures, while \cite{liu2024sat} did not differentiate moving and parking vehicles. Moreover, this study employs a class-specific vehicle detection algorithm to process satellite imagery data, improving the fidelity of density observations and enabling multi-class DODE, while \cite{liu2024sat} used aggregated density for all vehicle classes. Note that the results of this study and \cite{liu2024sat} cannot be compared directly due to the inherently underdetermined nature of DODE and different “ground truth” inputs.}

\section{Numerical Experiments}\label{sec:experiment}
\subsection{Toy network with synthetic data}
\label{sec:toy_net}
\begin{figure}[H]
    \centering
    \includegraphics[width=0.85\linewidth]{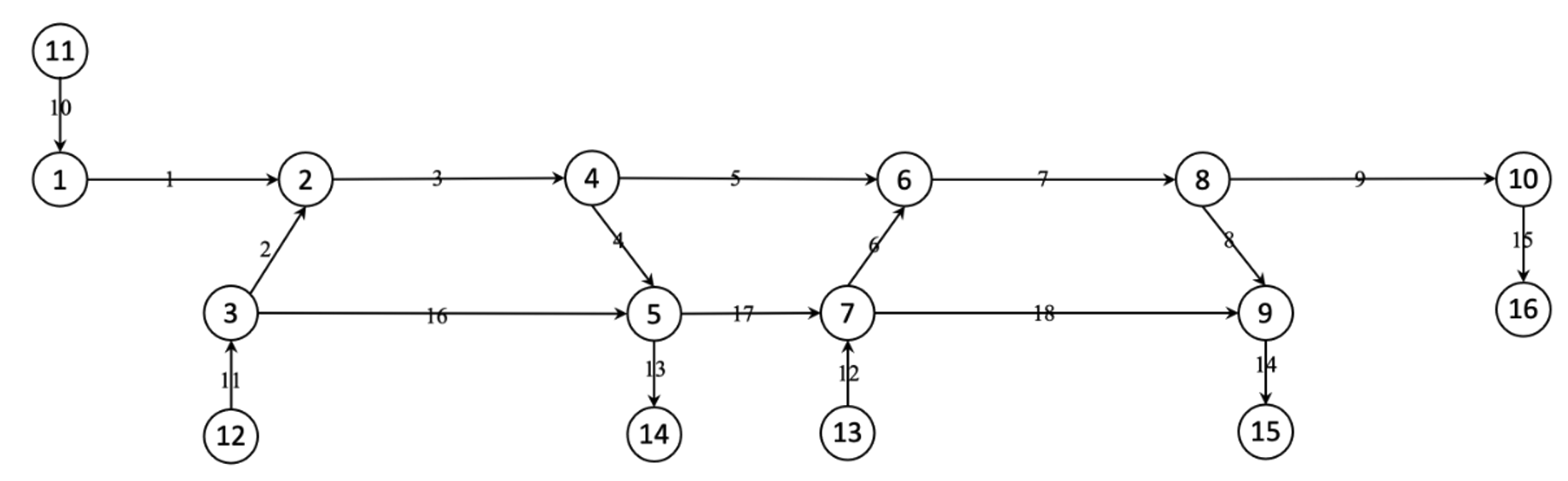}
    \caption{Small network}
    \label{fig:toy_net}
\end{figure}

We begin with a synthetic toy network to evaluate the effectiveness of incorporating density observations into DODE. The network, shown in Figure~\ref{fig:toy_net} is modified from \cite{zhang2020path} and consists of 18 links, including 12 road segments and 6 OD connectors. Detailed link configurations can be found in Table~\ref{tab:nie_network}. Time-varying traffic conditions, including traffic counts, travel times and densities, are generated based on synthetic demand and recorded at 15-minute intervals to serve as ground truth observations. As described in Section \ref{sec:model}, we aggregate moving and parking vehicles together to calculate the total remaining vehicles on the segment to represent link densities.

To simulate measurement noise, traffic count and travel time observations are corrupted by multiplying by random error terms $\epsilon_{10\%}$ drawn from a uniform distribution $\text{Unif}(0.9, 1.1)$. The densities are also corrupted by random multiplicative errors of three levels with corresponding uniform distributions: (1) $\pm 10\%$ with $\epsilon_{10\%}\sim\text{Unif}(0.9, 1.1)$ (2) $\pm 20\%$ with $\epsilon_{20\%}\sim\text{Unif}(0.8, 1.2)$ and (3) $\pm 50\%$ with $\epsilon_{50\%}\sim\text{Unif}(0.5, 1.5)$.

To simulate the sparsity of real-world observations, 6 out of 12 links are randomly selected to form the observed link set, denoted by $A_{obs}$. The remaining 6 links are unobserved by local sensors, denoted by the link set $A_{unobs}$. To show that incorporating density snapshots of the whole network can enhance overall DODE performance, we conduct two separate DODE tests under two data scenarios and compare estimation accuracy on both $A_{obs}$ and $A_{unobs}$. In scenario 1, we use only traffic counts and travel times from $A_{obs}$, represented as $\{x_a^t, h_a^t\}_{t\in T_d, a\in A_{obs}}$ to mimic the DODE process using localized sensor data. In scenario 2, the data is augmented with 10 density snapshots of all 12 links, taken at 15-minute intervals (the same as those of traffic count and travel time data), denoted by $\{k_a^t\}_{t\in T_d, a\in A}$. 

\begin{figure}[H]
    \centering
    \begin{subfigure}{0.495\textwidth}
        \centering
        \includegraphics[width=\linewidth]{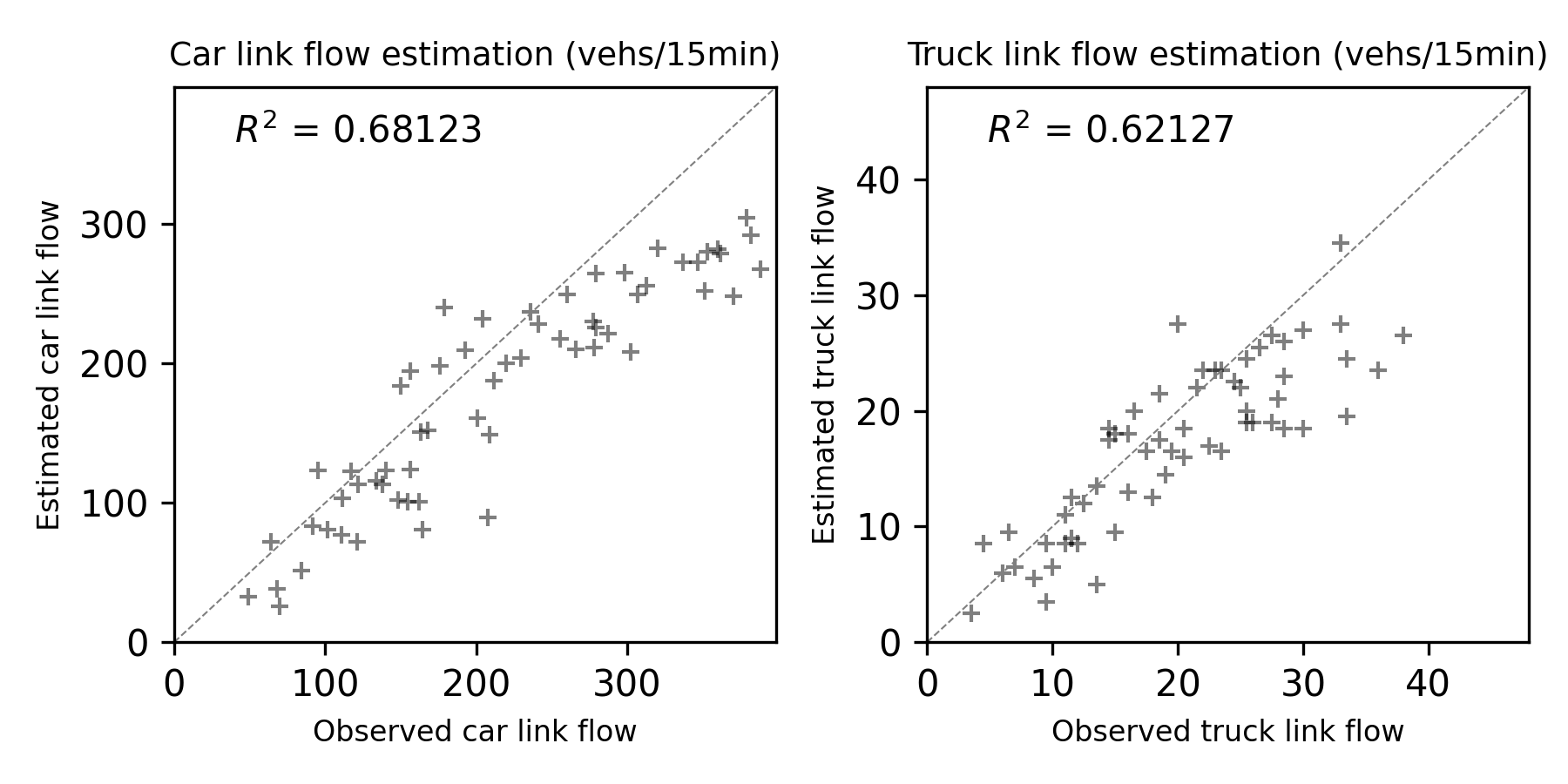}
        \caption{Scenario 1}
    \end{subfigure}
    \begin{subfigure}{0.495\textwidth}
        \centering
        \includegraphics[width=\textwidth]{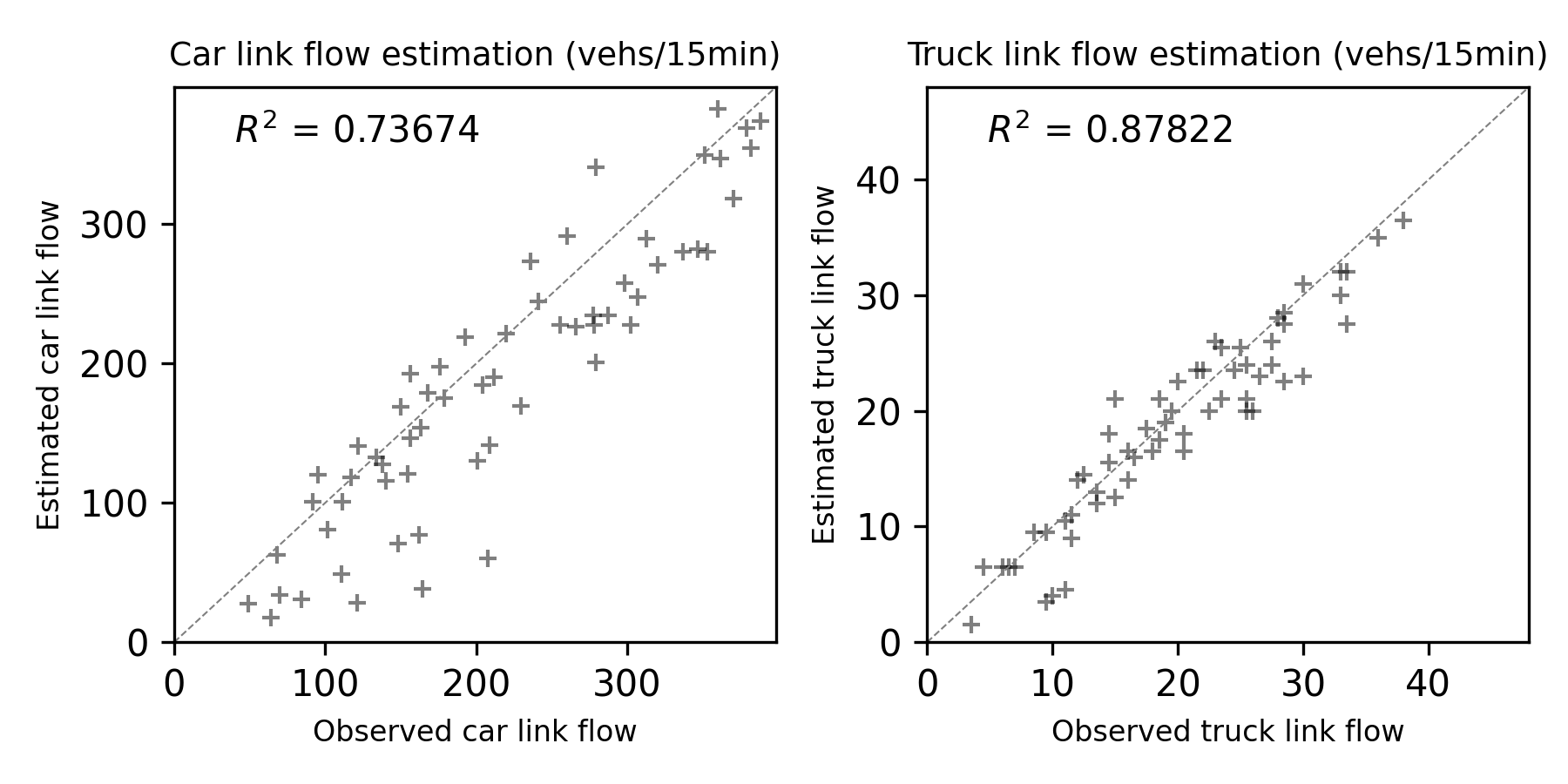}
        \caption{Scenario 2 ($\pm 10\%$ error in densities)}
    \end{subfigure}
    \begin{subfigure}{0.495\textwidth}
        \centering
        \includegraphics[width=\textwidth]{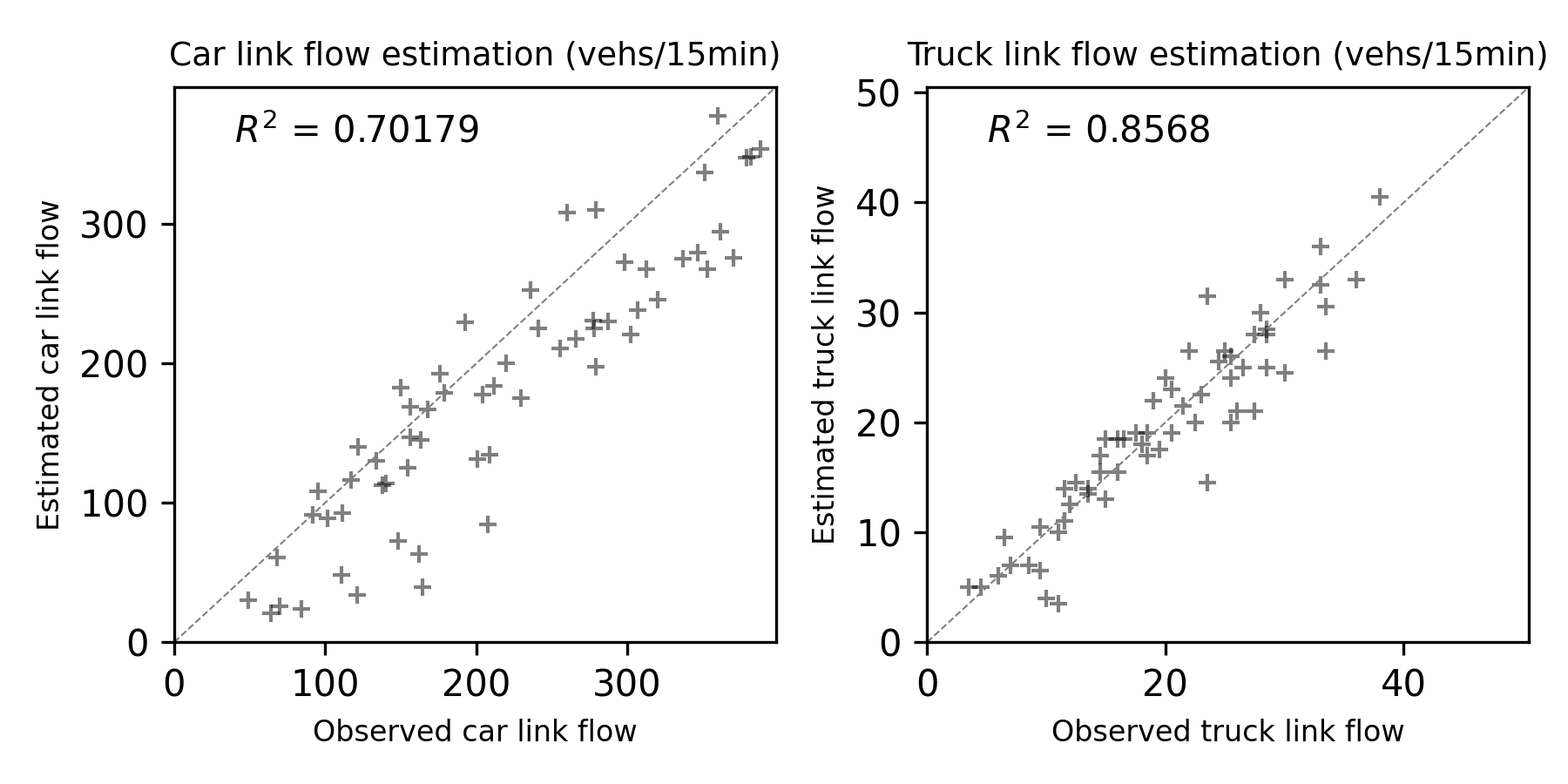}
        \caption{Scenario 2 ($\pm 20\%$ error in densities)}
    \end{subfigure}
    \begin{subfigure}{0.495\textwidth}
        \centering
        \includegraphics[width=\textwidth]{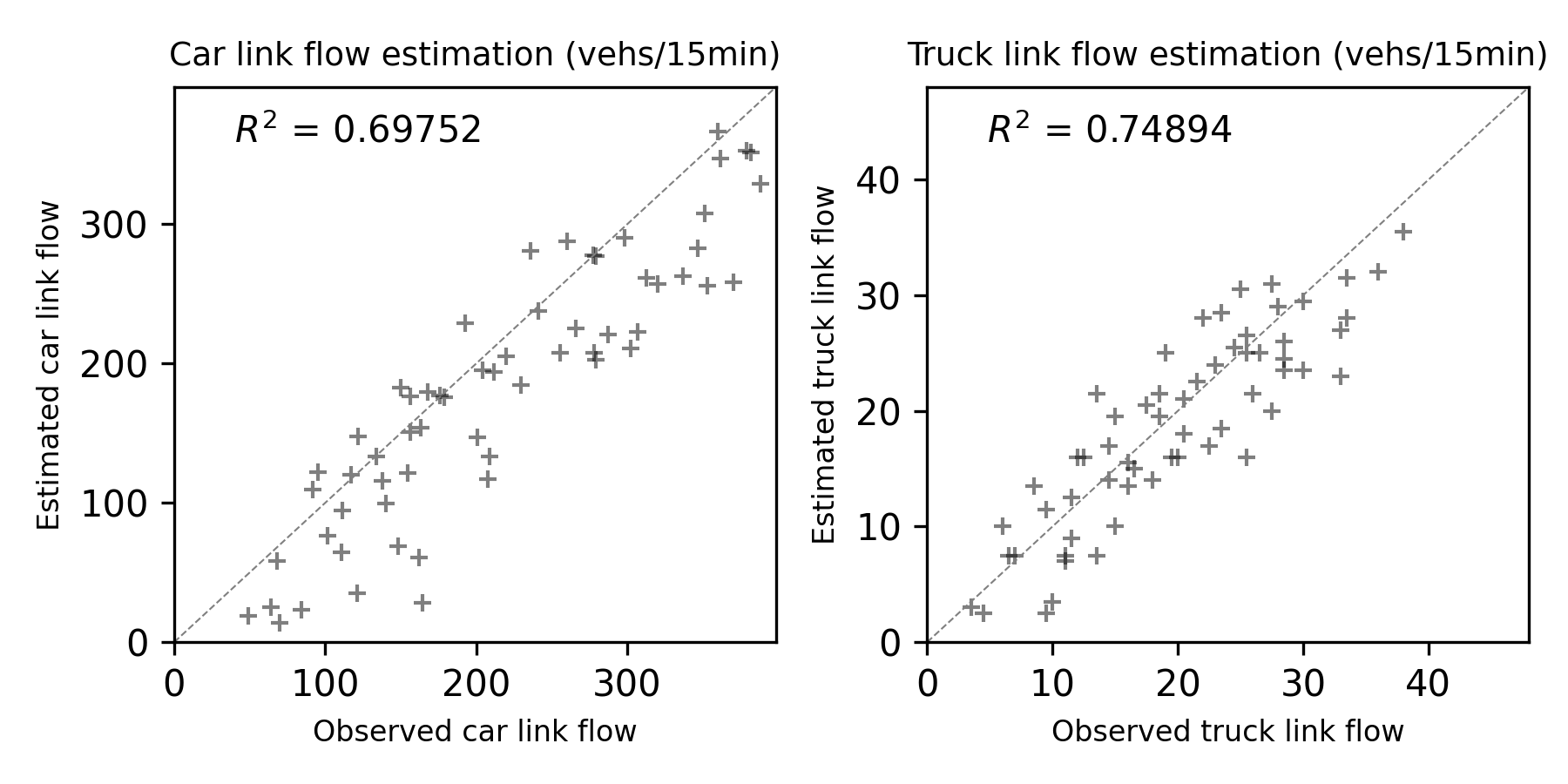}
        \caption{Scenario 2 ($\pm 50\%$ error in densities)}
    \end{subfigure}
    \caption{Comparison of count estimation for unobserved links}
    \label{fig:toy_count_oos}
\end{figure}

\begin{figure}[H]
    \centering
    \begin{subfigure}{0.495\textwidth}
        \centering
        \includegraphics[width=\linewidth]{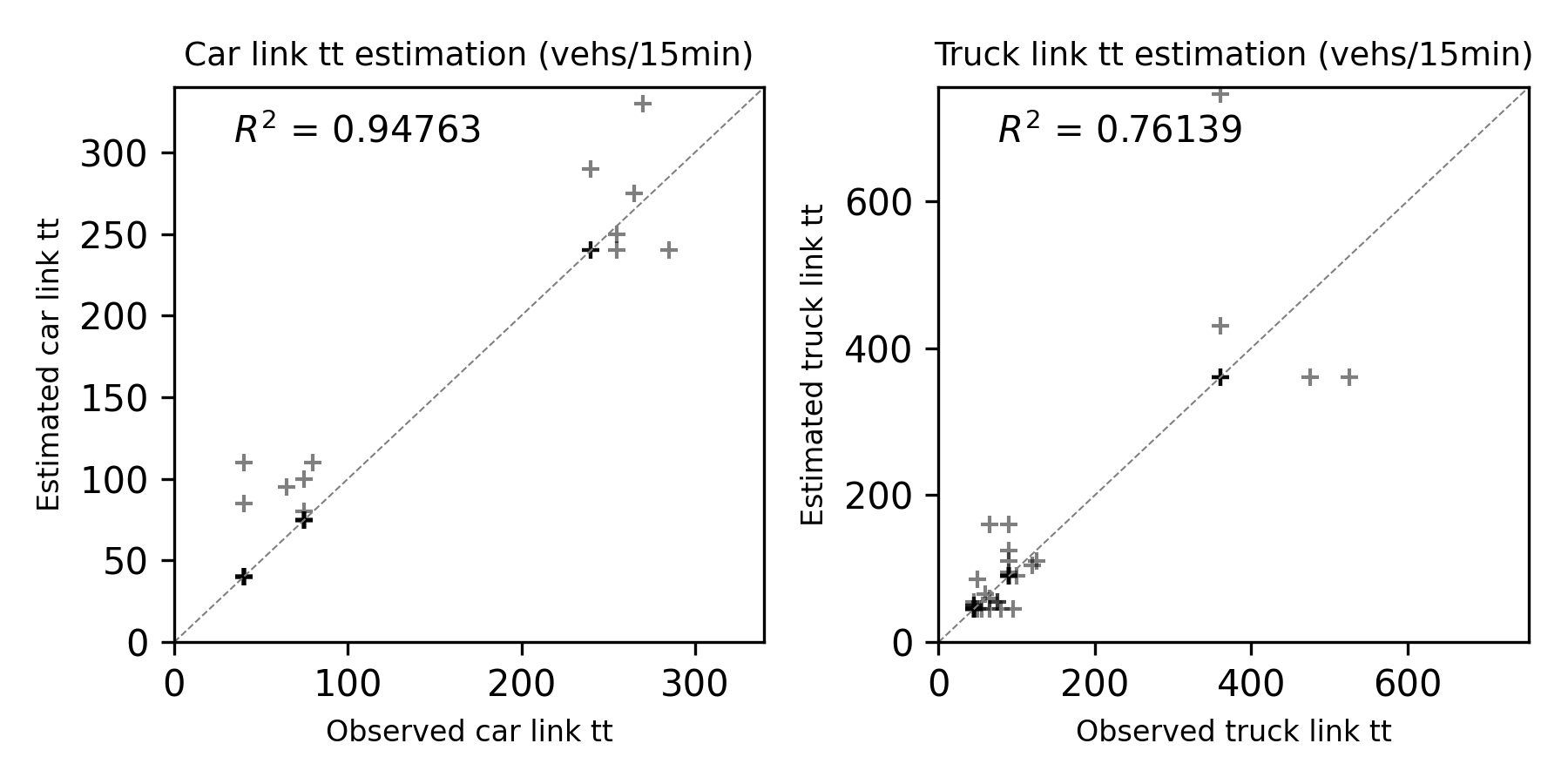}
        \caption{Scenario 1}
    \end{subfigure}
    \begin{subfigure}{0.495\textwidth}
        \centering
        \includegraphics[width=\textwidth]{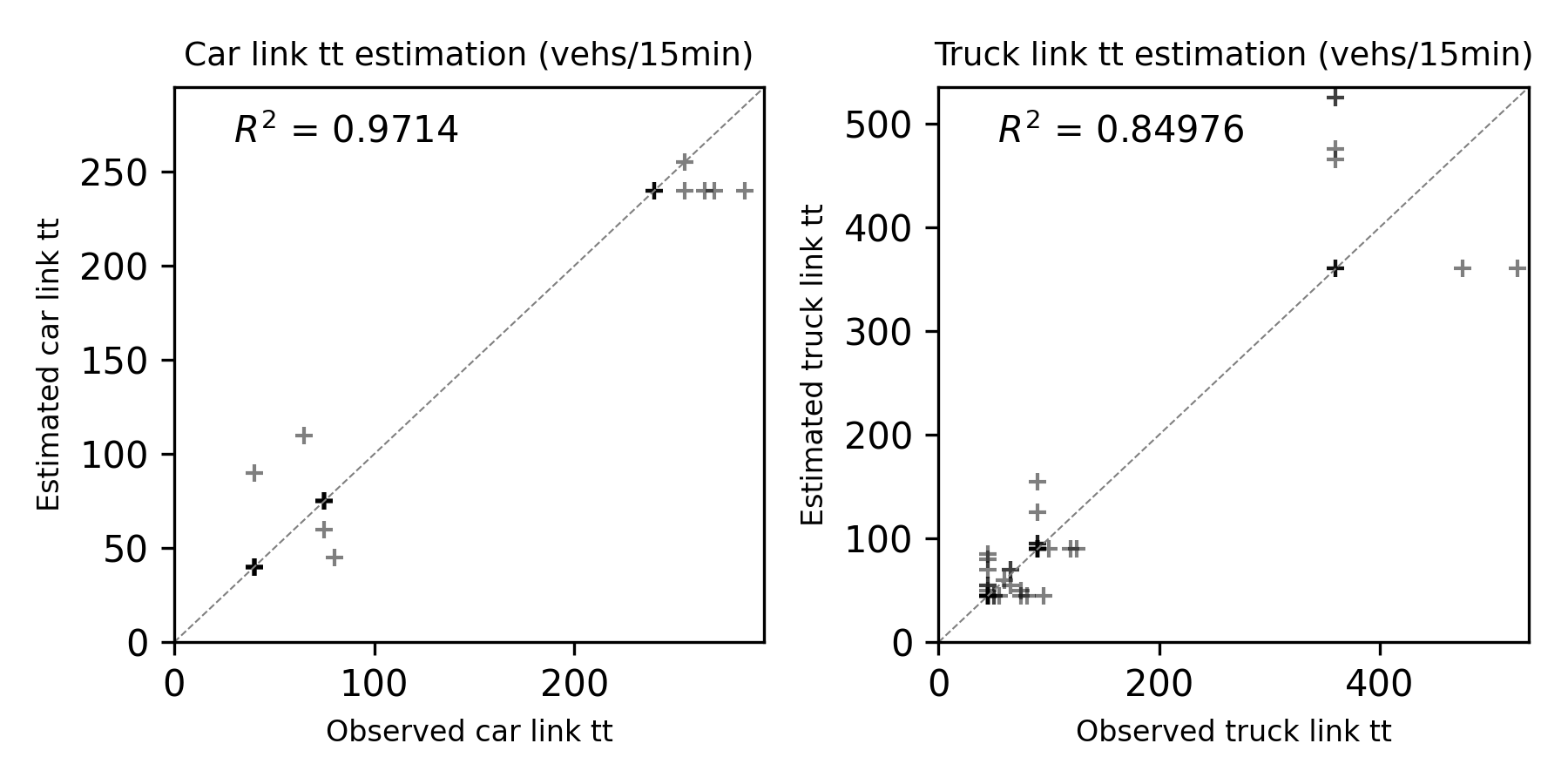}
        \caption{Scenario 2 ($\pm 10\%$ error in densities)}
    \end{subfigure}
    \begin{subfigure}{0.495\textwidth}
        \centering
        \includegraphics[width=\textwidth]{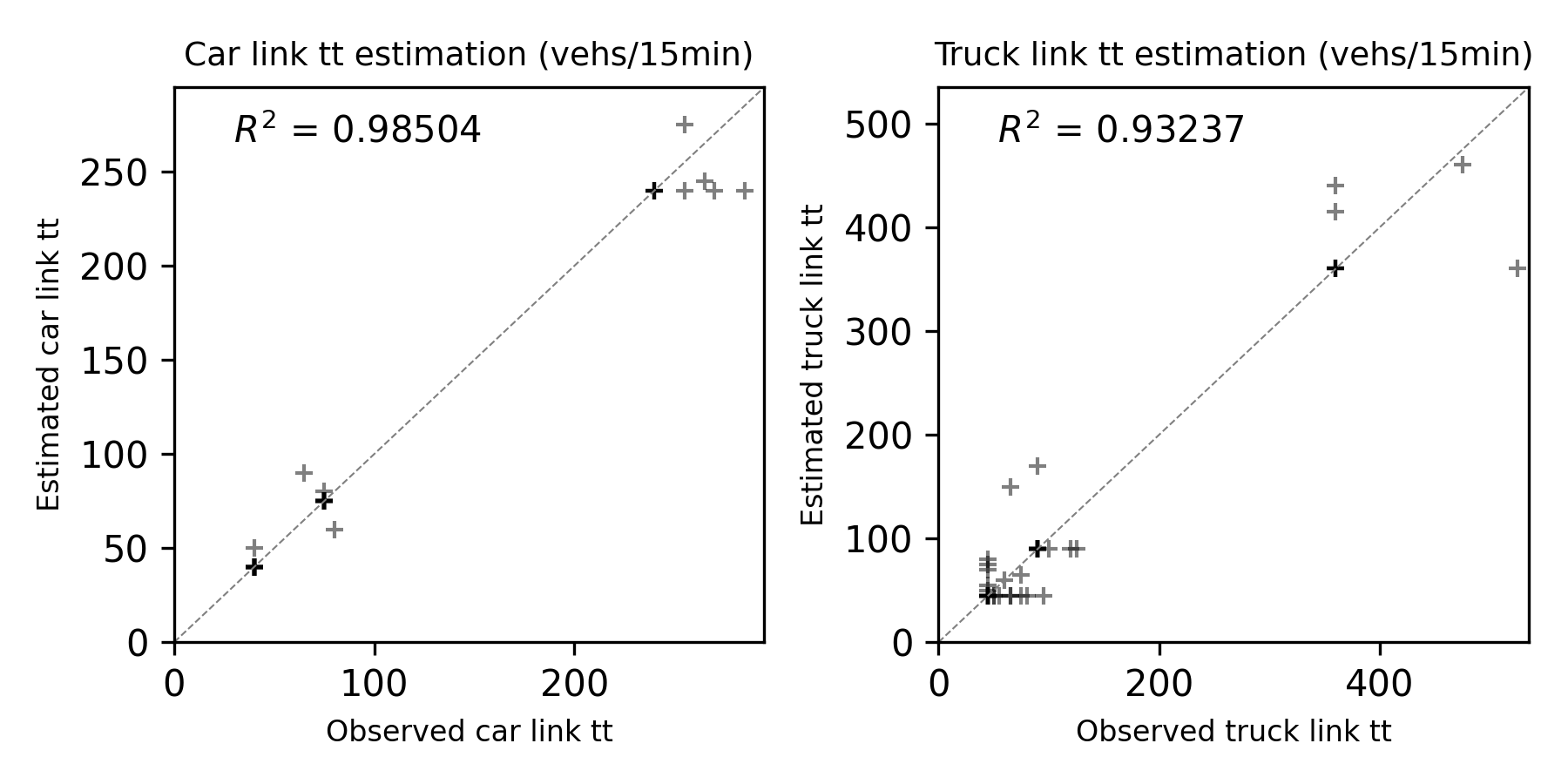}
        \caption{Scenario 2 ($\pm 20\%$ error in densities)}
    \end{subfigure}
    \begin{subfigure}{0.495\textwidth}
        \centering
        \includegraphics[width=\textwidth]{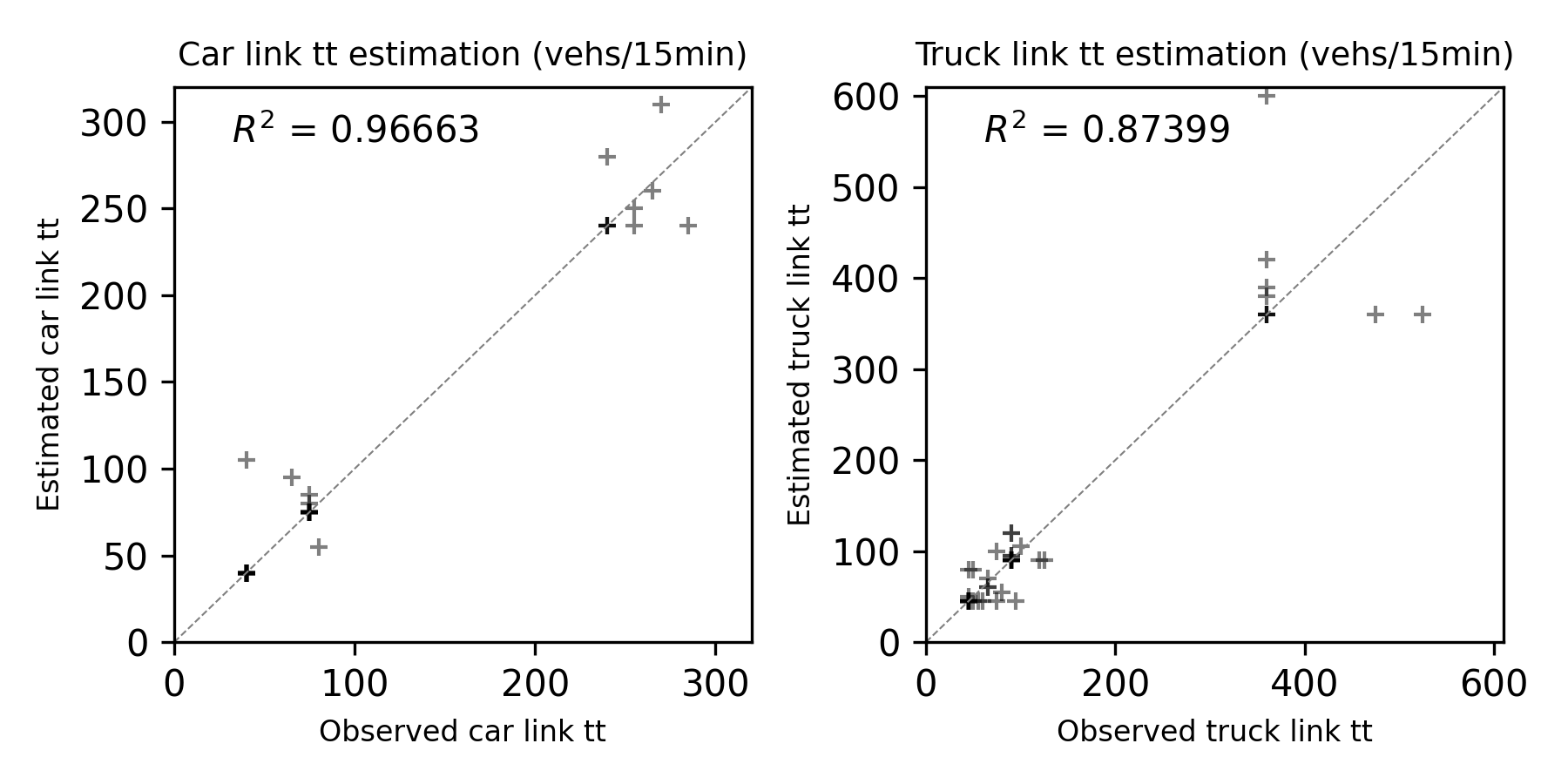}
        \caption{Scenario 2 ($\pm 50\%$ error in densities)}
    \end{subfigure}
    \caption{Comparison of travel time estimation for unobserved links}
    \label{fig:toy_tt_oos}
\end{figure}

\begin{figure}[H]
    \centering
    \begin{subfigure}{0.495\textwidth}
        \centering
        \includegraphics[width=\linewidth]{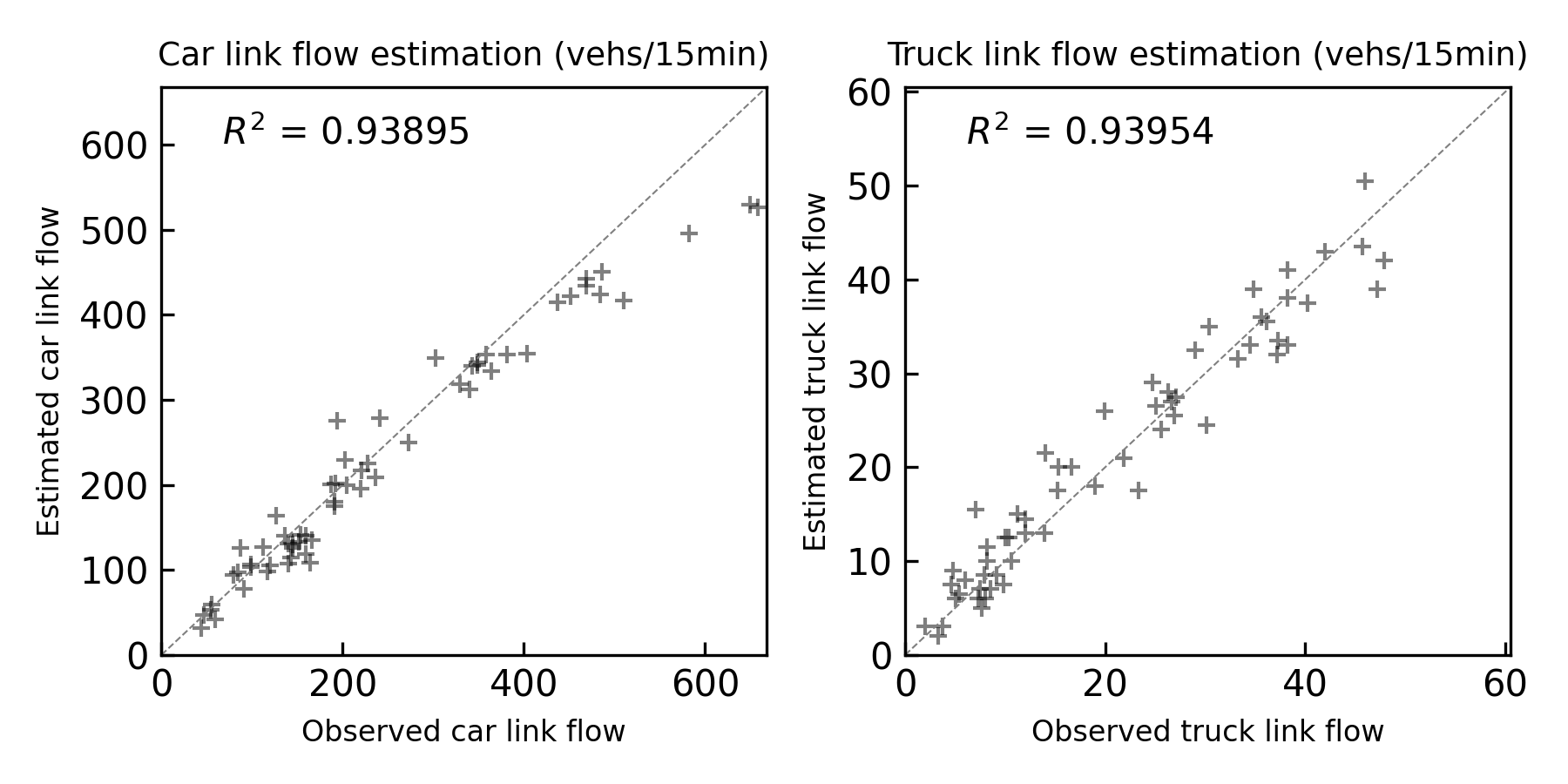}
        \caption{Scenario 1}
    \end{subfigure}
    \begin{subfigure}{0.495\textwidth}
        \centering
        \includegraphics[width=\textwidth]{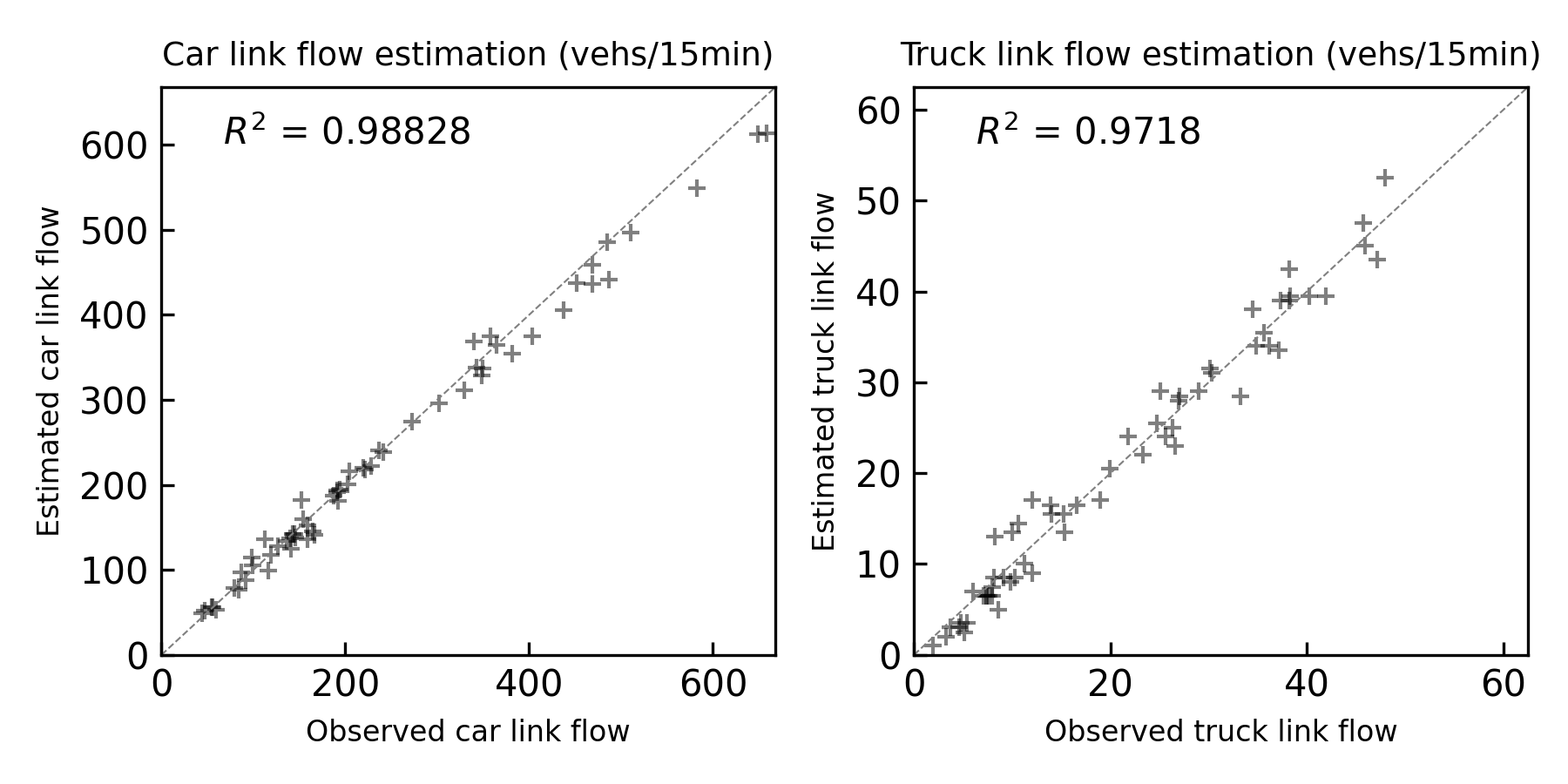}
        \caption{Scenario 2 (+/- 10\% error in densities)}
    \end{subfigure}
    \begin{subfigure}{0.495\textwidth}
        \centering
        \includegraphics[width=\textwidth]{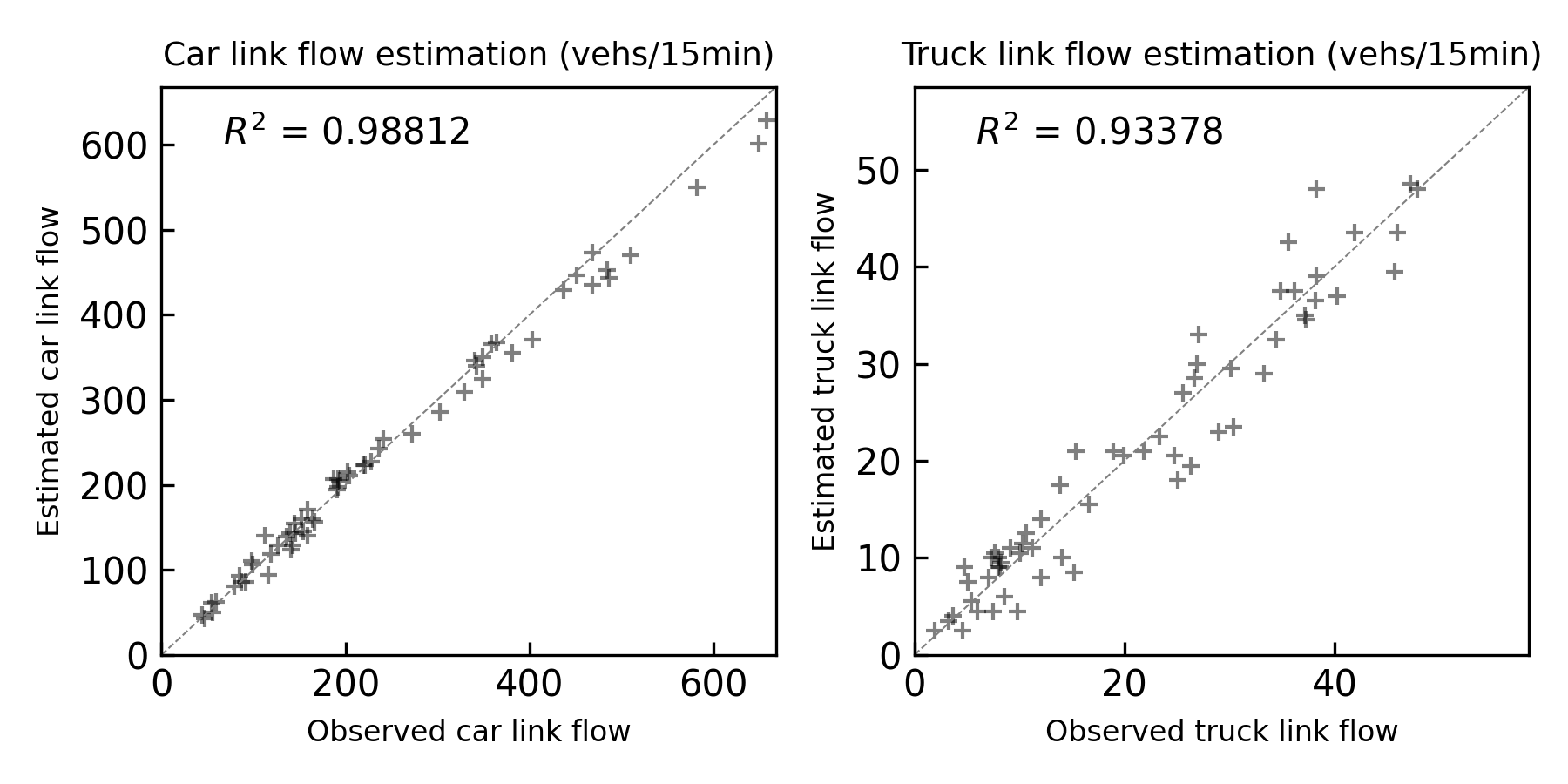}
        \caption{Scenario 2 (+/- 20\% error in densities)}
    \end{subfigure}
    \begin{subfigure}{0.495\textwidth}
        \centering
        \includegraphics[width=\textwidth]{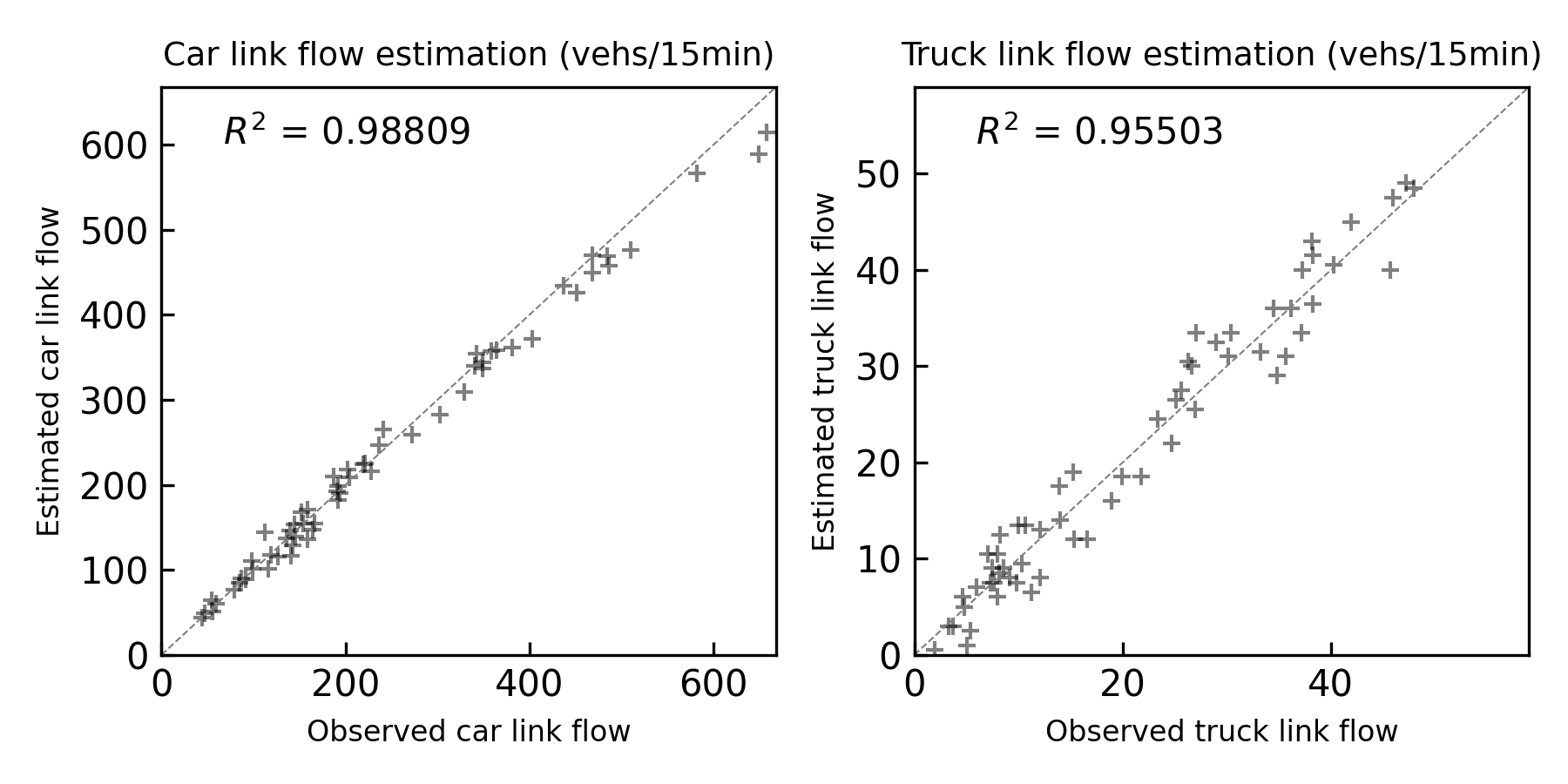}
        \caption{Scenario 2 (+/- 50\% error in densities)}
    \end{subfigure}
    \caption{Comparison of count estimation for observed links}
    \label{fig:toy_count_obs}
\end{figure}

\begin{figure}[H]
    \centering
    \begin{subfigure}{0.495\textwidth}
        \centering
        \includegraphics[width=\linewidth]{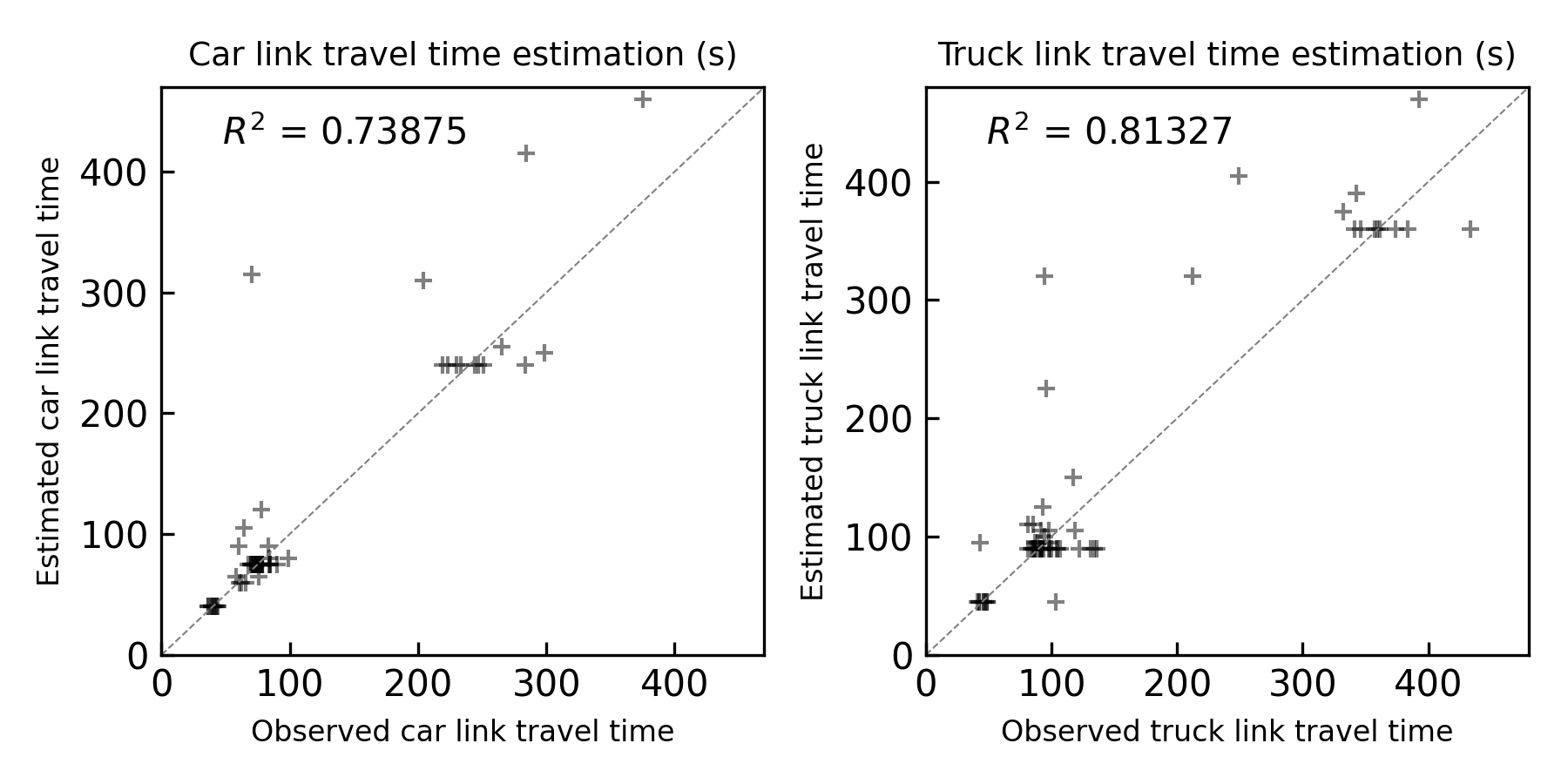}
        \caption{Scenario 1}
    \end{subfigure}
    \begin{subfigure}{0.495\textwidth}
        \centering
        \includegraphics[width=\textwidth]{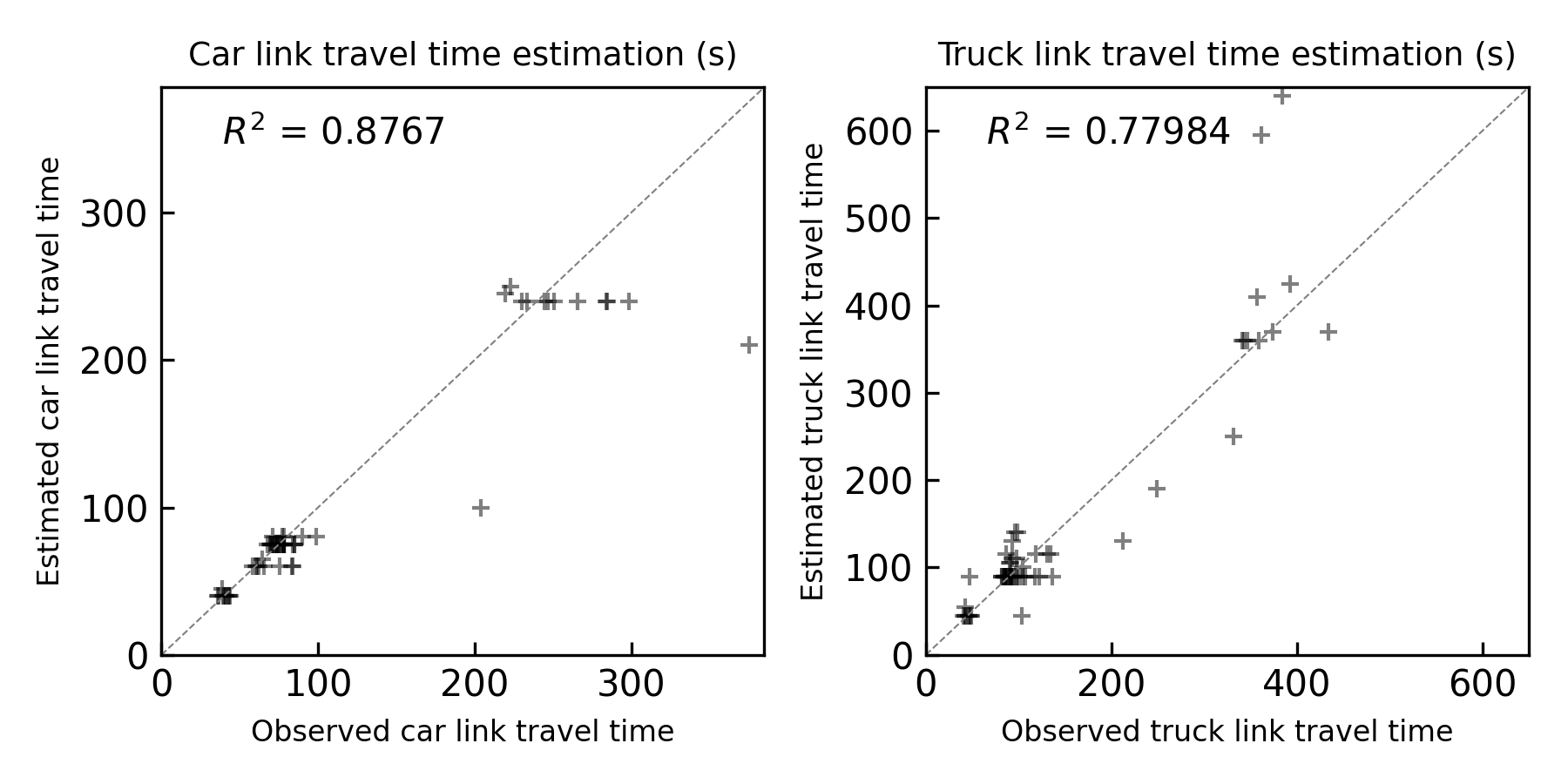}
        \caption{Scenario 2 (+/- 10\% error in densities)}
    \end{subfigure}
    \begin{subfigure}{0.495\textwidth}
        \centering
        \includegraphics[width=\textwidth]{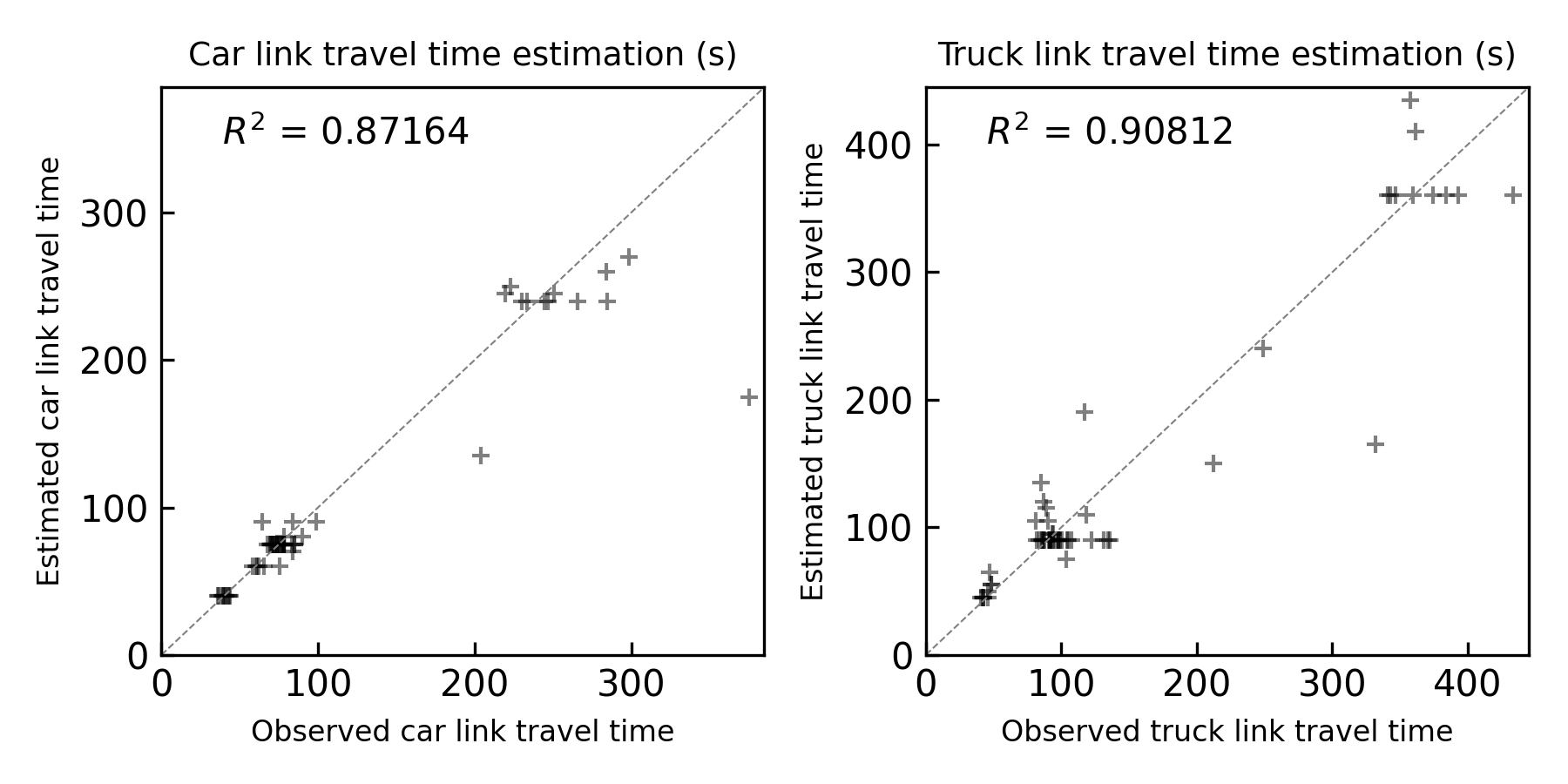}
        \caption{Scenario 2 (+/- 20\% error in densities)}
    \end{subfigure}
    \begin{subfigure}{0.495\textwidth}
        \centering
        \includegraphics[width=\textwidth]{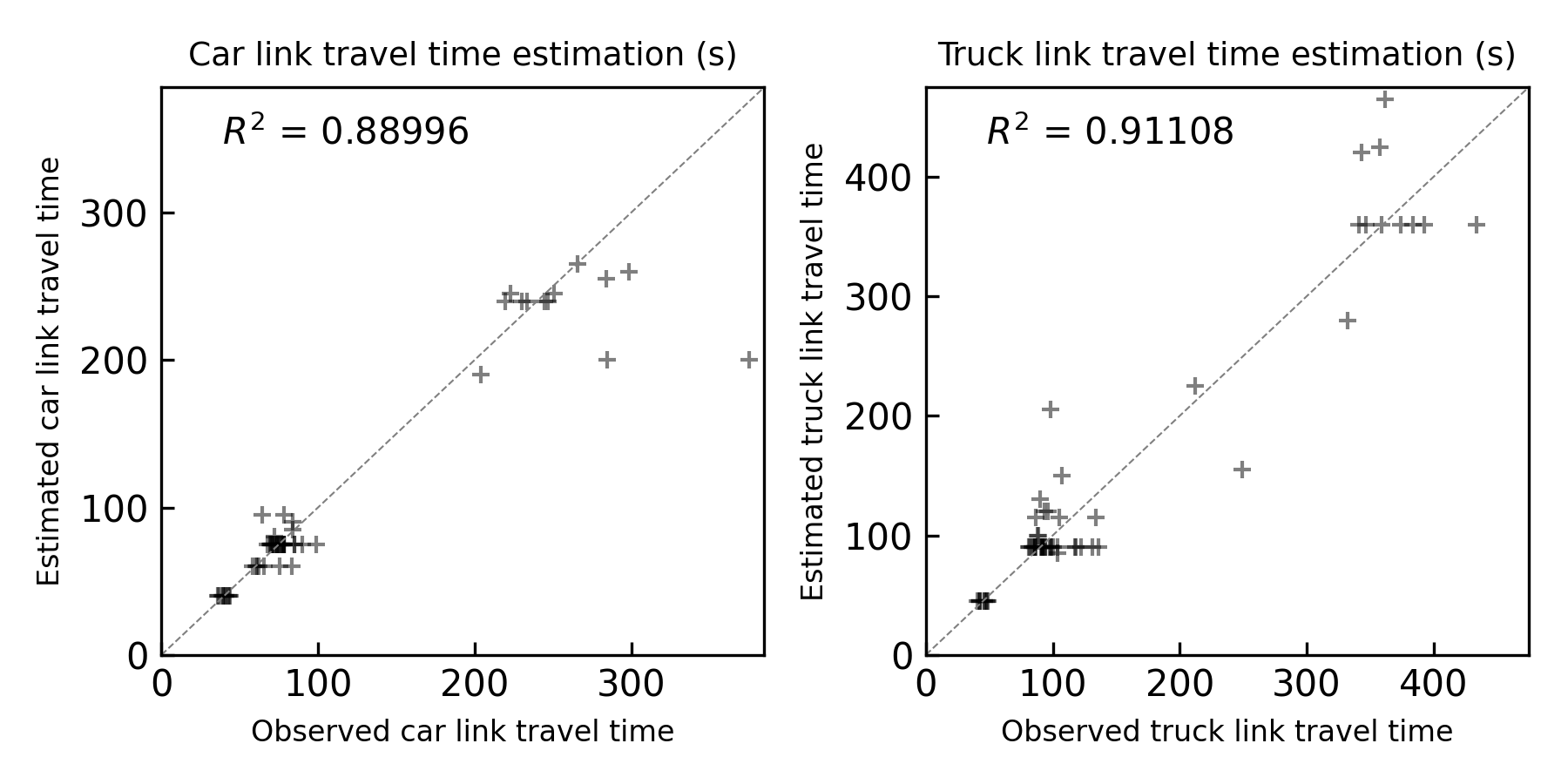}
        \caption{Scenario 2 (+/- 50\% error in densities)}
    \end{subfigure}
    \caption{Comparison of travel time estimation for observed links}
    \label{fig:toy_tt_obs}
\end{figure}

\begin{figure}[H]
    \centering
    \begin{subfigure}{0.495\textwidth}
        \centering
        \includegraphics[width=\linewidth]{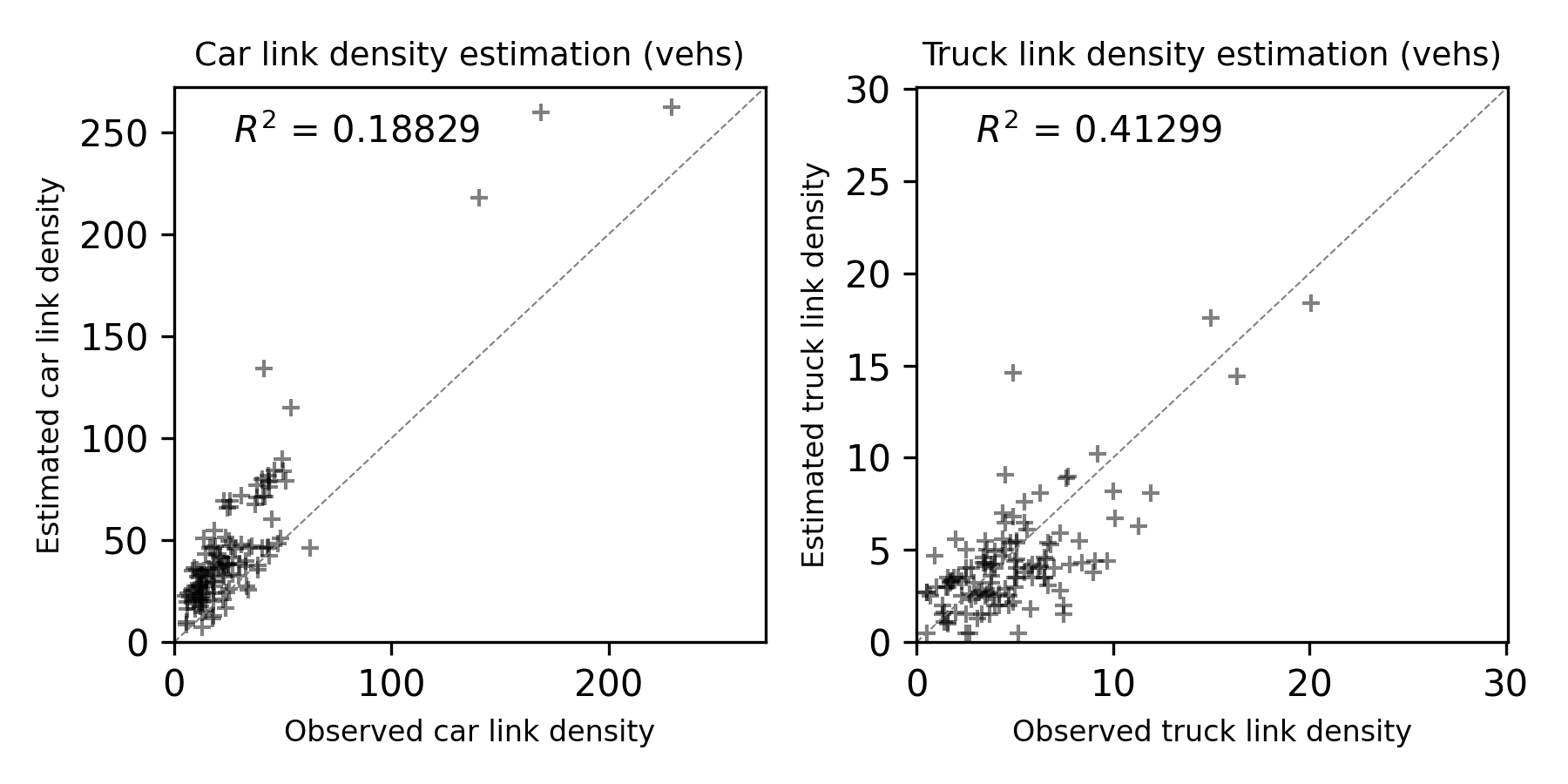}
        \caption{Scenario 1}
    \end{subfigure}
    \begin{subfigure}{0.495\textwidth}
        \centering
        \includegraphics[width=\textwidth]{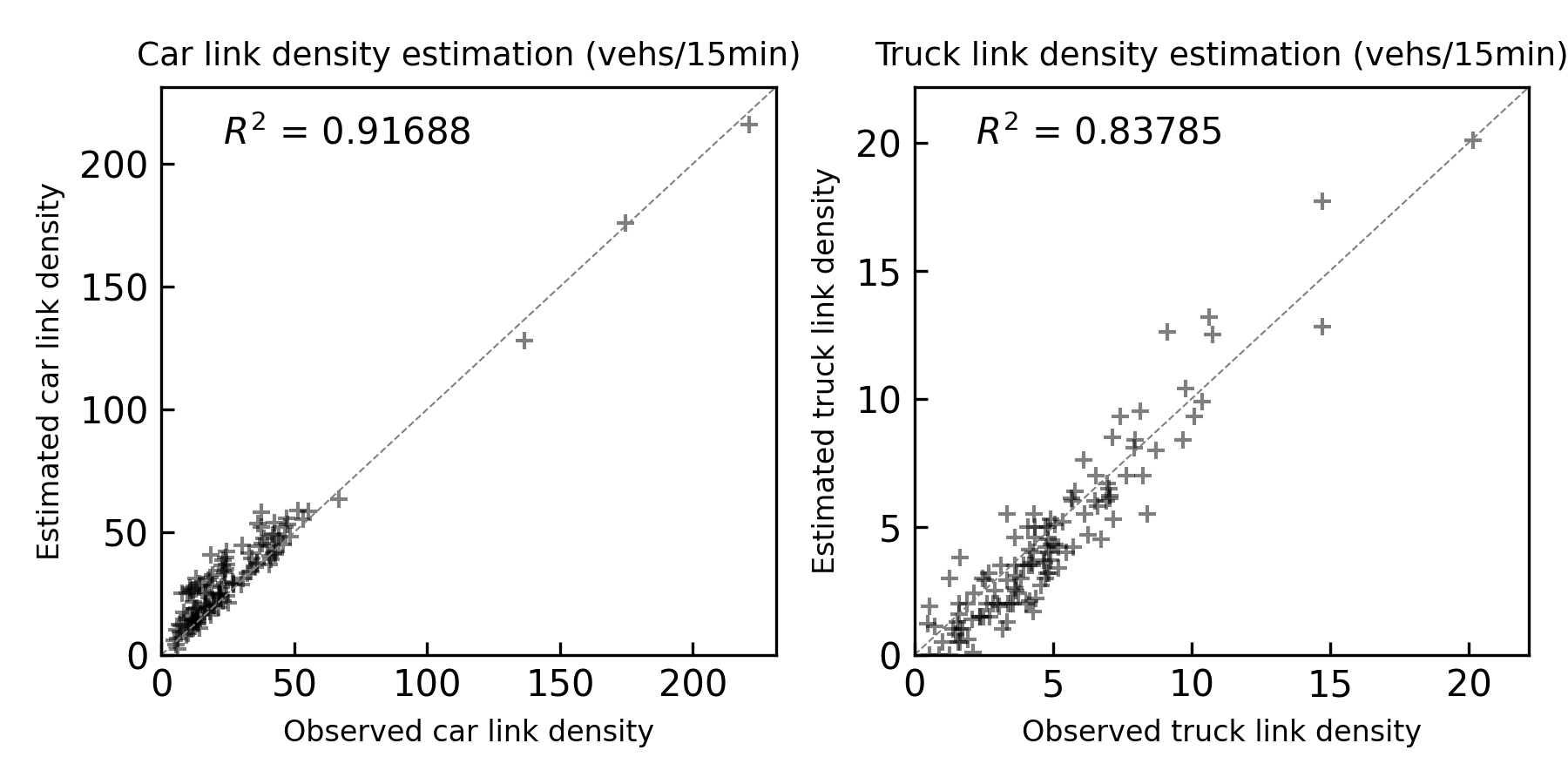}
        \caption{Scenario 2 (+/- 10\% error in densities)}
    \end{subfigure}
    \begin{subfigure}{0.495\textwidth}
        \centering
        \includegraphics[width=\textwidth]{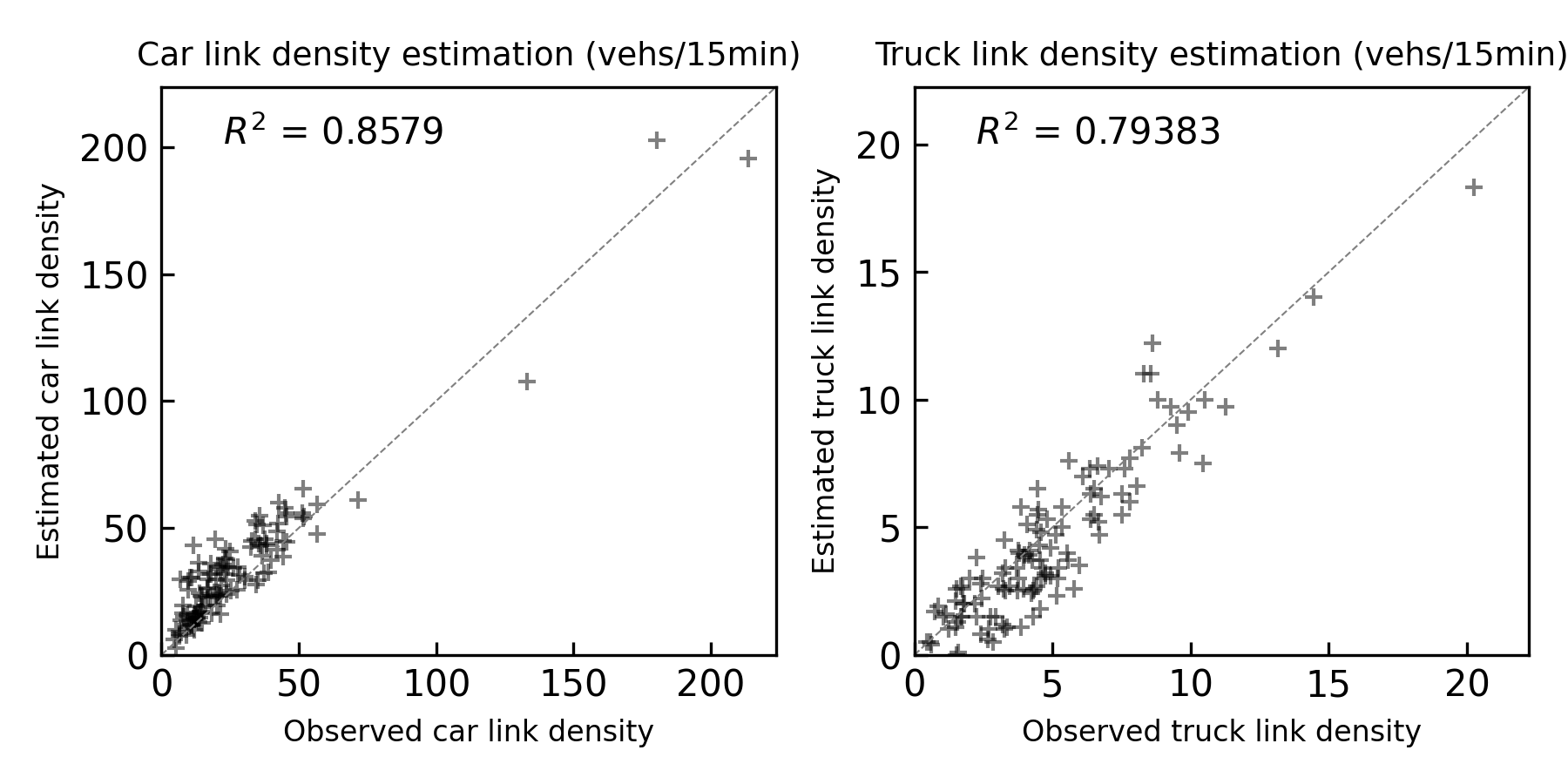}
        \caption{Scenario 2 (+/- 20\% error in densities)}
    \end{subfigure}
    \begin{subfigure}{0.495\textwidth}
        \centering
        \includegraphics[width=\textwidth]{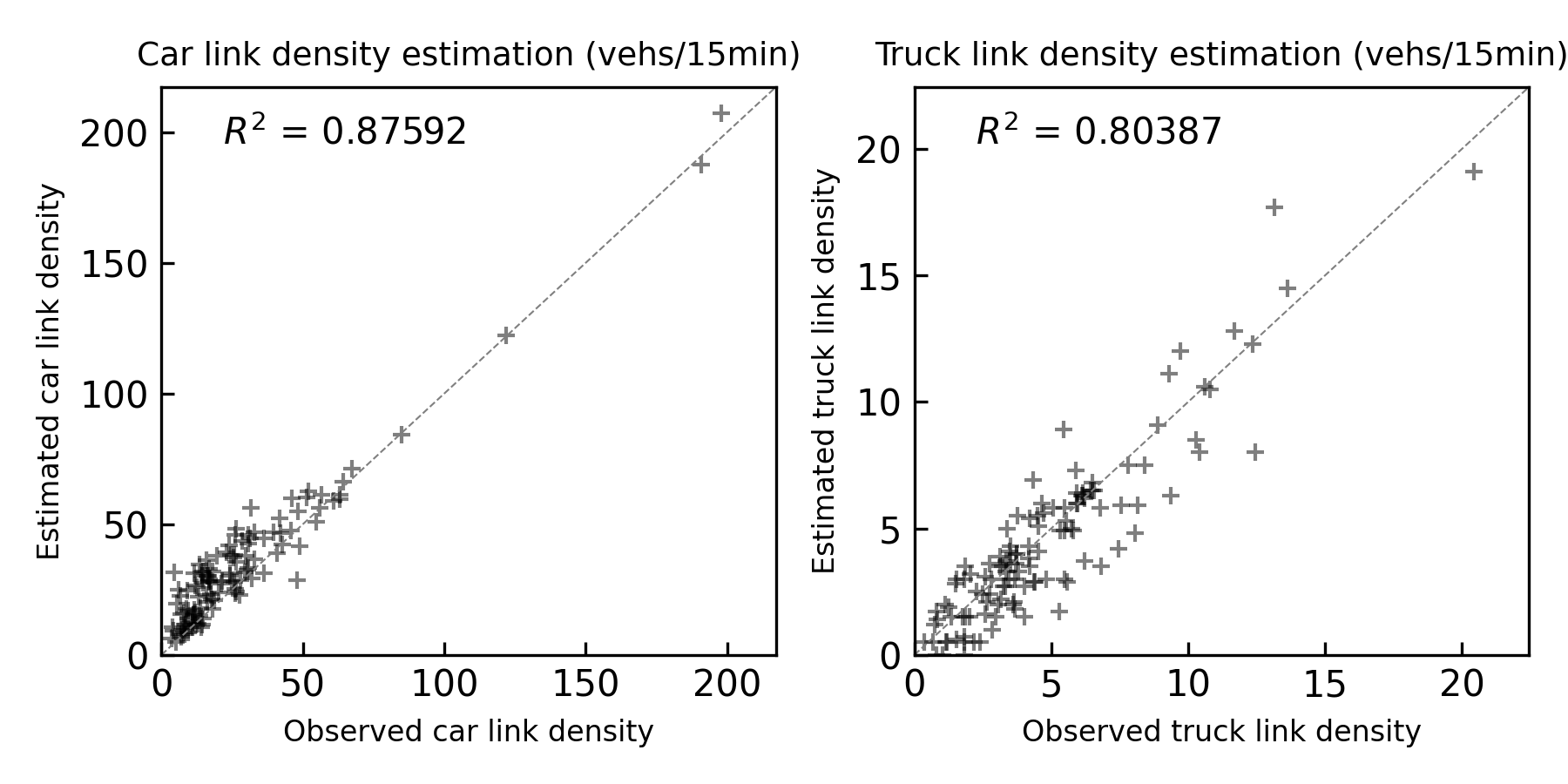}
        \caption{Scenario 2 (+/- 50\% error in densities)}
    \end{subfigure}
    \caption{Comparison of density estimation for all links}
    \label{fig:toy_k}
\end{figure}

\begin{table}[H]
    \centering
    \begin{tabular}{ccccc}
    \hline
      Vehicle class & Scenario 1 & Scenario 2 ($\pm 10\%$) & Scenario 2 ($\pm 20\%$) & Scenario 2 ($\pm 50\%$) \\
    \hline
      car  & 34.4 & 26.5 & 28.3 & 28.3\\
      truck & 2.9 & 2.4 & 2.4 & 2.7\\
    \hline
    \end{tabular}
    \caption{Mean absolute error (MAE) of multi-class demand under different scenarios}
    \label{tab:demand_mae}
\end{table}

 {The full OD ground truth is usually not available, which is typical in real-world networks. Therefore, in this study, the calibration and validation of the estimated multi-class OD demand in all experiments consistently rely on all observed traffic states (which can be either real-world data or synthetic data) that can be reproduced by the network model with estimated OD demand as inputs. Specifically, for each vehicle class, we compare modeled states (e.g., link flow, travel time, density) against observations and compute $R^2$ metrics to quantify the goodness-of-fit.}

Figure~\ref{fig:toy_count_oos}, \ref{fig:toy_tt_oos}, \ref{fig:toy_count_obs} and \ref{fig:toy_tt_obs} show the count and travel time estimation for unobserved and observed links under different scenario settings. It can be seen that augmenting the data with traffic density information for all links can improve the estimation accuracy of both unobserved and observed links, therefore improving the overall performance. However, the improvement relies on the sensing error level when extracting density information from imagery data. The improvement for unobserved links is overall better than that for observed links since density is the only data source for their estimation. Even though the test is performed on a toy network, the DODE is still underdetermined because only half of the links have traffic count and speed observations and the density information cannot differentiate parking and moving vehicles, and this information cannot fully reproduce the ground truth OD matrix. We calculate the mean squared errors for demand estimation under different scenario settings (shown in Table~\ref{tab:demand_mae}), demonstrating incorporating density can lead to estimated demand that is closer to ground truth, and that the improvement of car demand estimation is better than that for trucks due to the relatively higher car volumes in the network.

\subsection{Pittsburgh Downtown network with real-world data}
\subsubsection{Experiment Setting}
\label{sec:real_net}
\begin{figure}
    \centering
    \includegraphics[width=0.7\linewidth]{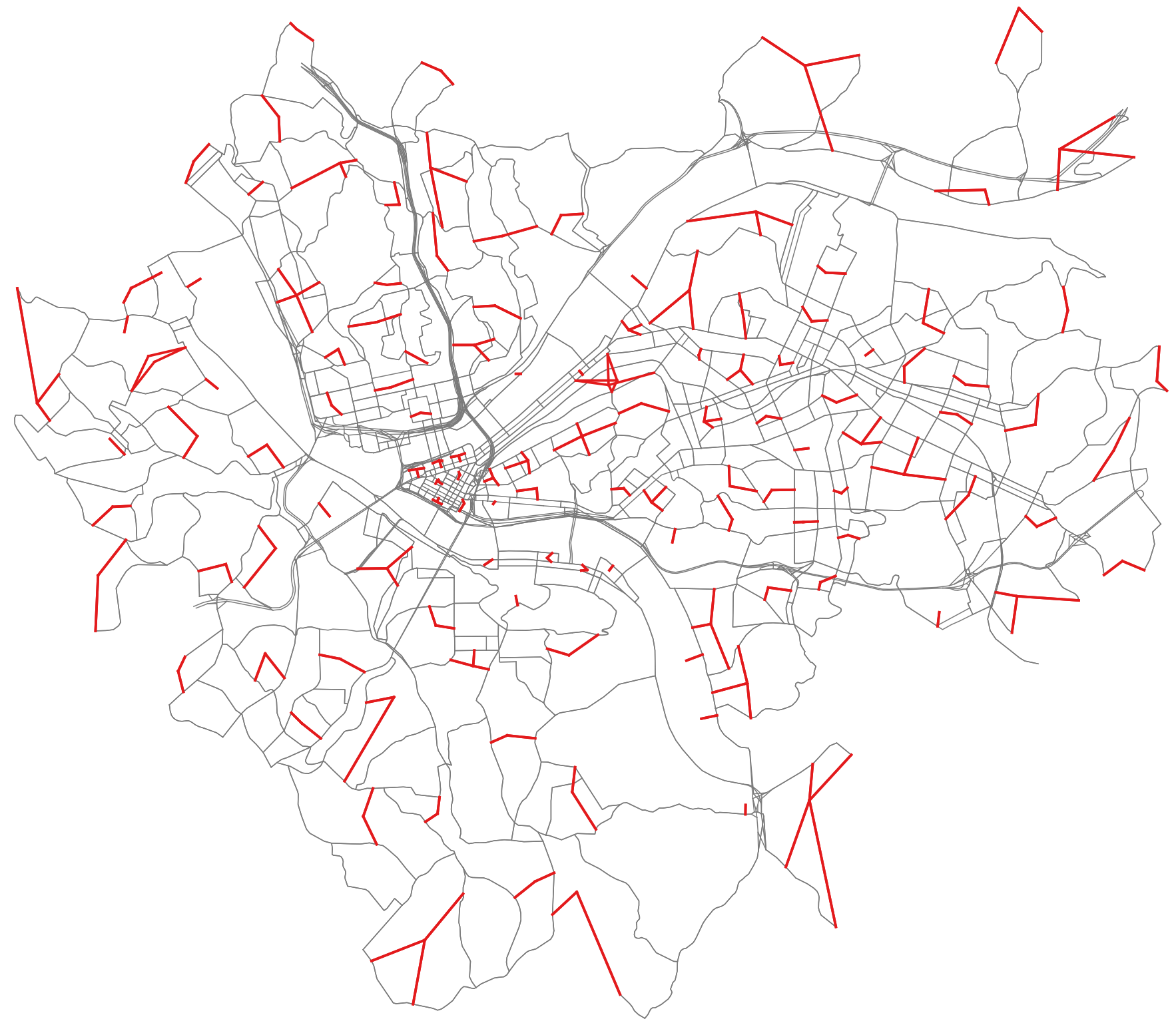}
    \caption{Pittsburgh network}
    \label{fig:pgh}
\end{figure}

The numerical experiment uses real-world and synthetic data on the Pittsburgh network to evaluate the proposed model. The network and sensing scales are shown in Figure~\ref{fig:pgh} and the detailed configuration is presented in Table~\ref{tab:pgh}. Two distinct real-world data sources are used: detector-based traffic counts and manual counts provided by the Southwestern Pennsylvania Commission, and two high-resolution satellite images provided by Airbus. The proposed computer vision pipeline is applied to extract link densities from the satellite images, distinguishing between vehicle classes but not parking and on-road vehicles. Therefore, the link density observations represent total vehicles of both parking and through traffic.

 {Due to data availability, two satellite images months apart are used in this real-world experiment. Our experiment therefore serves as a proof-of-concept demonstration under limited data availability, to show the benefit of incorporating satellite-derived, network-wide, class-specific density information into the DODE process. When more frequent and temporally aligned imagery data become available, more experiments can be conducted to support large-scale deployment in real world applications.}

\subsubsection{Traffic density observations from satellite imagery}
\begin{figure}[H]
    \centering
    \includegraphics[width=\linewidth, trim= {2cm 0.5cm 3.5cm 1cm}, clip]{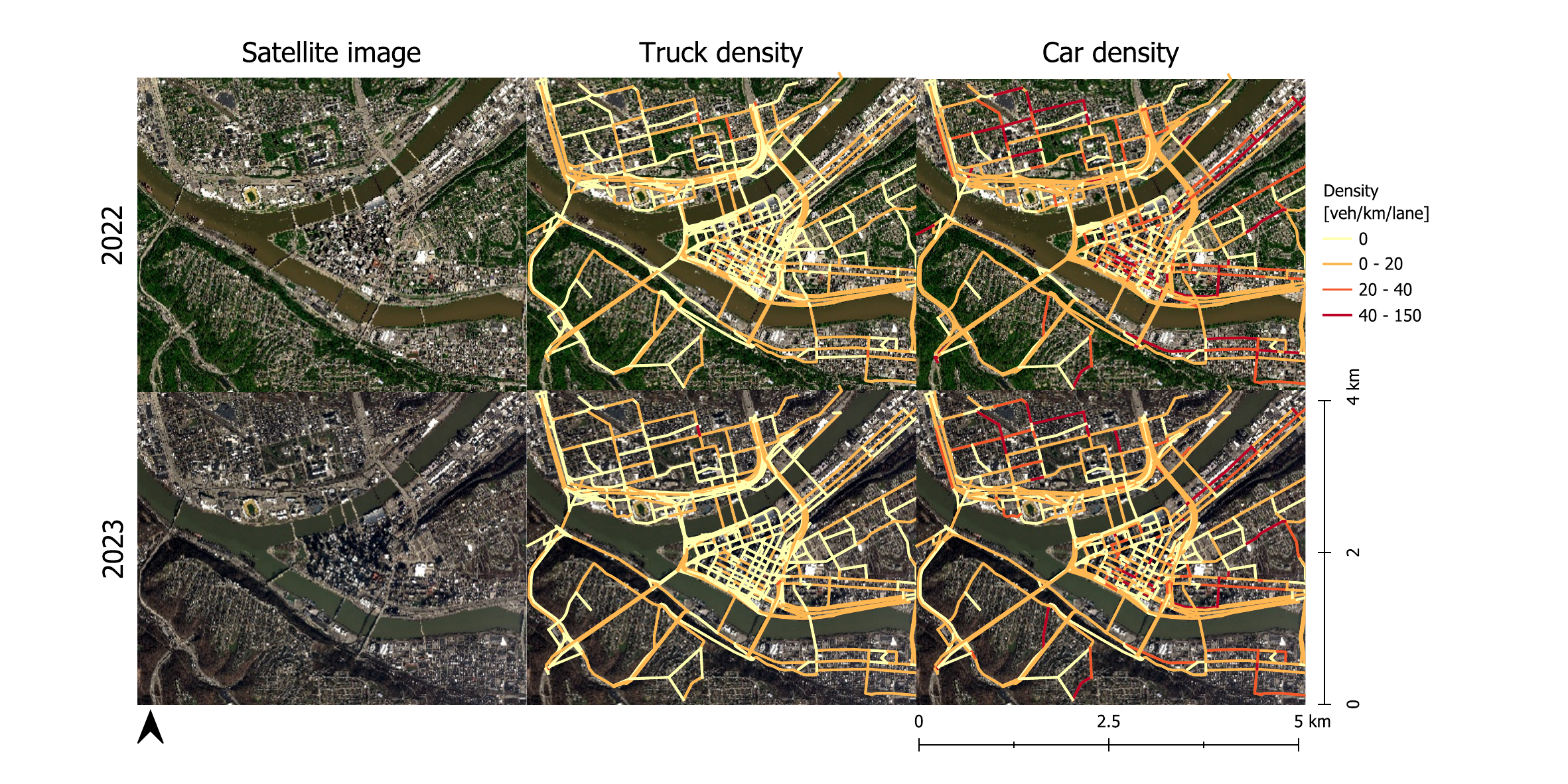}
    \caption{Link-level traffic density derived from satellite imagery, disaggregated by year (2022, 2023) and vehicle class (truck, car)\phantom{.........}}
    \label{fig:satellite-image-traffic-density}
\end{figure}

Two 30-cm resolution satellite images captured by a Pleiades Neo satellite from Airbus were used as the input for the computer vision pipeline. The images were collected on Wednesday, May 11, 2022 at 12:18 PM and Monday, March 6, 2023 at 12:17 PM Eastern Time. As shown in the first column of images in Figure \ref{fig:satellite-image-traffic-density}, each image has a width of 5 kilometers and a height of approximately 4 kilometers, spanning around 20 square kilometers and covering the downtown area in Pittsburgh, PA. The original images included four channels, but only the RGB channels were extracted to run the computer vision pipeline. 

To improve object detection results with Airbus imagery, the object detector was also trained with 30-cm resolution satellite imagery from the xView dataset \citep{lam_xview_2018}. Using satellite imagery with the same resolution for training and inference has been shown to improve object detection results \citep{van_etten_you_2018}. To ensure adequate representation of the positive classes, only the top 10 percent of images with the highest number of annotated vehicles were selected. This subset was then divided into training and validation sets, consisting of 46 images (80\%) and 13 images (20\%), respectively.

Table \ref{table:performance-metrics-object-detection} presents the precision, recall, and F1-score on the training set, disaggregated by vehicle type. The recall and precision for cars are comparable to those reported by \citet{liu2024sat}, which used a binary classifier for car detection. The lower precision observed for trucks relative to cars aligns with the trend in mean average precision reported by \citet{lam_xview_2018} for truck-related classes in the xView dataset. The last five columns of the table report the observed counts, predicted counts, true positives, false positives, and false negatives for each vehicle type, where the predicted count equals the sum of true and false positives, and the observed count equals the sum of true positives and false negatives.

\begin{table}[H]
\begin{adjustbox}{width=\linewidth} 
\renewcommand{\arraystretch}{1.05}
\centering
\begin{threeparttable}
\caption{Performance metrics for object detection in validation set}
\label{table:performance-metrics-object-detection}
\begin{tabular}{ccccccccc}
\hline
\multirow{2}{*}{\begin{tabular}[c]{@{}c@{}} \\ Vehicle \\ type \end{tabular}} 
& \multicolumn{8}{c}{Metric} \\ \cline{2-9} 
& \begin{tabular}[c]{@{}c@{}}Precision\end{tabular} 
& \begin{tabular}[c]{@{}c@{}}Recall\end{tabular} 
& \begin{tabular}[c]{@{}c@{}}F1 \\ score\end{tabular} 
& \begin{tabular}[c]{@{}c@{}}Observed \\ count\end{tabular} 
& \begin{tabular}[c]{@{}c@{}}Predicted \\ count\end{tabular} 
& \begin{tabular}[c]{@{}c@{}}True \\ positives\end{tabular} 
& \begin{tabular}[c]{@{}c@{}}False \\ positives\end{tabular} 
& \begin{tabular}[c]{@{}c@{}}False \\ negatives\end{tabular} 
\\ \hline
Cars
& $0.62$
& $0.66$
& $0.64$
& $15,562$
& $16,496$
& $10,256$
& $6,240$
& $5,306$
\\
Trucks
& $0.33$
& $0.31$
& $0.32$
& $1,177$
& $1,127$
& $369$
& $758$
& $808$
\\
Cars and trucks
& $0.60$
& $0.63$
& $0.62$
& $16,739$
& $17,623$
& $10,625$
& $6,998$
& $6,114$
\\
\hline
\end{tabular}
\end{threeparttable}
\end{adjustbox}
\end{table}

\begin{figure}[H]
    \centering
    \includegraphics[width=\linewidth, trim= {2cm 2.5cm 3.5cm 3cm}, clip]{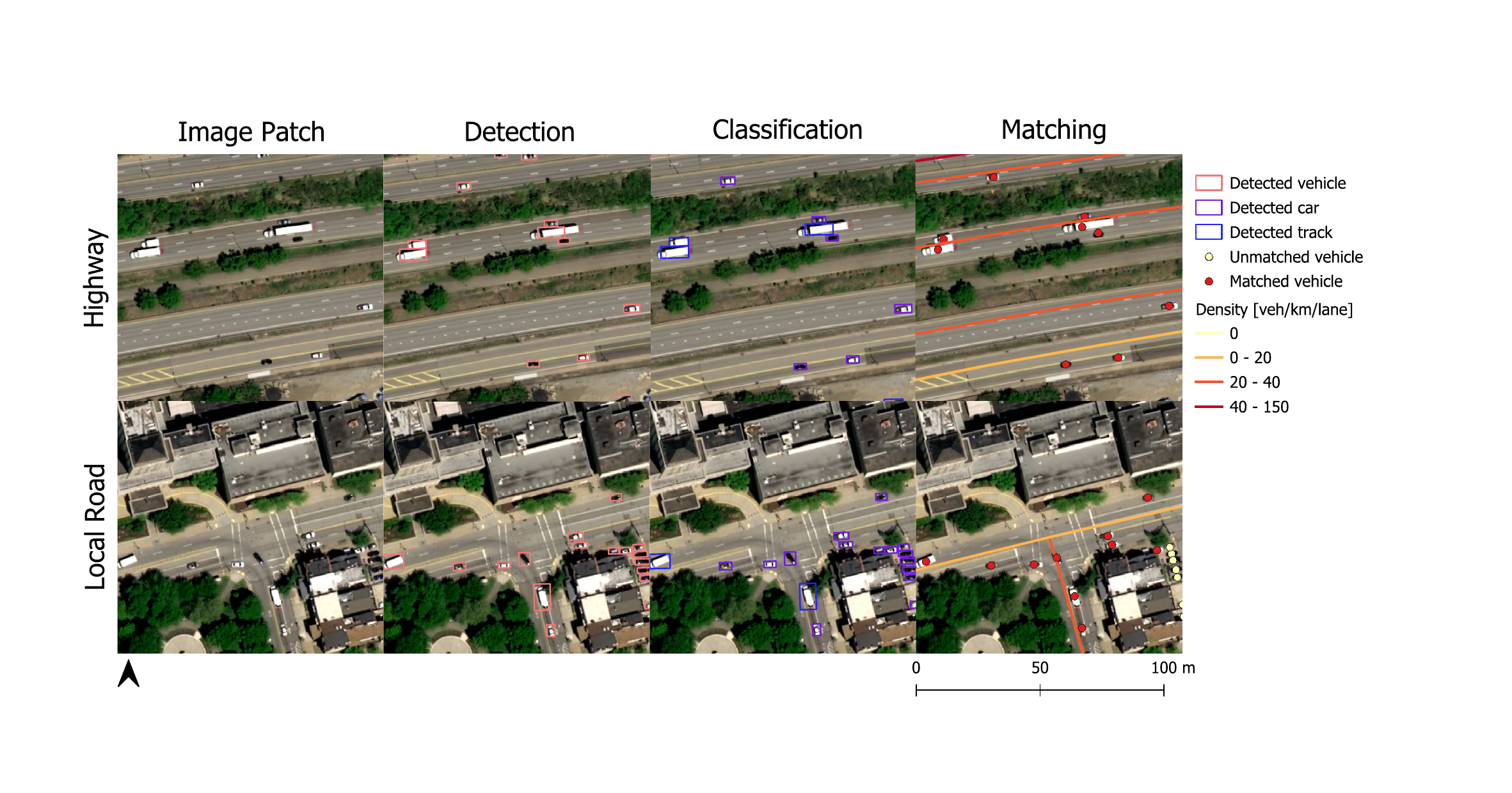}
    \caption{Vehicle detection, classification, and matching results using satellite image collected on May 11, 2022. \phantom{............................}}
    \label{fig:detection-matching-cases}
\end{figure}

Figure \ref{fig:detection-matching-cases} shows vehicle detection, classification, and matching results for image patches with a height and width of 100 meters. To analyze results across different types of streets, image patches were collected in areas containing local roads and highways. The yellow bounding boxes in the second column show the vehicle detection output. The third column shows the classification results, with purple and blue boxes representing cars and trucks, respectively. In the fourth column, the red points are the vehicles matched to the street network, while the yellow points represent vehicles that were detected but not matched to the street network. To exclude vehicles far from the road segments in the network, only vehicles near the road centerline are matched to the street network. Therefore, as shown in the image of the local road in the bottom right corner, vehicles parked near buildings or houses are not matched to the street network.

Figure \ref{fig:satellite-image-traffic-density} shows the traffic density at each link of the street network, disaggregated by trucks and cars, with the left column showing the satellite images collected for each year. Links with the darkest red color represent traffic densities higher than 40 vehicles per lane-kilometer and are associated with high congestion levels. The traffic density patterns for cars and trucks are consistent across images, which is expected given that both images were collected during the same hour of the day and on weekdays. In addition, the estimated traffic density is higher for cars than for trucks, which is in line with the share of cars and trucks observed in real-world transportation networks.

\begin{figure}[H]
    \centering
    \includegraphics[width=0.5\linewidth]{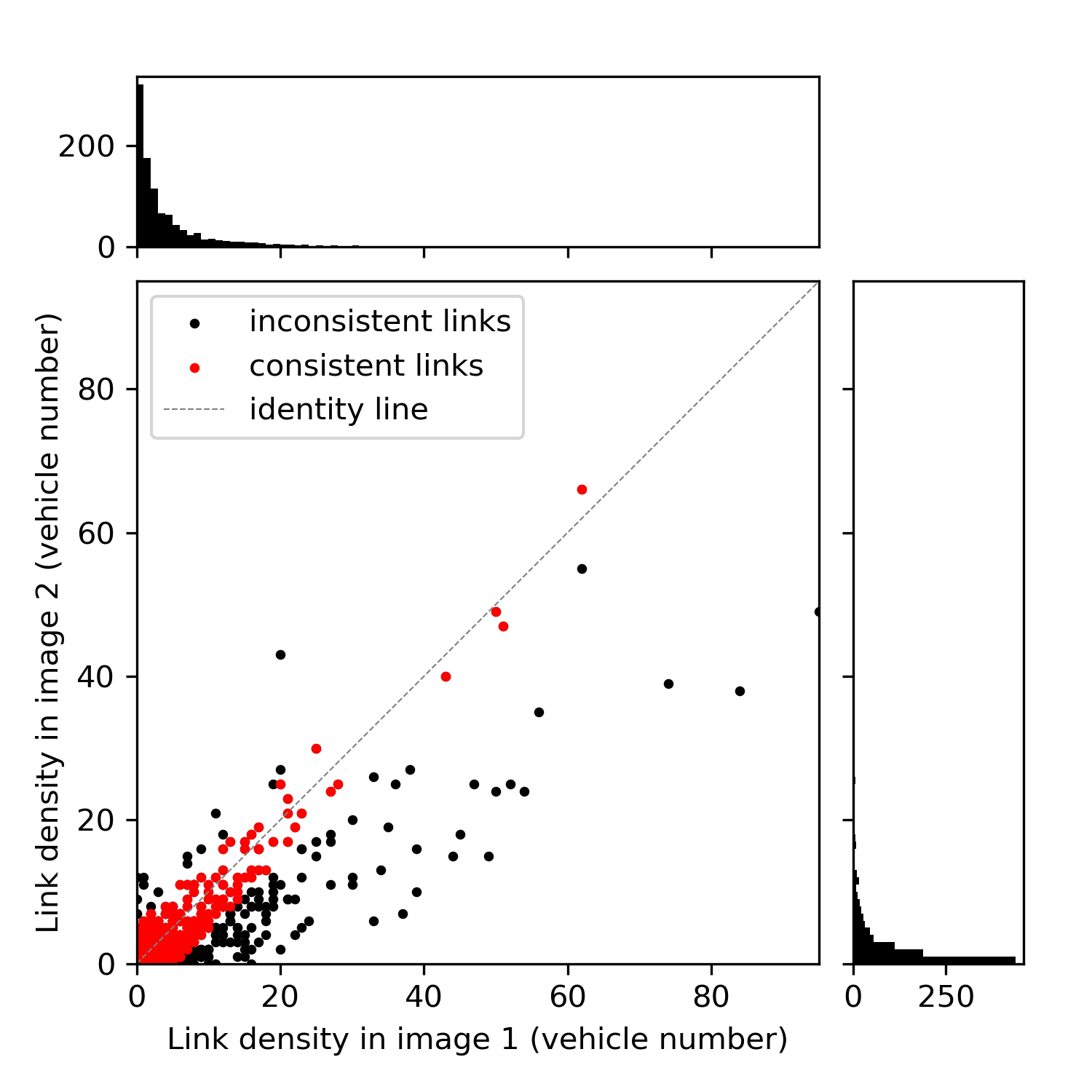}
    \caption{Density comparison in two images}
    \label{fig:density}
\end{figure}

After computer vision processing, two class-specific density snapshots are obtained within the same time interval (12:15-12:30 PM) on different dates. Figure~\ref{fig:density} illustrates the differences between the two samples. To mitigate the inconsistency, a threshold-based approach is adopted, where links with a difference of at most 5 are selected as consistent links, and their densities are used in the experiments. As a result, 900 out of 1044 links are considered consistent, shown as red dots in Figure~\ref{fig:density}.

To address the inconsistency introduced by multi-source data (e.g., different data providers and different sensing errors in data collection), we propose a two-stage calibration process using two satellite images separately. The first stage calibration is performed using one density snapshot and traffic count data, and we select a subset of links based on the following criteria: for links with non-zero density observations, we select links with estimated density no more than three times the observed value for both vehicle classes; for links with zero density observations, we select links with estimated densities no more than 10. As a result, 457 out of 900 links are selected for the second stage. In the second stage, we use the second densities of the selected link subset to perform a separate calibration together with the same traffic count data. This two-stage calibration process helps mitigate the inconsistencies between data sources and leads to more reliable results.

 {In practice, satellite imagery-based detection may be affected by different scenes such as network topology, traffic conditions, weather, lighting, occlusions etc. In future study, the CV pipeline can be further enhanced considering these complex conditions, and alternative or improved CV models can be substituted directly without changing the DODE formulation.}

\subsubsection{Results and Discussions}
To evaluate the performance of adding additional density data in DODE, we design two DODE scenarios similar to the toy network, but this time we only have information of observed links and the ground truth of unobserved links remains hidden. Scenario 1 uses only localized traffic counts for observed links, and scenario 2 uses both local sensors and a network-wide density snapshot. As DODE is a widely acknowledged underdetermined problem with unknown true OD demand and traffic states of unobserved links, we propose the following conditions to be satisfied, justifying the claim that incorporating satellite imagery can enhance DODE performance: (1) adding density into DODE cannot do harm to other data sources, meaning that the accuracy of other traffic condition estimation cannot be largely reduced. (2) the accuracy of density estimation should be improved for observed links, i.e., larger R-squared values for links with density observations. (3) the estimated OD demand pattern in scenario 2 is different from the one in scenario 1.

\begin{figure}[H]
    \centering
    \begin{subfigure}{0.495\textwidth}
        \centering
        \includegraphics[width=\linewidth]{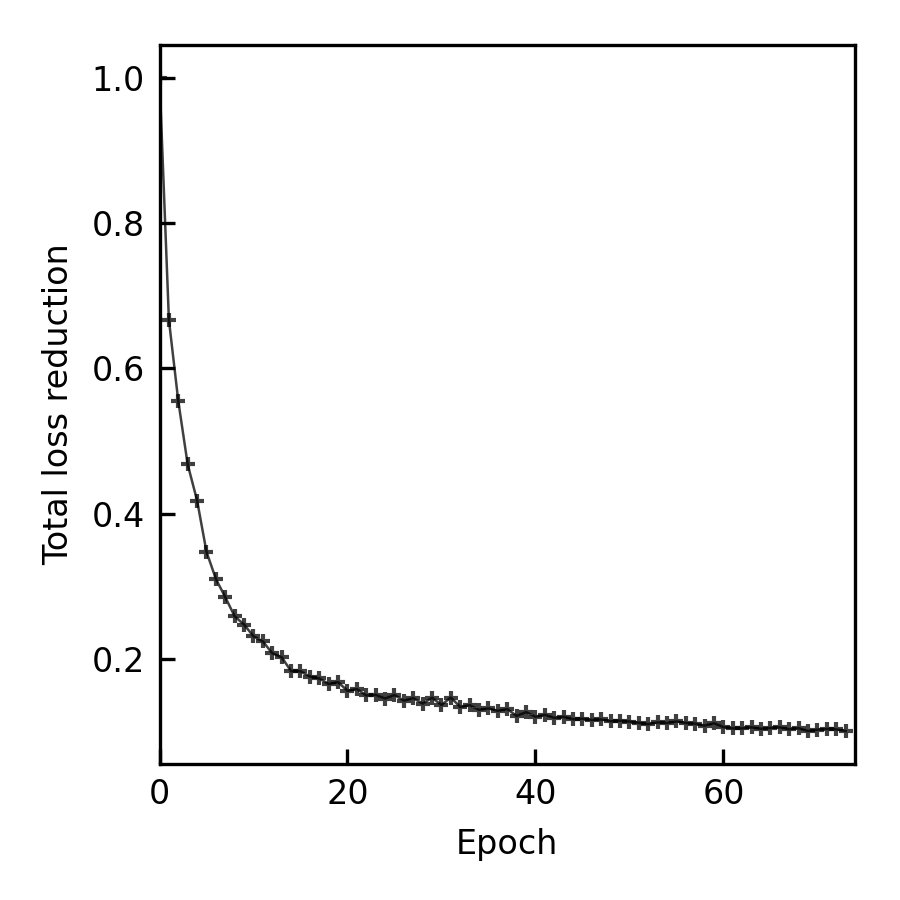} 
        \caption{DODE without density}
        \label{fig:converge_no_k}
    \end{subfigure}
    \begin{subfigure}{0.495\textwidth}
        \centering
        \includegraphics[width=\linewidth]{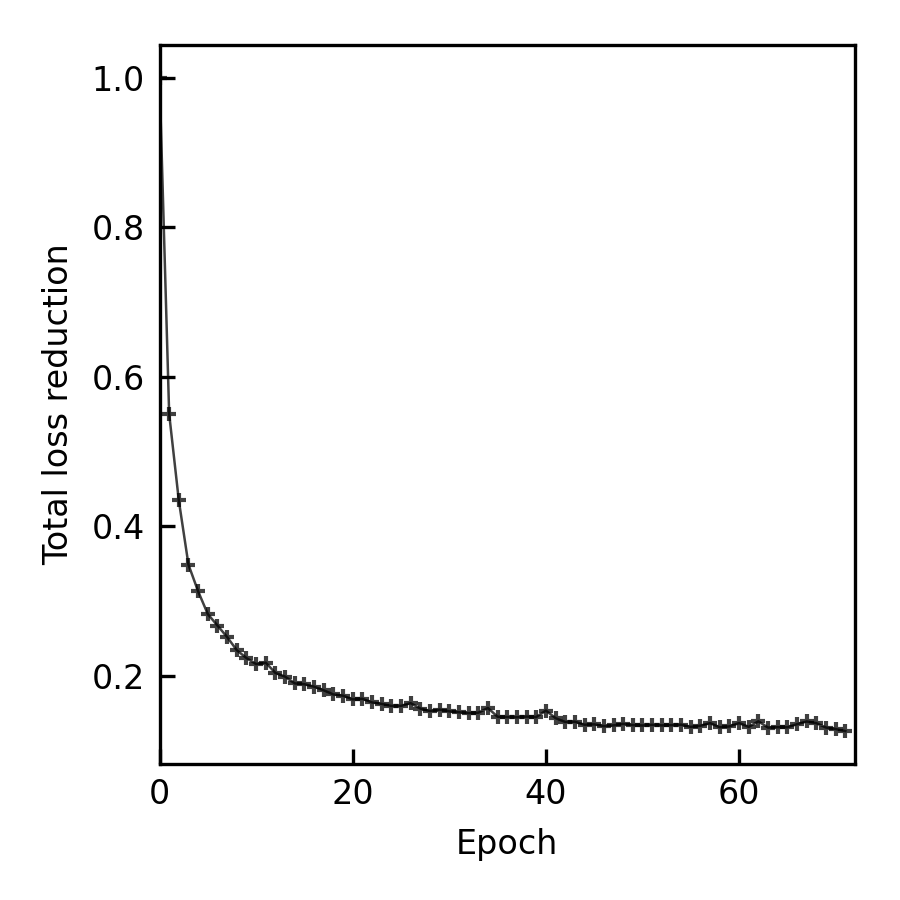}
        \caption{DODE with density}
        \label{fig:converge_k}
    \end{subfigure}    
    \caption{Convergence curve}
    \label{fig:convergence}
\end{figure}

\begin{figure}[H]
    \centering
    \begin{subfigure}{0.495\textwidth}
        \centering
        \includegraphics[width=\linewidth]{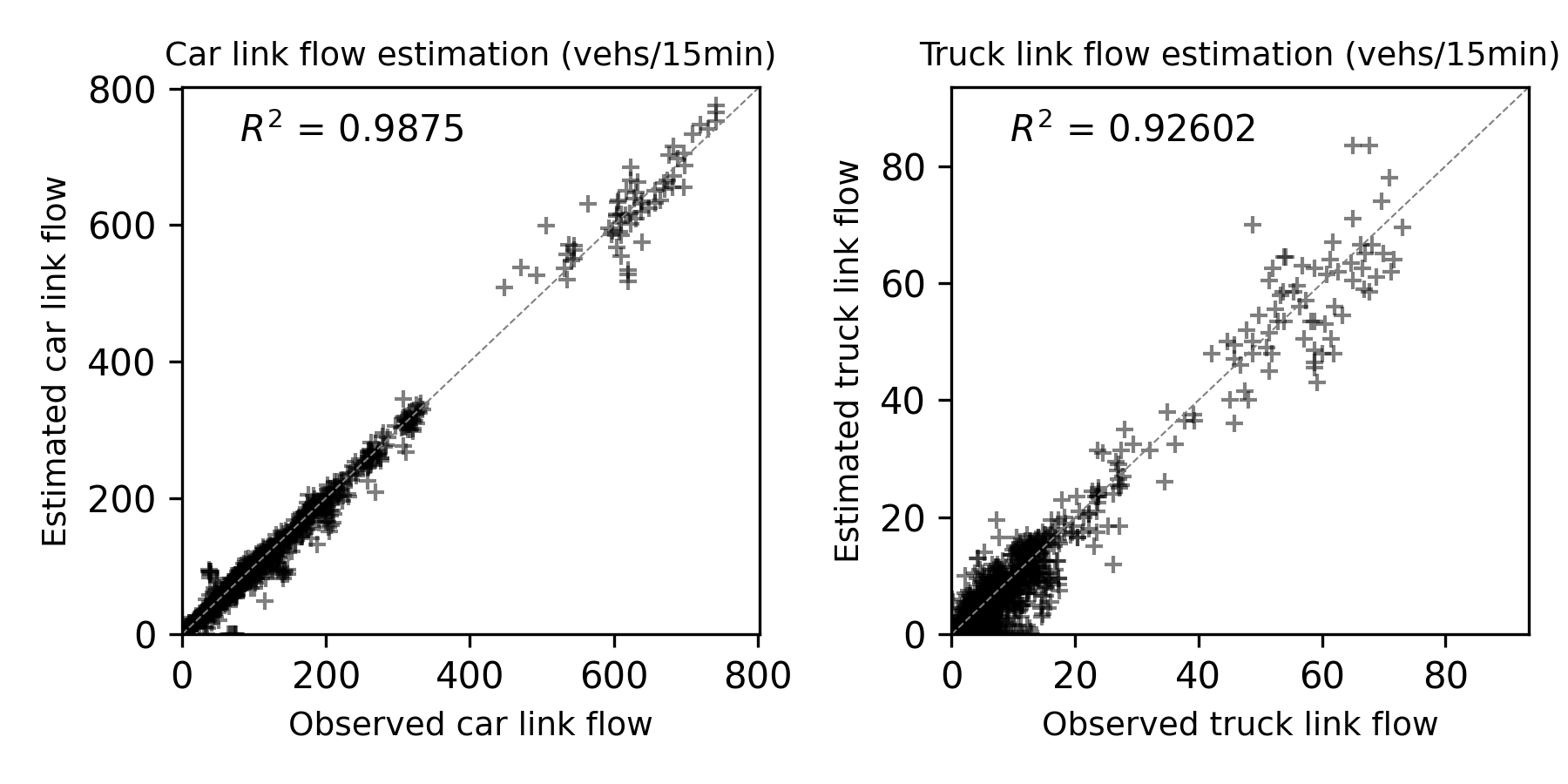}
        \caption{link flow estimation}
    \end{subfigure}
    \begin{subfigure}{0.495\textwidth}
        \centering
        \includegraphics[width=\linewidth]{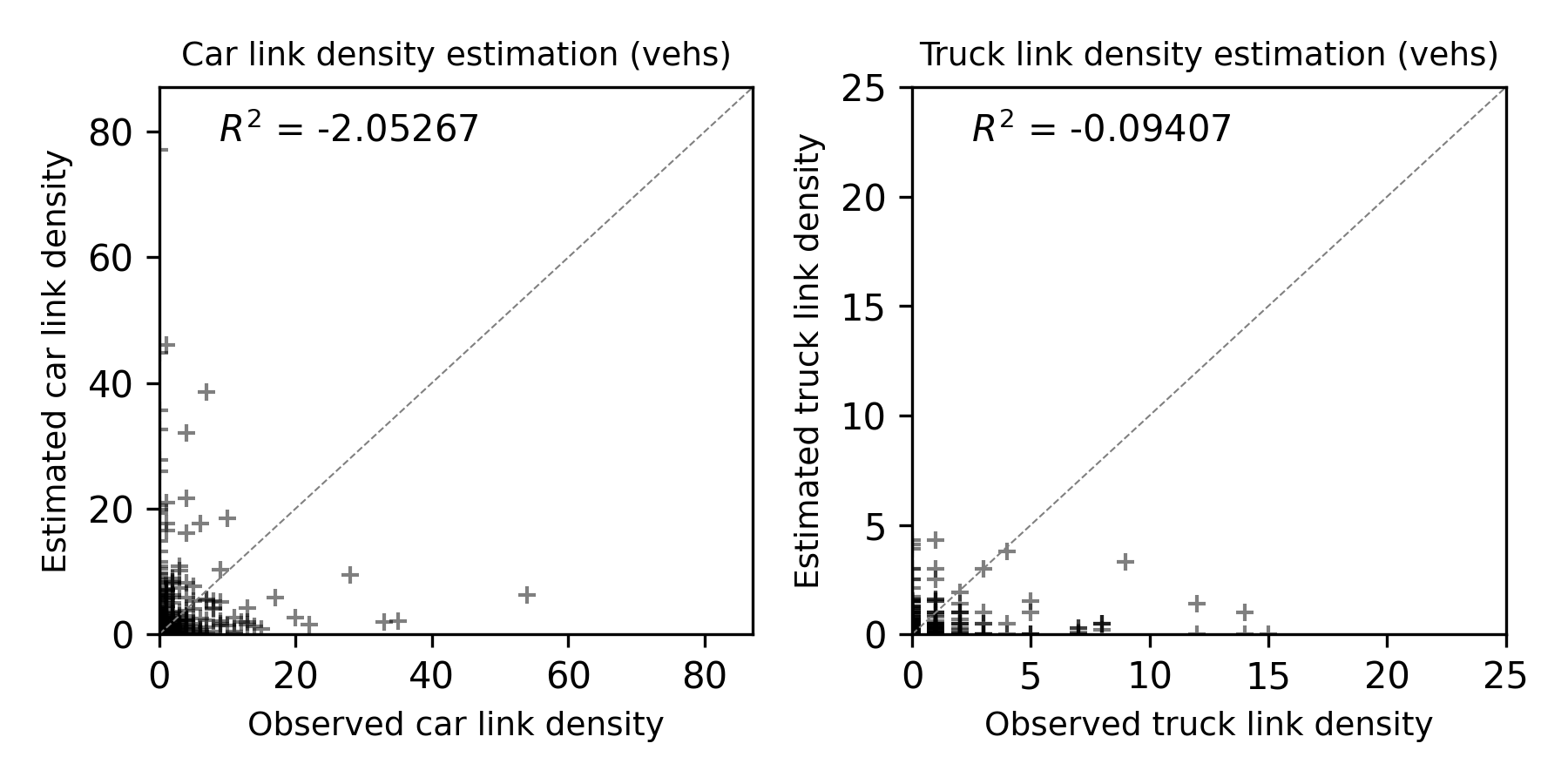}
        \caption{link density estimation}
    \end{subfigure}
    \caption{ {DODE without density data}}
    \label{fig:dode_wo_k}
\end{figure}

Figure~\ref{fig:convergence} shows the convergence curves for both DODE tests in the two scenarios. Both curves converge after around 70 epochs and the total reduction is similar, indicating that adding an additional density loss term in the objective function of DODE will not hurt the convergence performance. We further plot the estimation results of link-level conditions. Figure~\ref{fig:dode_wo_k} shows the accuracy of estimated conditions in DODE without density. It can be seen that link flow estimations match observations well, with R-squared values of above 0.9, but using the same network OD flow pattern cannot reproduce the densities. 

\begin{figure}[H]
    \centering
    \begin{subfigure}{0.495\textwidth}
        \centering
        \includegraphics[width=\linewidth]{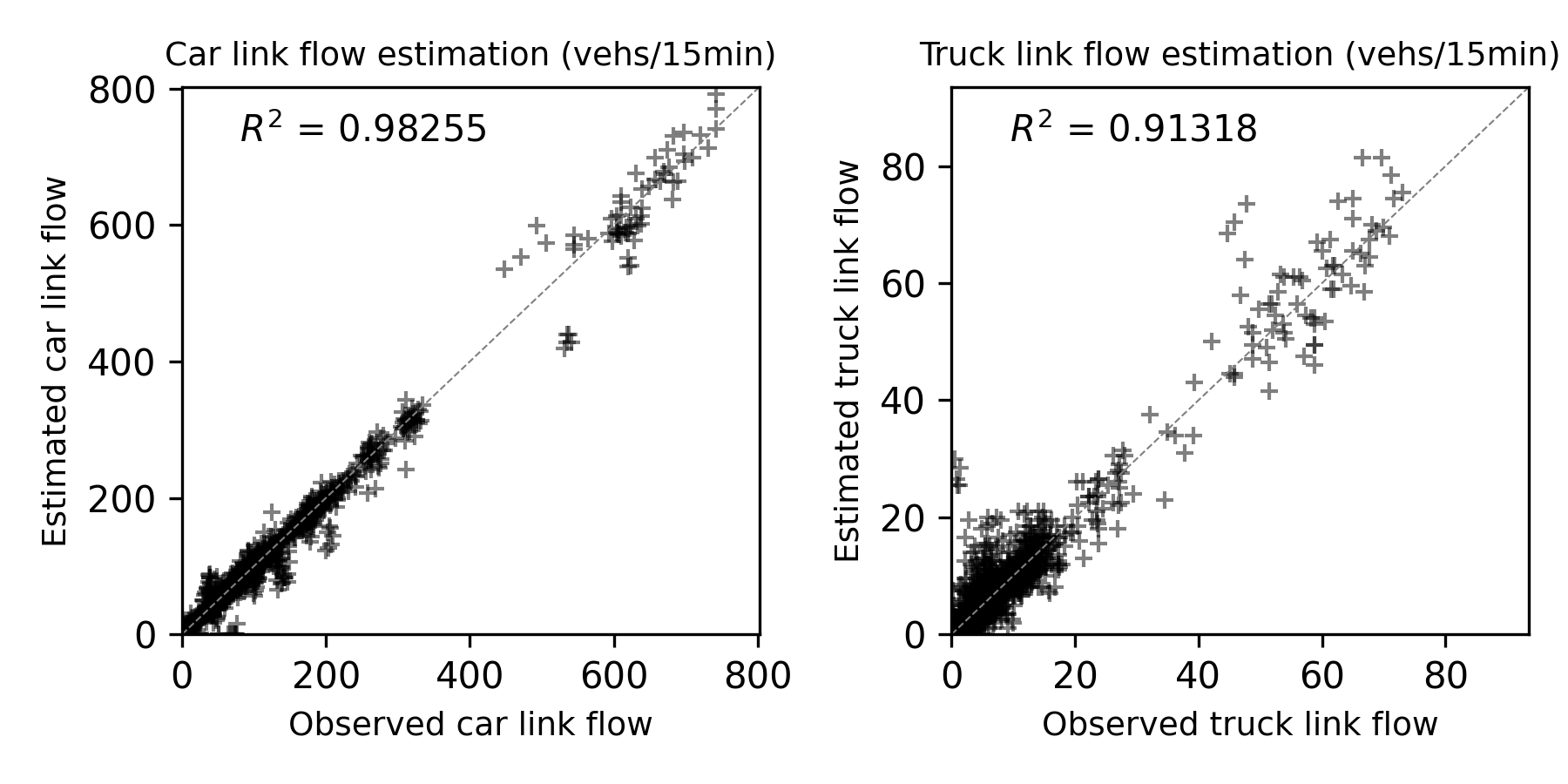}
        \caption{link flow estimation}
    \end{subfigure}
    \begin{subfigure}{0.495\textwidth}
        \centering
        \includegraphics[width=\linewidth]{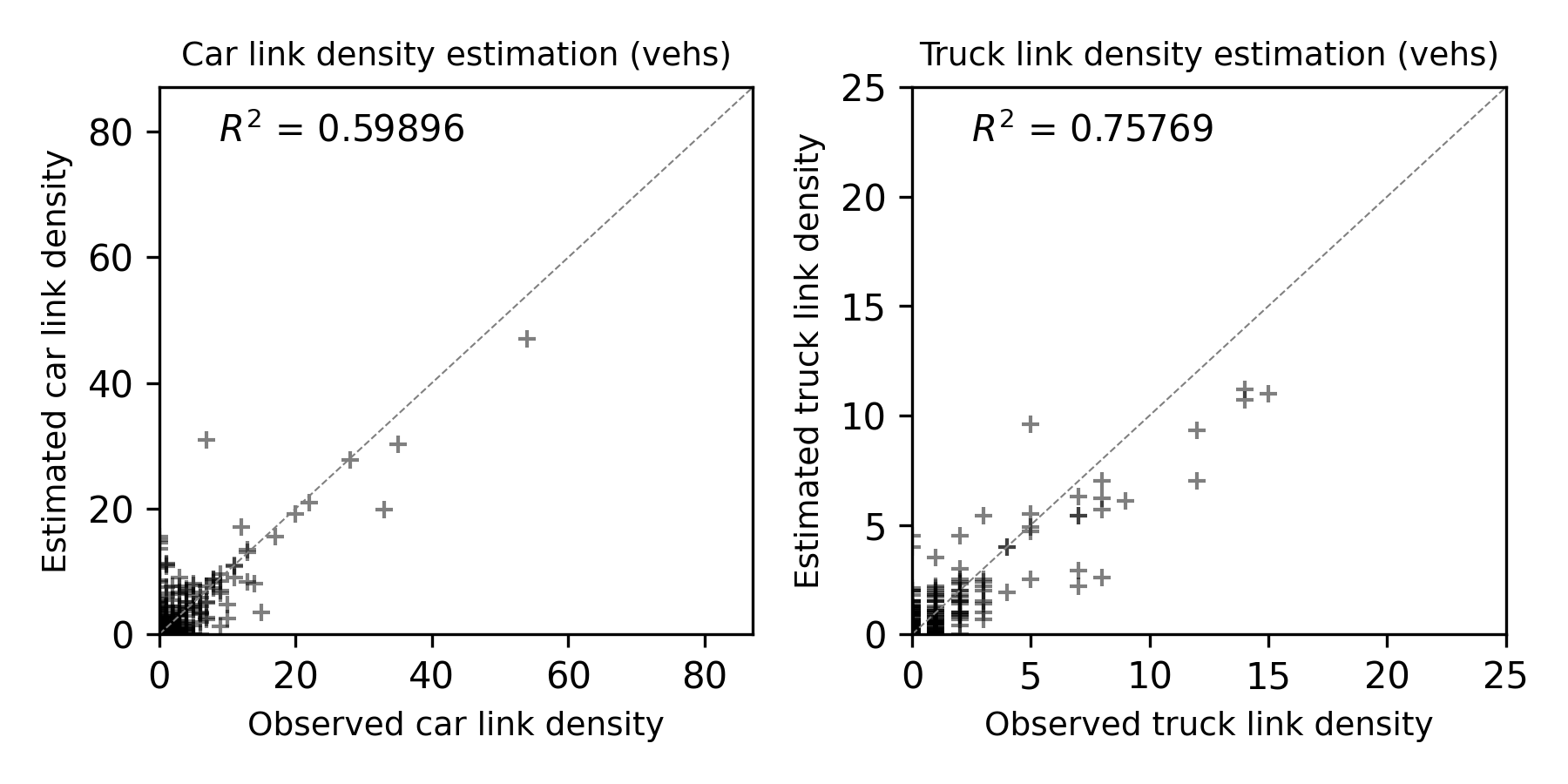}
        \caption{link density estimation}
    \end{subfigure}
    \caption{ {DODE with density data}}
    \label{fig:dode_w_k}
\end{figure}

\begin{figure}[H]
    \centering
    \includegraphics[width=1.0\linewidth]{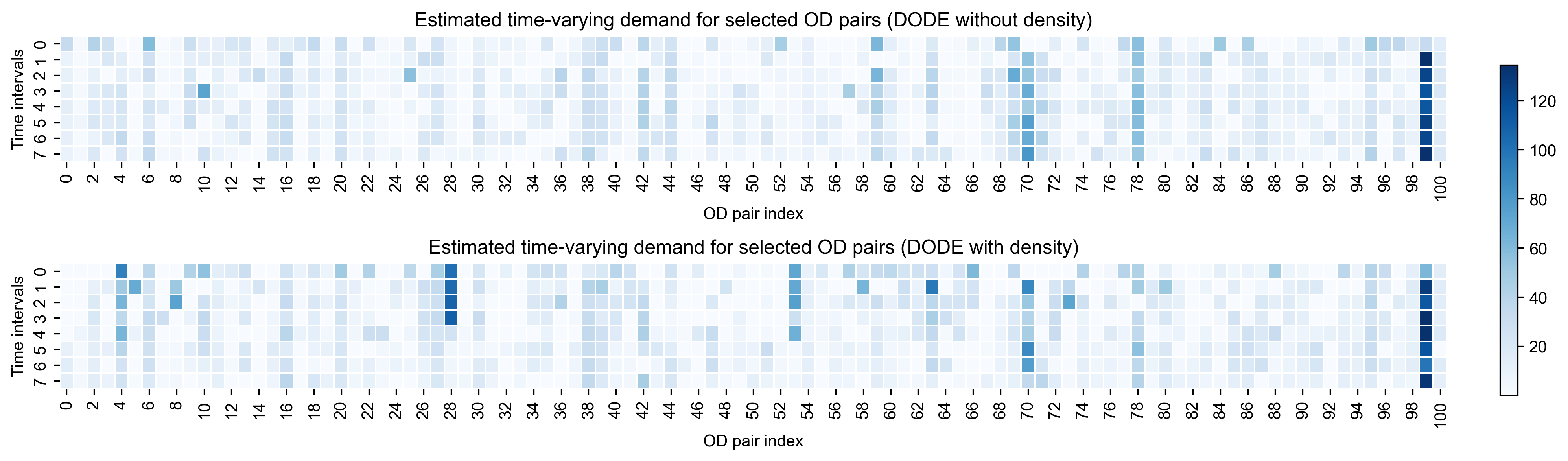}
    \caption{OD demand results from two scenario (car)}
    \label{fig:car_OD}
\end{figure}

\begin{figure}[H]
    \centering
    \includegraphics[width=1.0\linewidth]{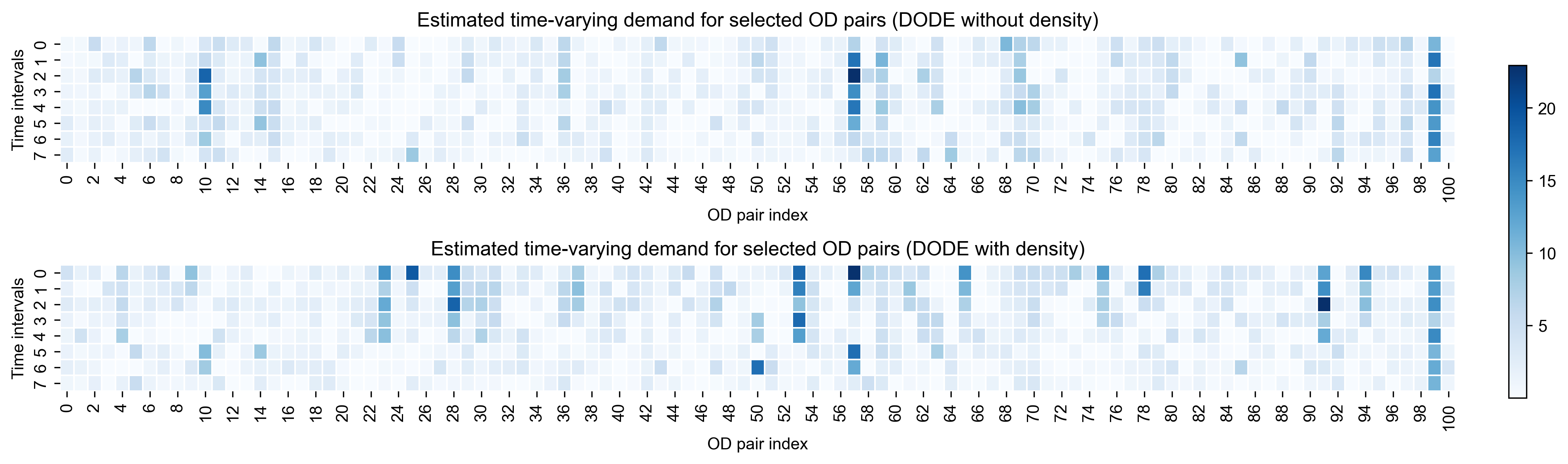}
    \caption{OD demand results from two scenario (truck)}
    \label{fig:truck_OD}
\end{figure}

However, when we incorporate density observations in the DODE, the results (Figure~\ref{fig:dode_w_k}) show that the estimated flows still match observed traffic counts (with a slight reduction in accuracy) and more importantly, the estimated density can match observations with much higher accuracy. This can be explained by the underdetermined nature of DODE problems. Utilizing partial observations (i.e., local sensor data) cannot guarantee the uniqueness of the OD matrix. 
The different demand patterns might lead to the same local traffic conditions, resulting in highly biased estimation, which is difficult to recognize.
Adding a new data source (i.e., density) adds more constraints to the problem, leading to a smaller feasible region and estimation results can be closer to the true values. We further plot the estimated OD patterns of these two DODE runs and observe some discrepancies for some OD pairs in Figure~\ref{fig:car_OD} and \ref{fig:truck_OD}.  {This experiment serves as a self-benchmarking ablation analysis to isolate the marginal contribution of satellite-derived density. The increased accuracy of density estimation demonstrates the value added by satellite imagery as a complementary data source.}

\begin{figure}[H]
    \centering
    \includegraphics[width=\linewidth]{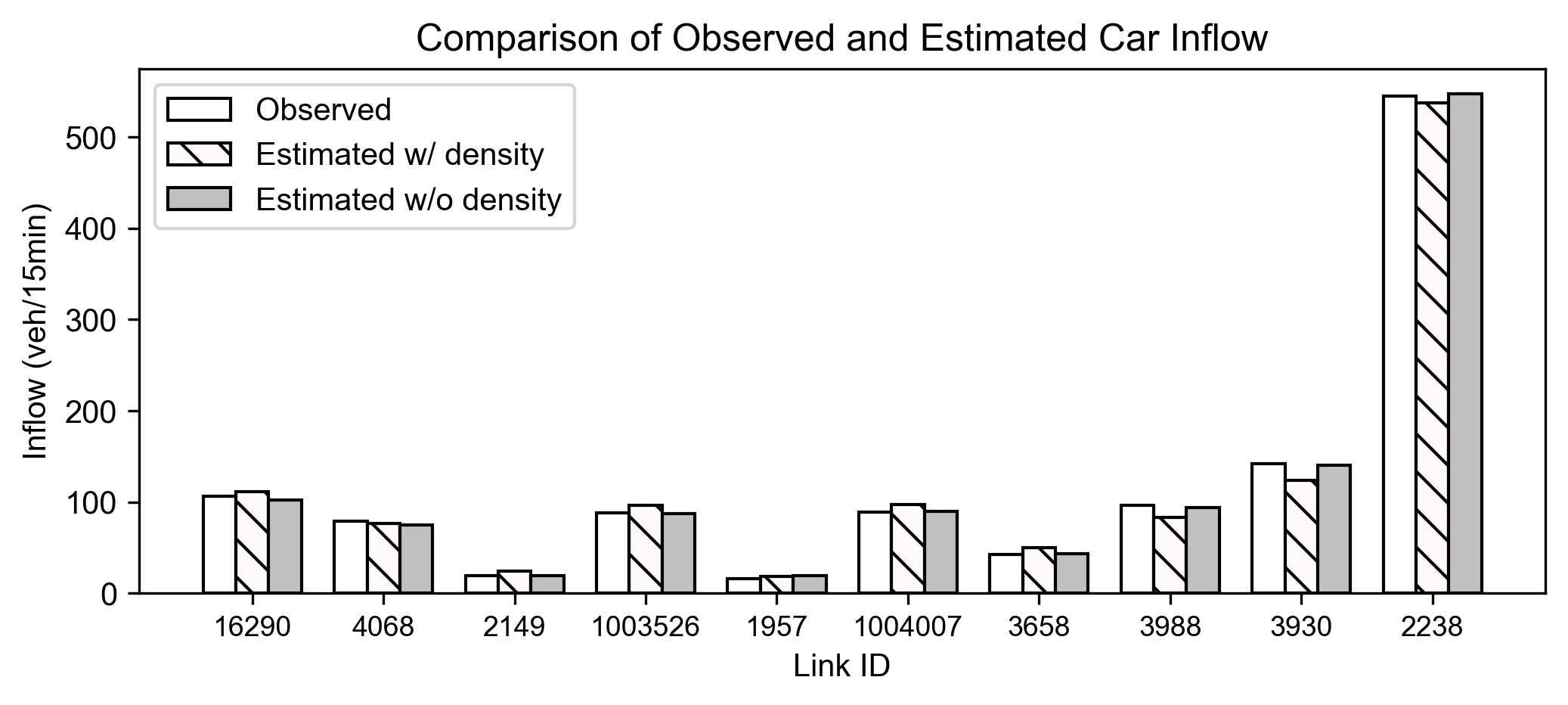}
    \caption{Car flow estimation of overlapping links}
    \label{fig:flow_car}
\end{figure}

\begin{figure}[H]
    \centering
    \includegraphics[width=\linewidth]{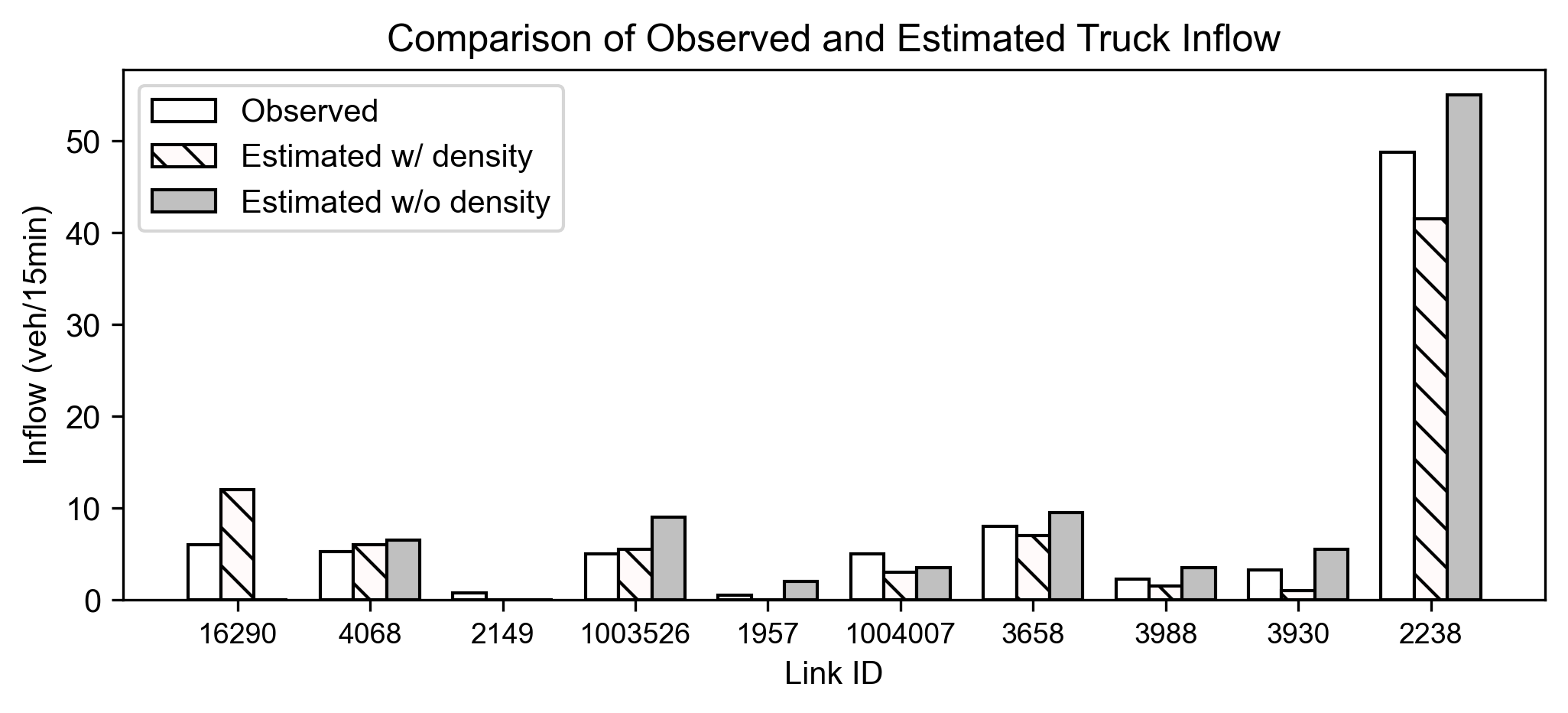}
    \caption{Truck flow estimation of overlapping links}
    \label{fig:flow_truck}
\end{figure}

We also investigate how adding density observations impacts the accuracy of the same localized traffic flow estimation due to the "trade-off" of fitting multi-source data. We focus on ten overlapping links with both traffic count and density observations and plot the estimation results in two scenarios. Because the satellite image only provides one density snapshot in the time interval of 12:15-12:30 PM, we compare the traffic flow estimation in this interval, shown in Figures~\ref{fig:flow_car} and \ref{fig:flow_truck}. The results show that the link flow estimations do not have large discrepancies between the two scenarios, while truck flow estimations perform worse than cars especially for observed flows that are smaller than 10. This is due to the stochasticity in the DNL model and the relatively low detection accuracy for trucks in the CV pipeline. Unknown inconsistencies between data sources also contribute to this difference.

 {From a computational cost perspective, incorporating density data does not change the main structure of the CG-based DODE framework. The dominant computation cost still comes from DNL runs in the forward pass that generates traffic state variables, and the computation of gradient backpropagation in the backward pass. Density is one of the state variables produced by the DNL model (extracted from cumulative curves), and its gradient is computed using the same set of DAR matrices that map path flows to link inflow/outflow. Therefore, adding the density loss does not introduce much additional computation cost in solving DODE.}

 {Moreover, satellite image processing is performed offline as a preprocessing step that converts raw images into link-level density observations, which can be used in the offline DODE. In the Pittsburgh case, the vehicle detection for each image took approximately 7 minutes end-to-end. In our implementation, the CV preprocessing time is smaller than the time for running DNL per iteration, so the overall runtime is dominated by OD estimation rather than image processing. The CV pipeline uses a sliding-window (tiling) strategy, so its complexity can be characterized by the number of tiles needed to cover the spatial extent of the network. The total inference cost in the CV pipeline grows roughly linearly with the area covered by the satellite imagery. Importantly, scaling to larger networks does not increase per-tile complexity and instead only increases the number of tiles. This makes the preprocessing step easy to parallelize across CPU cores or GPUs, and straightforward to scale using batch inference or distributed execution when needed.}

 {\subsubsection{Comparison with the low rank-based method}}
 {We use the Principal Component-based Simultaneous Perturbation Stochastic Approximation (PC-SPSA) method proposed by \cite{qurashi2019pc, qurashi2022dynamic} as a benchmark. PC-SPSA first learns a low-dimensional representation of dynamic OD demand by extracting principal components (PCs) from a set of historical OD samples, and then performs OD estimation in this reduced PC space. 
This benchmark represents a different line of work from our approach, as described in the literature review section. Our CG-based framework mitigates the underdetermined issue by incorporating heterogeneous, independent observations (traffic counts and satellite-derived densities) as regularization, whereas PC-SPSA relies on a low-rank assumption to reduce the DODE problem dimension via PCA.}

 {In the benchmark experiment, we use the same real-world observations (traffic counts and satellite-derived densities) as in our proposed model for PC-SPSA. We also follow the same initialization process by starting from a randomly generated dynamic OD matrix rather than a pre-specified seed OD. Because the two methods do not necessarily start from the same random point, their absolute objective values are not directly comparable across runs. Therefore, we report the relative objective reduction (normalized by the initial objective value) to show the convergence behaviors.}

 {For the Pittsburgh network, historical dynamic OD matrices are not available, so the low-rank structure of dynamic OD demand required by PCA cannot be directly inferred from real observations. To construct a feasible PCA-based benchmark, we generate a set of OD samples following the synthetic data generation procedure in \cite{qurashi2022dynamic}. Specifically, we first obtain one dynamic OD matrix from a prior DODE run and use it as a template to generate additional OD samples (i.e., it is not used as the initialization in solving DODE with PC-SPSA). We generate 299 additional dynamic OD matrices, each covering a 2-hour time horizon (8 intervals at 15-minute resolution). This yields 300 OD matrices in total, corresponding to 2400 interval-level OD vectors for each vehicle class, which are used to compute the PCs. We compare the two methods in terms of (1) convergence speed, assessed by the normalized objective function reduction over the same number of iterations, and (2) goodness-of-fit between estimated and observed traffic states, quantified by $R^2$ for class-specific link flows and densities.}

\begin{figure}[H]
    \centering
    \includegraphics[width=\linewidth]{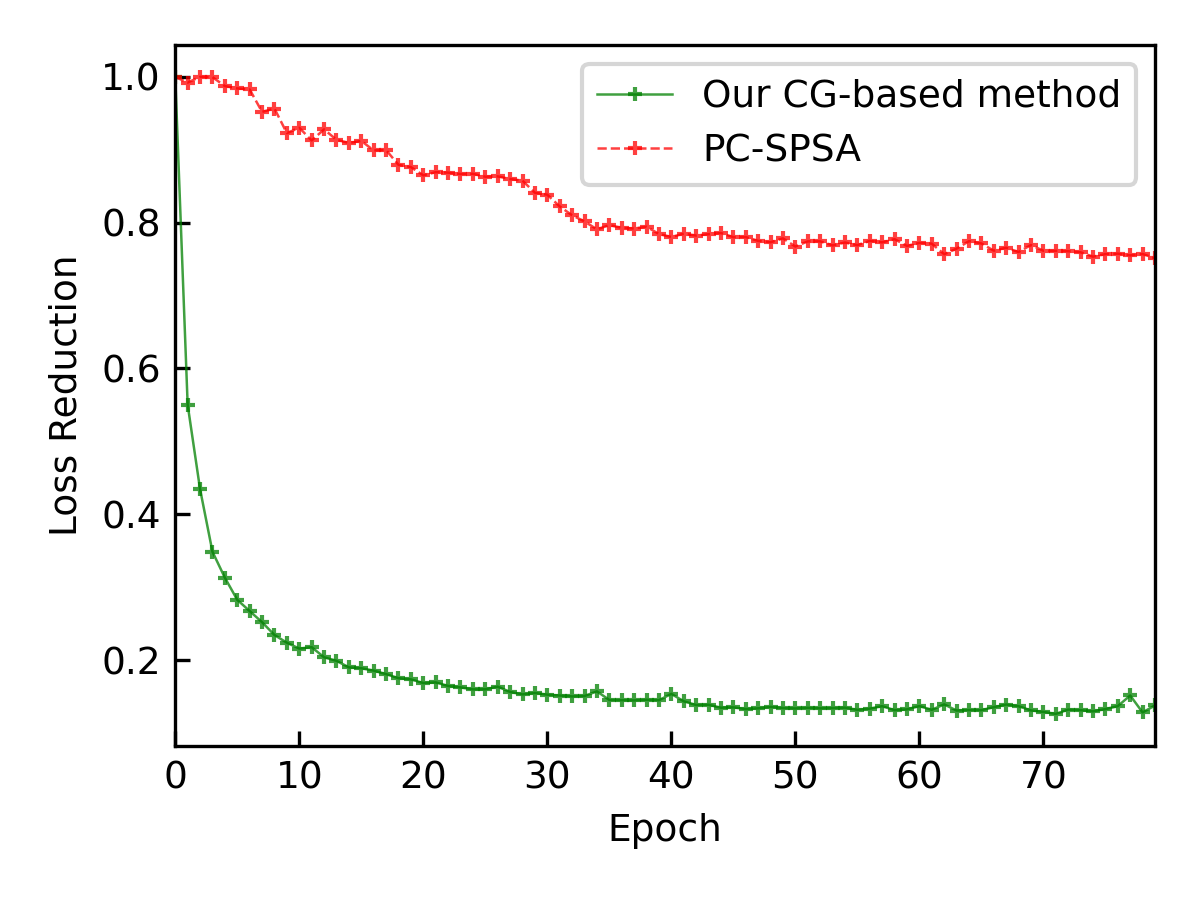}
    \caption{Normalized objective reduction (relative to the initial objective) for CG-based method vs. PC-SPSA on the Pittsburgh network}
    \label{fig:loss_bm}
\end{figure}

\begin{figure}[H]
    \centering
    \includegraphics[width=\linewidth]{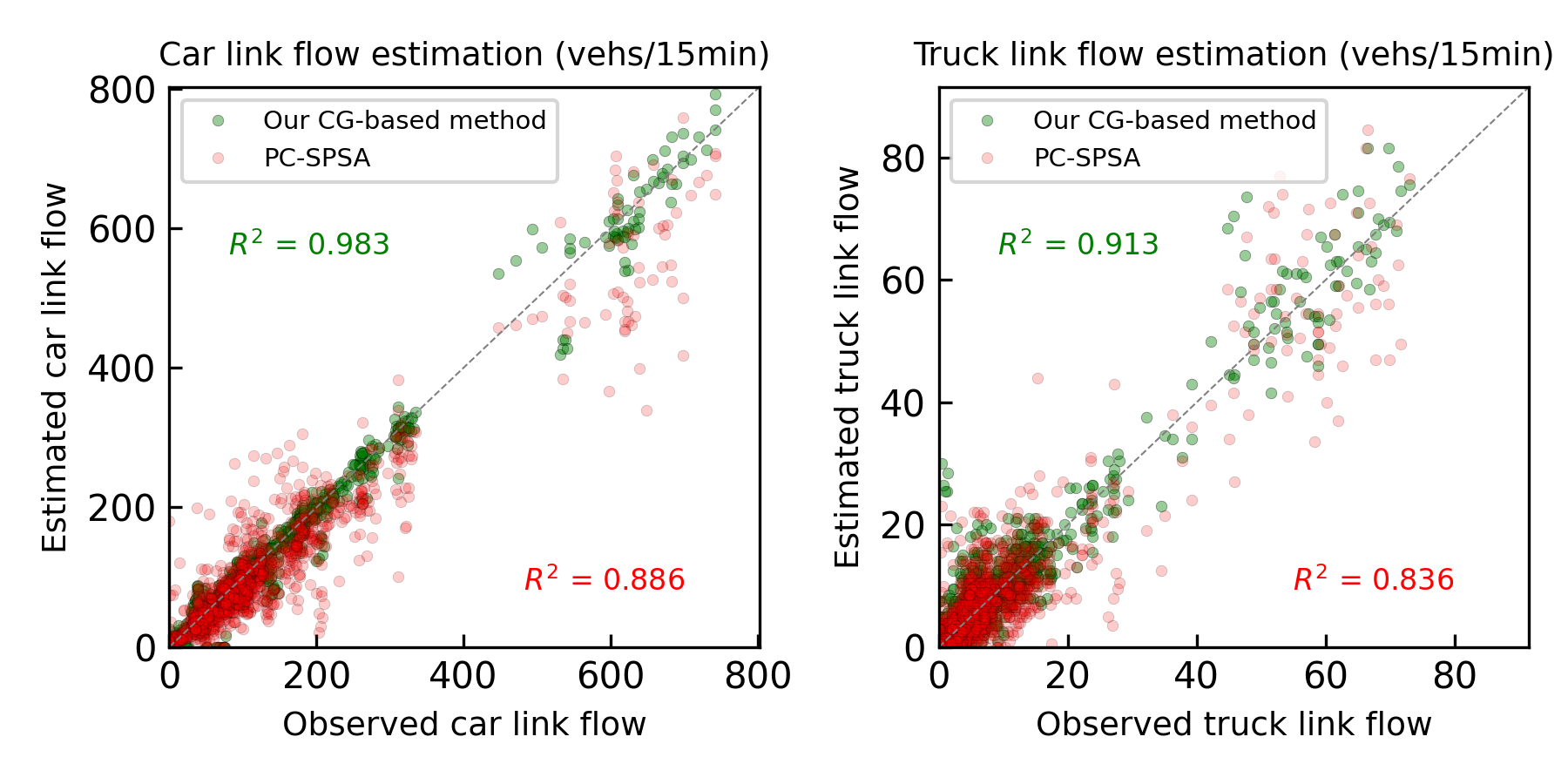}
    \caption{Comparison for class-specific link flow estimation (CG-based vs. PC-SPSA)}
    \label{fig:benchmark_flow}
\end{figure}

\begin{figure}[H]
    \centering
    \includegraphics[width=\linewidth]{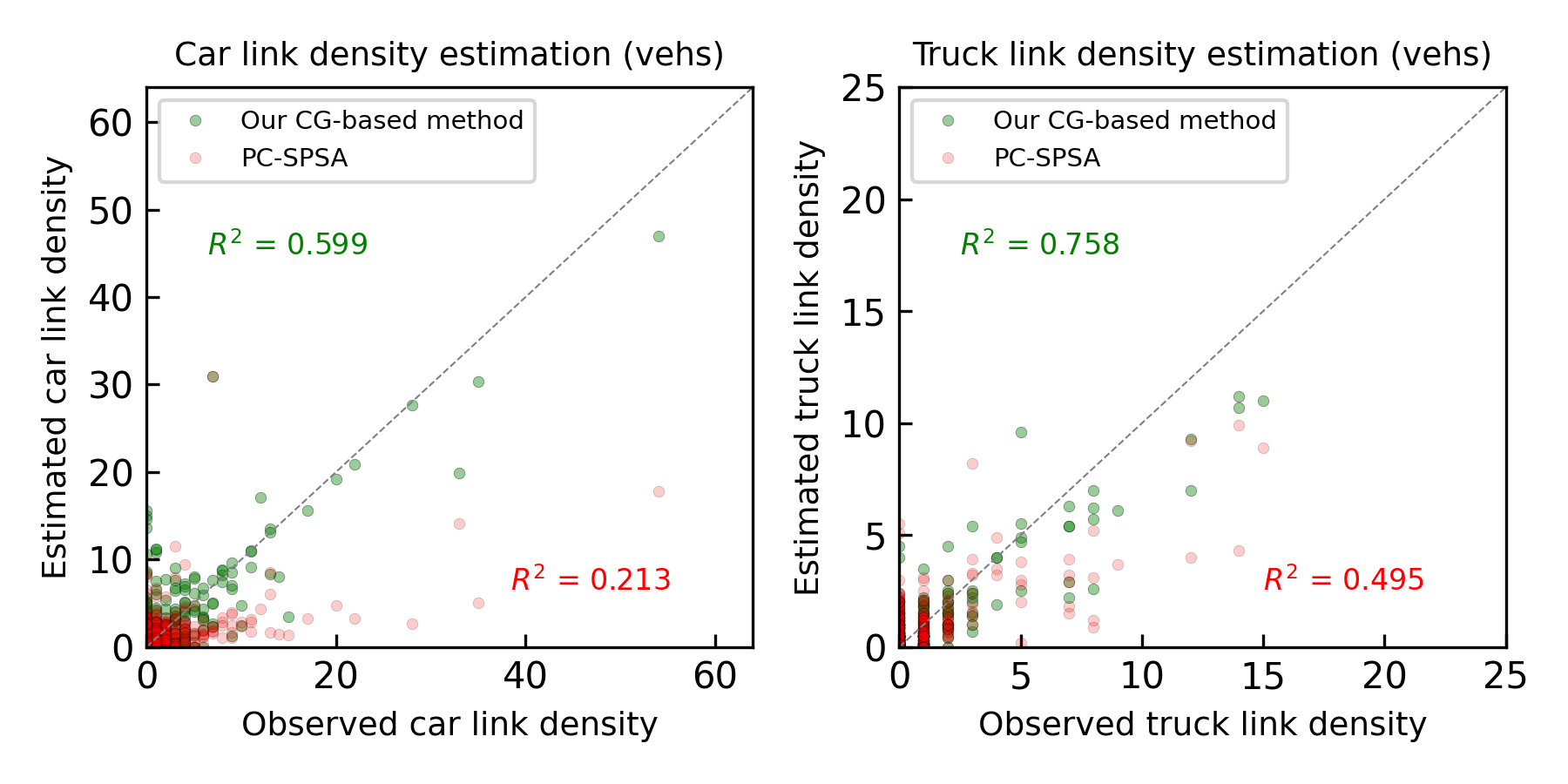}
    \caption{Comparison for class-specific link density estimation (CG-based vs. PC-SPSA)}
    \label{fig:benchmark_k}
\end{figure}

 {Results show that PC-SPSA underperforms our method on the Pittsburgh network in both convergence speed (Figure~\ref{fig:loss_bm}) and state estimation accuracy (Figures~\ref{fig:benchmark_flow} and \ref{fig:benchmark_k}). This finding is consistent with \cite{Ma2020} who reported worse performance of the SPSA method compared with CG-based approaches on a large-scale network. We explain this performance gap from three perspectives. First, the SPSA method estimates gradients of the objective function via stochastic perturbations, while our method computes exact gradients with respect to flow and density through the DAR matrices after each DNL run, yielding a deterministic and interpretable mapping from OD flows to link-level traffic states. Although PCA reduces the DODE problem to a lower dimension, the stochastic-gradient nature of SPSA can still slow convergence and degrade fitting, especially for density observations that couple in and out link flows across multiple time intervals through accumulation and propagation effects. Second, without real-world historical OD data, it remains unclear whether the true OD patterns in Pittsburgh have a low-rank structure. If this assumption is not valid, PCA-based compression may discard spatio-temporal variations of OD demand and degrade the estimation. Third, PC-SPSA performance is sensitive to hyperparameter choices; despite testing settings reported in \cite{qurashi2019pc, qurashi2022dynamic} and additional tuning on our dataset, improvements were limited and more extensive tuning may be required to yield better performance. Overall, our method yields better performance on the Pittsburgh network and we leave more extensive benchmarking experiments to future work when richer real-world historical OD datasets become available.}

\subsection{Pittsburgh Downtown network with synthetic data}
Given the limited coverage of real-world traffic data in Section~\ref{sec:real_net}, this experiment uses synthetic data to conduct a sensitivity analysis and evaluate the robustness of the proposed DODE framework incorporating density observations. We use the estimated multi-class OD demand from Scenario 2 in Section~\ref{sec:real_net} (which incorporates both traffic count and density snapshots) as the "ground truth" OD demand. This demand is fed into the DNL model to generate time-varying traffic conditions across all links, serving as the synthetic ground truth observations. 

Following the data generation process described in Section~\ref{sec:toy_net}, we randomly sample 400 links to form the observation link set $A_{obs}$, and introduce stochastic errors to traffic count and density data to simulate sensing errors. The remaining links form the link set $A_{unobs}$ for which only density observations with stochastic errors are available. These corrupted observations, including traffic counts of observed links and density on all links, are used as input for DODE.

We evaluate the accuracy of estimated traffic conditions separately for observed and unobserved links. This section focuses on analyzing the sensitivity of the DODE framework to varying data conditions, especially the level of sensing error and the frequency of imagery-based density snapshots. To reduce the bias introduced by the spatial distribution of local sensors, each DODE test is repeated with five different random samples of $A_{obs}$ under each experimental setting. 

We first assess the sensitivity of DODE to sensing errors in density observations. As discussed in Section~\ref{sec:data_availability}, density observations may require aggregation when serving as inputs for DODE and aggregating moving and through vehicles together can introduce sensing errors. Additionally, inaccuracies in class-specific vehicle detection further contribute to noise in sensing results. 

To evaluate the impact of such errors, this section considers two levels of sensing errors: (1) $\pm 10\%$ with perturbation $\epsilon_{10\%}\sim\text{Unif}(0.9, 1.1)$ (2) $\pm 20\%$ with perturbation $\epsilon_{20\%}\sim\text{Unif}(0.8, 1.2)$. For each error level, 5 DODE runs are conducted using independently sampled observed link sets ($A_{\text{obs}}$) to mitigate the impact related to the spatial sampling of local sensors. Consistent with prior sections, $R^2$ is used to evaluate model performance. To assess model generalization, we report separate $R^2$ values for observed and unobserved links to evaluate generalization.

\begin{figure}
    \centering
    \begin{subfigure}{0.32\textwidth}
        \centering
        \includegraphics[width=\linewidth]{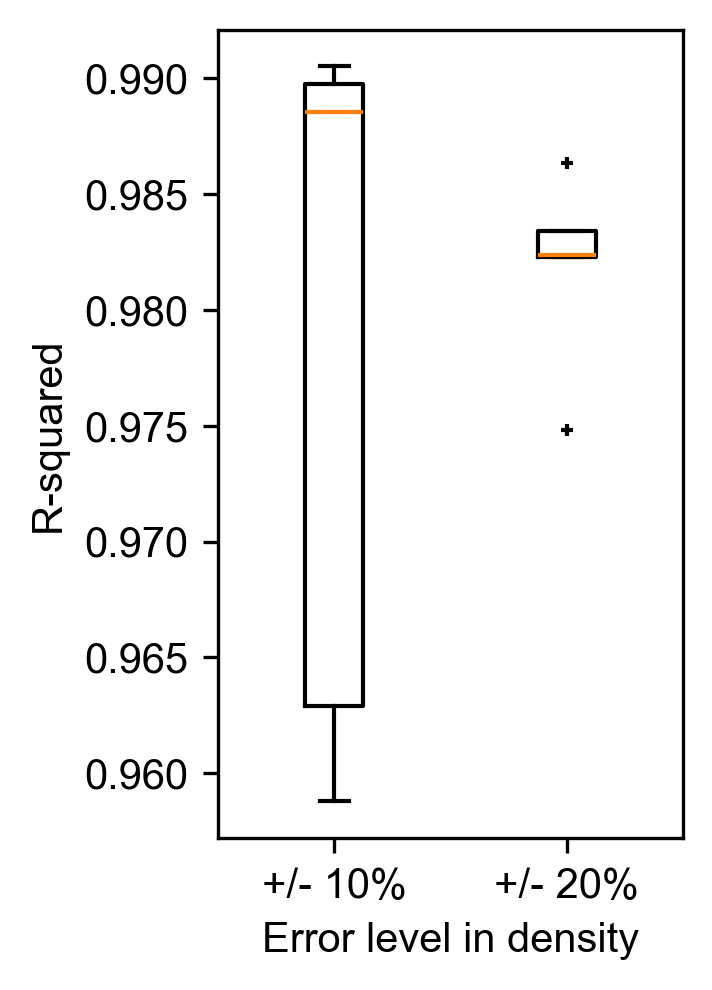}
        \caption{observed car flow}
        \label{fig:observed_car_flow}
    \end{subfigure}
\hfill
    \begin{subfigure}{0.32\textwidth}
        \centering
        \includegraphics[width=\linewidth]{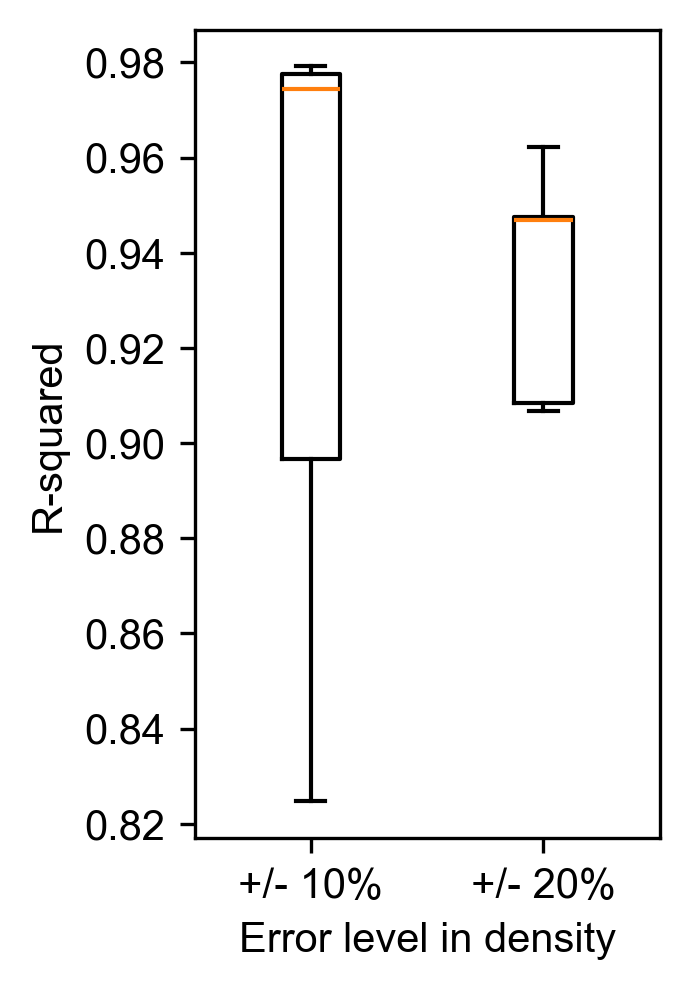}
        \caption{observed truck flow}
        \label{fig:observed_truck_flow}
    \end{subfigure}
\hfill
    \begin{subfigure}{0.32\textwidth}
        \centering
        \includegraphics[width=\linewidth]{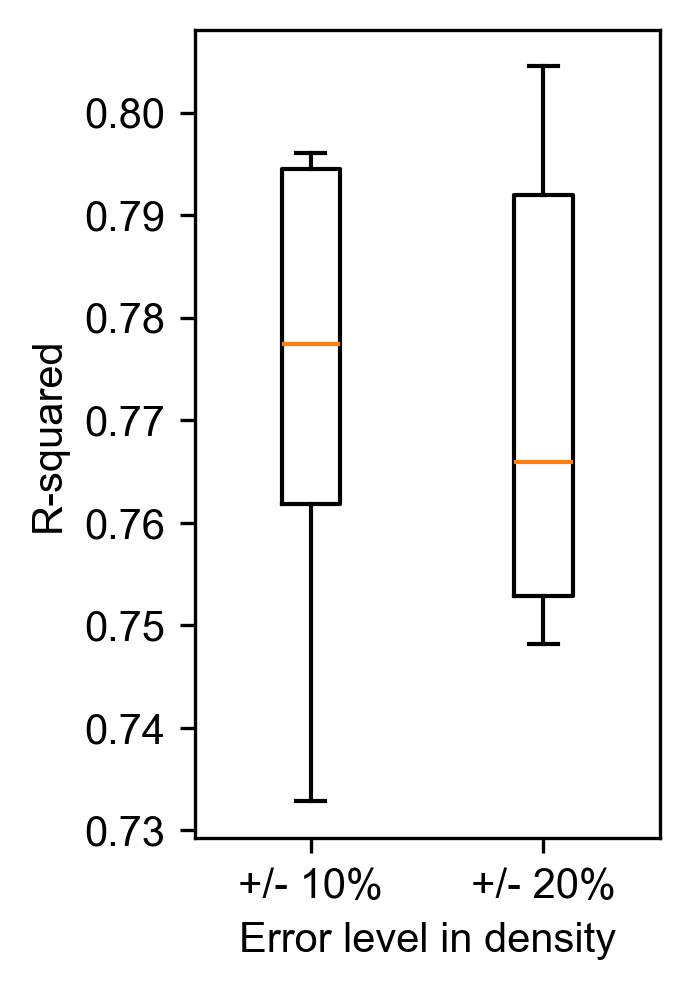}
        \caption{unobserved car flow}
        \label{fig:unobserved_car_flow}
    \end{subfigure}
\hfill
    \begin{subfigure}{0.32\textwidth}
        \centering
        \includegraphics[width=\linewidth]{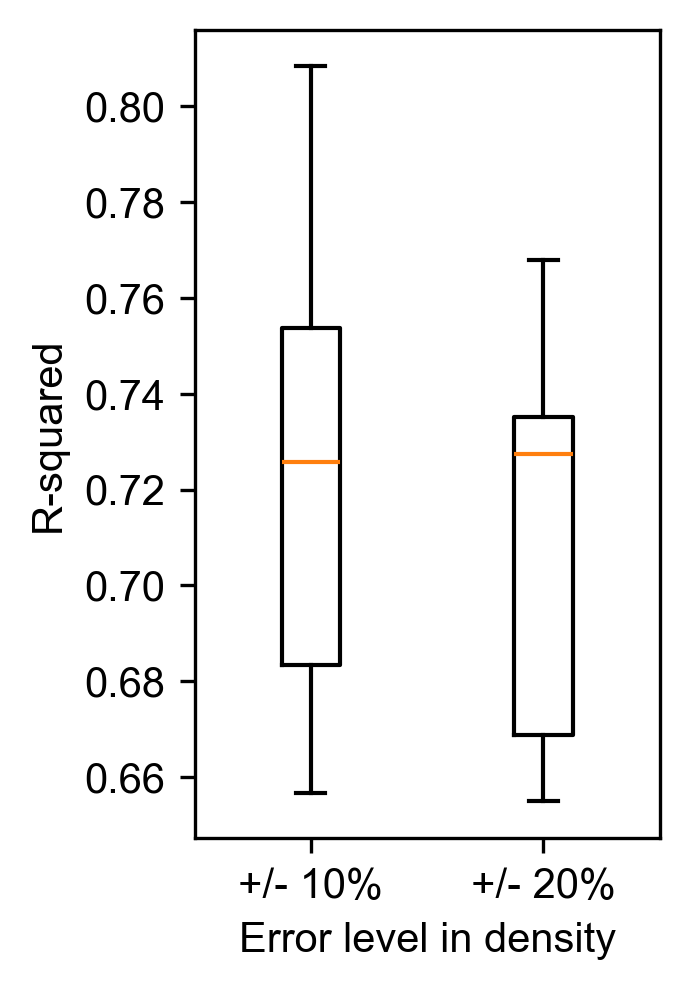}
        \caption{unobserved truck flow}
        \label{fig:unobserved_truck_flow}
    \end{subfigure}
\hfill
    \begin{subfigure}{0.32\textwidth}
        \centering
        \includegraphics[width=\linewidth]{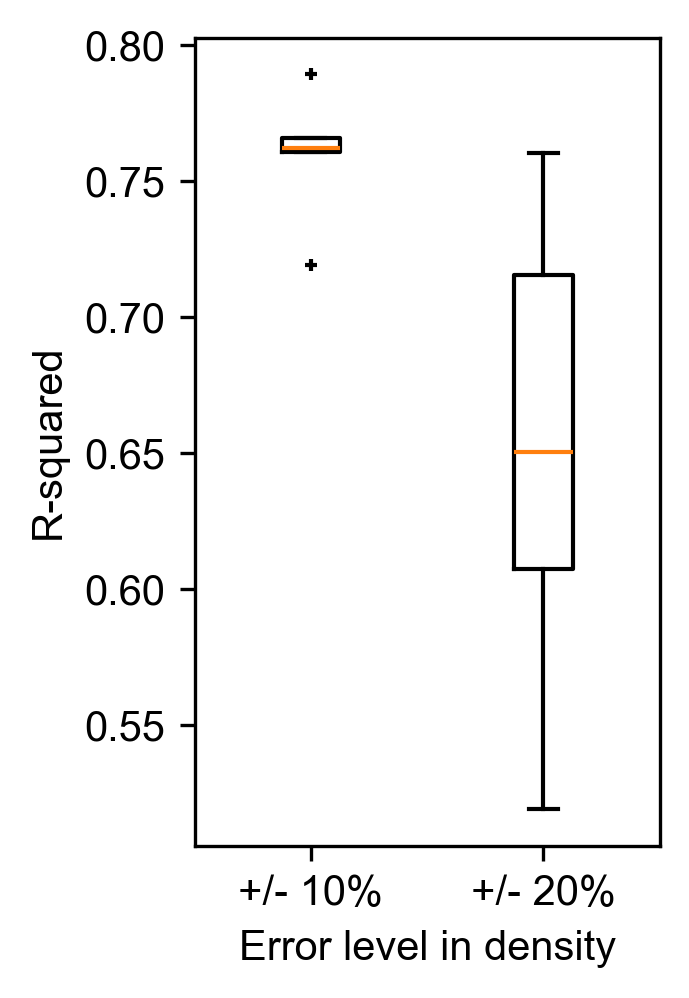}
        \caption{observed car density}
        \label{fig:observed_car_k}
    \end{subfigure}
\hfill
    \begin{subfigure}{0.32\textwidth}
        \centering
        \includegraphics[width=\linewidth]{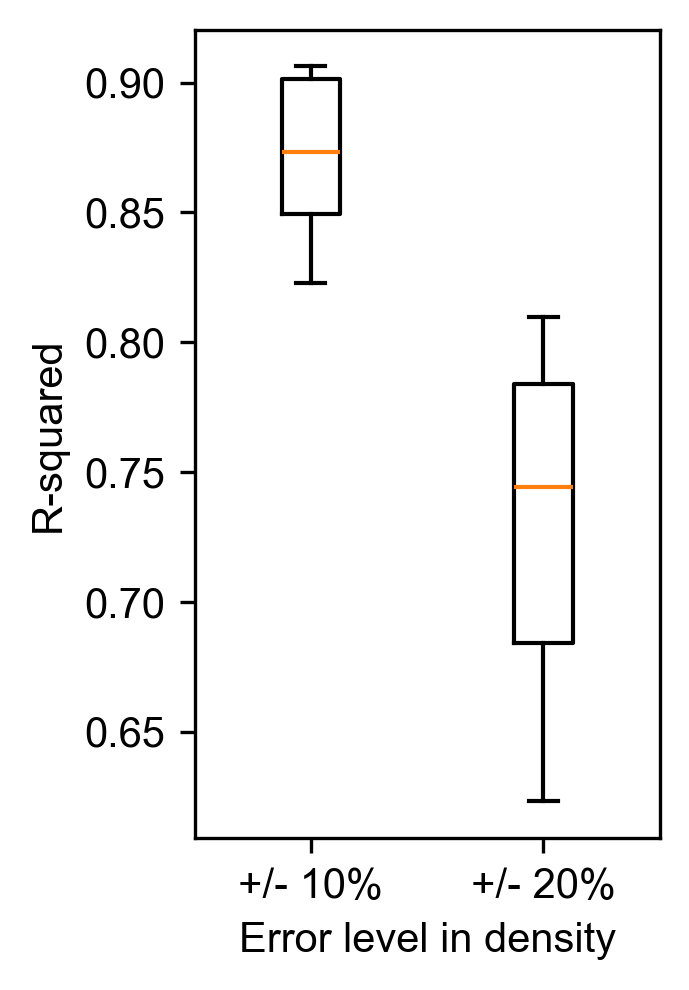}
        \caption{observed truck density}
        \label{fig:observed_truck_k}
    \end{subfigure}
    \caption{Sensitivity analysis for density sensing error}
    \label{fig:sa_error}
\end{figure}

Figure~\ref{fig:sa_error} shows a clear trend that increasing sensor error level from $\pm 10\%$ to $\pm 20\%$ leads to a noticeable decline in $R^2$ values, indicating that estimation accuracy is reduced. The flow estimates for unobserved links are relatively more robust, maintaining similar performance under both error level settings, but for density estimation, $R^2$ deteriorates with higher error level because sensing errors can reduce the consistency in network-wide density observations. 
 {The perturbation errors also represent the vehicle detection errors and our experiments show that estimation accuracy for cars and trucks (on both observed and unobserved links by local sensors) degrades only modestly under substantial error perturbations.}
Overall, these results demonstrate that the sensing errors in density extraction from imagery data do not have a large impact on the local traffic estimation at unobserved locations but do impact overall density estimation itself.

Next sensitivity analysis evaluates the impact of the capture frequency of satellite imagery. We fix the level of sensing errors in density observations at $\pm 10\%$ and change density snapshot frequencies to two different levels: (1) one density snapshot per 15 min and (2) one density snapshot per 30 min. Similarly, 5 DODE runs are conducted in each data setting, and separate $R^2$ values for observed and unobserved links are reported to assess model performance.

\begin{figure}
    \centering
    \begin{subfigure}{0.32\textwidth}
        \centering
        \includegraphics[width=\linewidth]{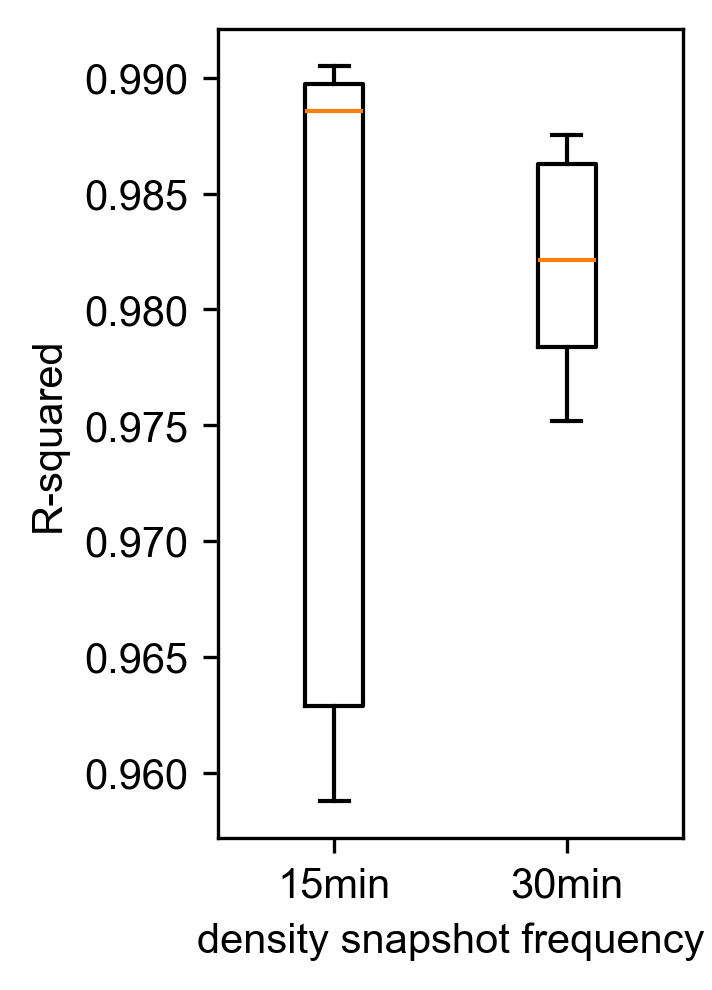}
        \caption{observed car flow}
        \label{fig:observed_car_flow_freq}
    \end{subfigure}
\hfill
    \begin{subfigure}{0.32\textwidth}
        \centering
        \includegraphics[width=\linewidth]{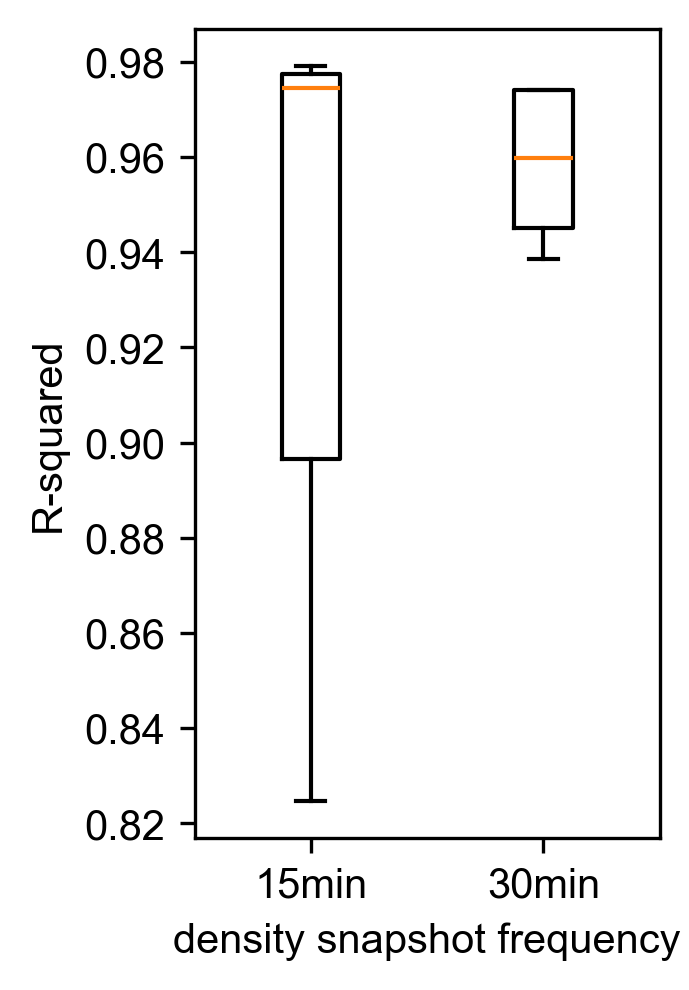}
        \caption{observed truck flow}
        \label{fig:observed_truck_flow_freq}
    \end{subfigure}
\hfill
    \begin{subfigure}{0.32\textwidth}
        \centering
        \includegraphics[width=\linewidth]{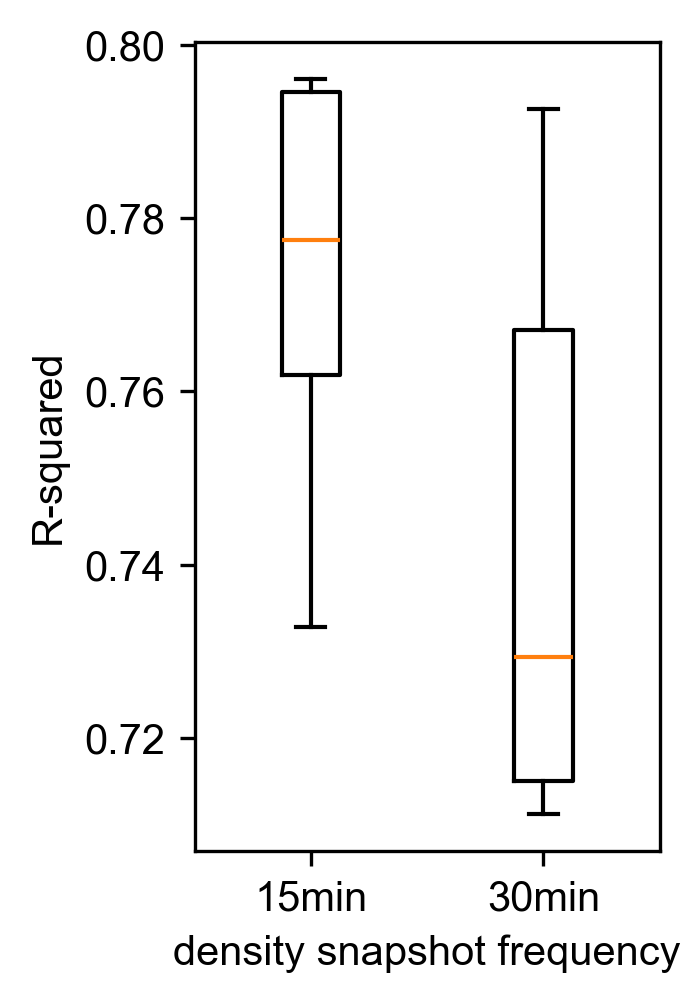}
        \caption{unobserved car flow}
        \label{fig:unobserved_car_flow_freq}
    \end{subfigure}
\hfill
    \begin{subfigure}{0.32\textwidth}
        \centering
        \includegraphics[width=\linewidth]{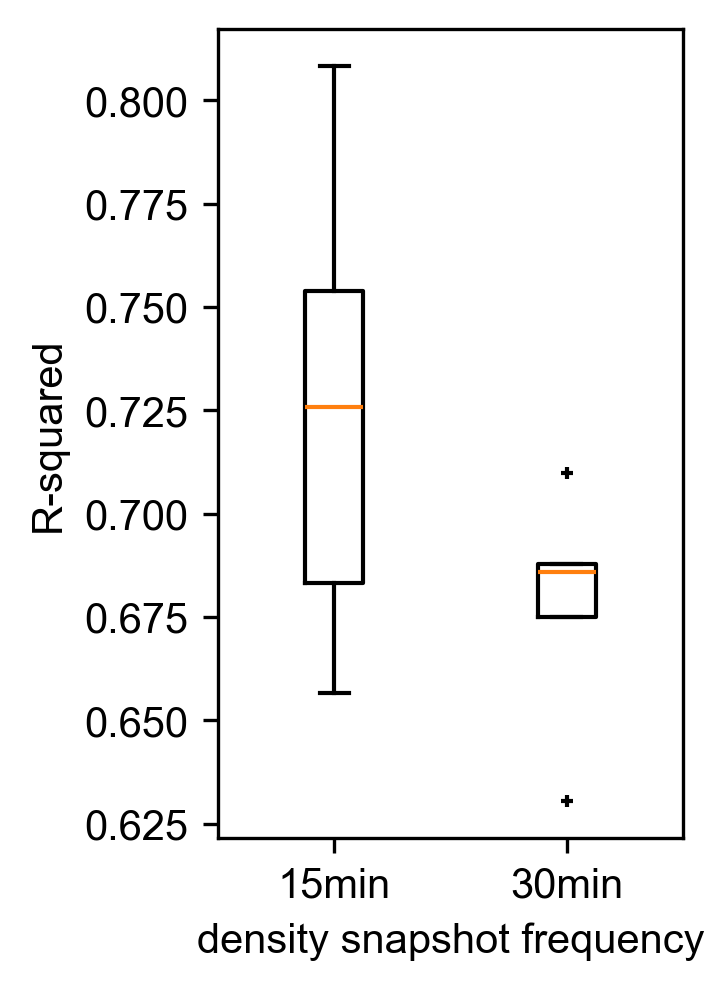}
        \caption{unobserved truck flow}
        \label{fig:unobserved_truck_flow_freq}
    \end{subfigure}
\hfill
    \begin{subfigure}{0.32\textwidth}
        \centering
        \includegraphics[width=\linewidth]{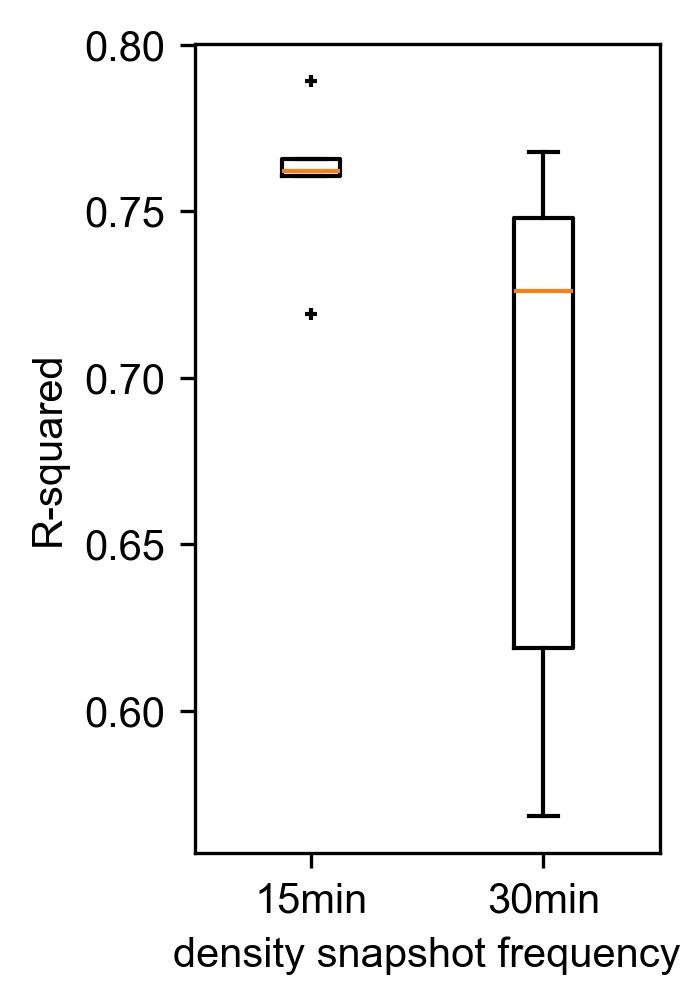}
        \caption{observed car density}
        \label{fig:observed_car_k_freq}
    \end{subfigure}
\hfill
    \begin{subfigure}{0.32\textwidth}
        \centering
        \includegraphics[width=\linewidth]{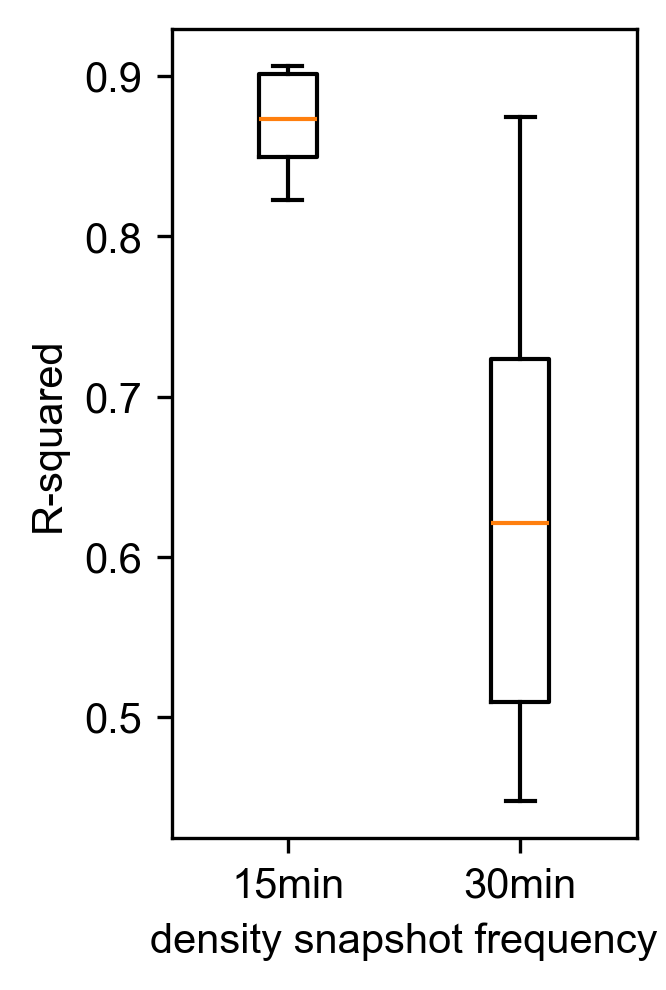}
        \caption{observed truck density}
        \label{fig:observed_truck_k_freq}
    \end{subfigure}
    \caption{Sensitivity analysis for density snapshot frequency}
    \label{fig:sa_freq}
\end{figure}

Figure~\ref{fig:sa_freq} shows that across all metrics, estimation accuracy, as measured by R-squared, tends to decline as the density snapshot frequency decreases from 15 to 30 minutes. This indicates that denser temporal coverage contributes to more accurate calibration.The impact is more pronounced for truck flow estimation than for car flow, as truck volumes are generally lower and thus more sensitive to observational constraints. For density estimation, the 30-min snapshot frequency results in noticeably poorer performance compared to the 15-minute case, likely due to increased uncertainty in the aggregated density observations. Higher snapshot frequency captures more detailed spatiotemporal traffic conditions, which improves estimation accuracy. Similar trends can be observed for car and truck flow estimation on unobserved links, suggesting that more frequent density snapshots enhance estimation accuracy for links lacking local sensor data. Nevertheless, estimation accuracy for unobserved links remains lower than that for observed links, reflecting the underdetermined nature of the DODE problem.

\section{Conclusion}\label{sec:conclusion}
 {This study proposes a computational graph-based framework to estimate dynamic origin-destination demand (DODE) by integrating multi-modal data, including emerging high-resolution satellite imagery and conventional traffic data collected from ground-based localized sensors. Satellite imagery data provides city-wide class-specific traffic density information of both on-road and parking vehicles, which can be extracted using computer vision techniques and used as an additional data source in DODE, overcoming the limitations of local traffic sensor data. A series of numerical experiments using both synthetic and real-world data on networks of various sizes demonstrate that incorporating satellite imagery-derived density improves the fit of link density and recovers more plausible dynamic OD patterns, while maintaining satisfactory accuracy for local traffic counts and speeds, especially for segments without local sensor coverage. In the real-world benchmarking, our computational-graph framework also shows faster and more stable convergence and achieves better state estimation accuracy than the PC-SPSA benchmark under the same observation setting, suggesting practical advantages when historical OD patterns are unavailable. We also conduct sensitivity analysis in terms of sensing error and snapshot frequency in imagery-derived density observations to evaluate the robustness of our model to various data availability conditions. The sensitivity analysis results indicate that the proposed framework is relatively insensitive to moderate systematic errors in density sensing, so that even with imperfect class-specific detection, the resulting OD estimates remain satisfactory under realistic levels of noise.}

 {The primary use case of this study is calibration of city-wide dynamic network models, which serve as a testbed that reproduces status quo traffic states to support scenario analysis and policy evaluation. Especially when fixed detectors are sparse, unevenly deployed, or difficult to expand, satellite imagery with broader spatial coverage can be a promising data source used in developing these network models. From the data perspective, satellite images used in this study are infrequent, so our experiment is a proof-of-concept study under limited data availability. 
When more city-wide satellite imagery data are available, more tests can be conducted to evaluate model robustness, especially in large-scale real-world applications. Moreover, the proposed model can also accommodate video-based data, which provides traffic information of higher resolution, but this requires more experiments with only high-frequency traffic density information. From a specific application perspective, satellite images only capture the conditions of curbside spaces and outdoor parking lots. Fusing other parking information, such as indoor parking and parking meter transactions, will enable more comprehensive parking choice modeling. From a data preprocessing perspective, when class-specific information is unreliable or not available, the framework can also accommodate aggregated density information with satisfactory performance, as demonstrated in a preliminary study by \cite{liu2024sat}. The computer vision pipeline can be further enhanced for accurate vehicle classification and road segmentation for map matching, particularly for freight vehicles, and its systematic detection biases can be explicitly corrected to further reduce potential bias in truck OD estimation.}

\section*{Acknowledgment}
This research is sponsored by Fujitsu Research of America. ChatGPT 5.2 was
used to assist in improving the clarity, conciseness, and overall quality of the writing in this manuscript. After using this tool, the authors reviewed and edited the content as needed and take full responsibility for the content of this manuscript.

\newpage
\bibliography{elsarticle-template}

@article{cantelmo2018utility,
  title={A utility-based dynamic demand estimation model that explicitly accounts for activity scheduling and duration},
  author={Cantelmo, Guido and Viti, Francesco and Cipriani, Ernesto and Nigro, Marialisa},
  journal={Transportation Research Part A: Policy and Practice},
  volume={114},
  pages={303--320},
  year={2018},
  publisher={Elsevier}
}

@article{cascetta2013quasi,
  title={Quasi-dynamic estimation of o--d flows from traffic counts: Formulation, statistical validation and performance analysis on real data},
  author={Cascetta, Ennio and Papola, Andrea and Marzano, Vittorio and Simonelli, Fulvio and Vitiello, Iolanda},
  journal={Transportation Research Part B: Methodological},
  volume={55},
  pages={171--187},
  year={2013},
  publisher={Elsevier}
}

@article{krishnakumari2020data,
  title={A data driven method for OD matrix estimation},
  author={Krishnakumari, Panchamy and Van Lint, Hans and Djukic, Tamara and Cats, Oded},
  journal={Transportation Research Part C: Emerging Technologies},
  volume={113},
  pages={38--56},
  year={2020},
  publisher={Elsevier}
}

@article{sanandaji2016compressive,
  title={Compressive origin-destination estimation},
  author={Sanandaji, Borhan M and Varaiya, Pravin},
  journal={Transportation Letters},
  volume={8},
  number={3},
  pages={148--157},
  year={2016},
  publisher={Taylor \& Francis}
}

@article{djukic2012application,
  title={Application of principal component analysis to predict dynamic origin--destination matrices},
  author={Djukic, Tamara and Van Lint, JWC and Hoogendoorn, SP},
  journal={Transportation research record},
  volume={2283},
  number={1},
  pages={81--89},
  year={2012},
  publisher={SAGE Publications Sage CA: Los Angeles, CA}
}

@inproceedings{djukic2012efficient,
  title={Efficient real time OD matrix estimation based on Principal Component Analysis},
  author={Djukic, Tamara and Fl{\"o}tter{\"o}d, Gunnar and Van Lint, Hans and Hoogendoorn, Serge},
  booktitle={2012 15th International IEEE Conference on Intelligent Transportation Systems},
  pages={115--121},
  year={2012},
  organization={IEEE}
}

@article{prakash2017reducing,
  title={Reducing the dimension of online calibration in dynamic traffic assignment systems},
  author={Prakash, A Arun and Seshadri, Ravi and Antoniou, Constantinos and Pereira, Francisco C and Ben-Akiva, Moshe E},
  journal={Transportation Research Record},
  volume={2667},
  number={1},
  pages={96--107},
  year={2017},
  publisher={SAGE Publications Sage CA: Los Angeles, CA}
}

@article{prakash2018improving,
  title={Improving scalability of generic online calibration for real-time dynamic traffic assignment systems},
  author={Prakash, A Arun and Seshadri, Ravi and Antoniou, Constantinos and Pereira, Francisco C and Ben-Akiva, Moshe},
  journal={Transportation Research Record},
  volume={2672},
  number={48},
  pages={79--92},
  year={2018},
  publisher={SAGE Publications Sage CA: Los Angeles, CA}
}

@article{qurashi2019pc,
  title={PC--SPSA: Employing dimensionality reduction to limit SPSA search noise in DTA model calibration},
  author={Qurashi, Moeid and Ma, Tao and Chaniotakis, Emmanouil and Antoniou, Constantinos},
  journal={IEEE Transactions on Intelligent Transportation Systems},
  volume={21},
  number={4},
  pages={1635--1645},
  year={2019},
  publisher={IEEE}
}

@article{qurashi2022dynamic,
  title={Dynamic demand estimation on large scale networks using principal component analysis: The case of non-existent or irrelevant historical estimates},
  author={Qurashi, Moeid and Lu, Qing-Long and Cantelmo, Guido and Antoniou, Constantinos},
  journal={Transportation Research Part C: Emerging Technologies},
  volume={136},
  pages={103504},
  year={2022},
  publisher={Elsevier}
}

@article{yang1991analysis,
  title={An analysis of the reliability of an origin-destination trip matrix estimated from traffic counts},
  author={Yang, Hai and Iida, Yasunori and Sasaki, Tsuna},
  journal={Transportation Research Part B: Methodological},
  volume={25},
  number={5},
  pages={351--363},
  year={1991},
  publisher={Elsevier}
}

@article{zhou2025flow,
  title={Flow-through tensors: A unified computational graph architecture for multi-layer transportation network optimization},
  author={Zhou, Xuesong Simon and Kim, Taehooie and Ameli, Mostafa and Zhu, Henan Bety and Honma, Yudai and Pendyala, Ram M},
  journal={Artificial Intelligence for Transportation},
  volume={1},
  pages={100006},
  year={2025},
  publisher={Elsevier}
}

@article{xiong2025multi,
  title={Multi-Source Urban Traffic Flow Forecasting with Drone and Loop Detector Data},
  author={Xiong, Weijiang and Fonod, Robert and Alahi, Alexandre and Geroliminis, Nikolas},
  journal={arXiv preprint arXiv:2501.03492},
  year={2025}
}

@article{espadaler2025accurate,
  title={An accurate safety and congestion monitoring framework with a swarm of drones},
  author={Espadaler-Clap{\'e}s, Jasso and Fonod, Robert and Barmpounakis, Emmanouil and Geroliminis, Nikolas},
  journal={Transportation Research Interdisciplinary Perspectives},
  volume={32},
  pages={101490},
  year={2025},
  publisher={Elsevier}
}

@article{fonod2025advanced,
  title={Advanced computer vision for extracting georeferenced vehicle trajectories from drone imagery},
  author={Fonod, Robert and Cho, Haechan and Yeo, Hwasoo and Geroliminis, Nikolas},
  journal={Transportation Research Part C: Emerging Technologies},
  volume={178},
  pages={105205},
  year={2025},
  publisher={Elsevier}
}

@article{wu2024participatory,
  title={Participatory traffic control: Leveraging connected and automated vehicles to enhance network efficiency},
  author={Wu, Minghui and Wang, Ben and Yin, Yafeng and Lynch, Jerome P},
  journal={Transportation Research Part C: Emerging Technologies},
  volume={166},
  pages={104757},
  year={2024},
  publisher={Elsevier}
}

@article{wu2025multiday,
  title={Multiday User Equilibrium with Strategic Commuters},
  author={Wu, Minghui and Yin, Yafeng and Lynch, Jerome P},
  journal={Transportation Science},
  volume={59},
  number={2},
  pages={413--432},
  year={2025},
  publisher={INFORMS}
}

@inproceedings{kawamura2025rn,
  title={RN-Sam: Road Network-Aided Sam Optimization for Road Segmentation In Satellite Imagery},
  author={Kawamura, Ryosuke and Guarda, Pablo and Narwade, Pradeep and Patel, Yash and Niinuma, Koichiro},
  booktitle={2025 IEEE International Conference on Image Processing (ICIP)},
  pages={1318--1323},
  year={2025},
  organization={IEEE}
}

@incollection{ye2019gating,
  title={Gating Control for Congested Network with Multiple Sub-Regions and Dynamic Boundaries},
  author={Ye, Hanjun and Liao, Nannan and Liu, Jiachao and An, Chengchuan and Xia, Jingxin},
  publisher={CICTP 2019},
  pages={2850--2861},
  year={2019}
}

@article{he2024efficient,
  title={Efficient and Robust Freeway Traffic Speed Estimation Under Oblique Grid Using Vehicle Trajectory Data},
  author={He, Yang and An, Chengchuan and Jia, Yuheng and Liu, Jiachao and Lu, Zhenbo and Xia, Jingxin},
  journal={IEEE Transactions on Intelligent Transportation Systems},
  year={2024},
  publisher={IEEE}
}

@article{liu2023optimal,
  title={Optimal curbside pricing for managing ride-hailing pick-ups and drop-offs},
  author={Liu, Jiachao and Ma, Wei and Qian, Sean},
  journal={Transportation Research Part C: Emerging Technologies},
  volume={146},
  pages={103960},
  year={2023},
  publisher={Elsevier}
}

@article{wu2018hierarchical,
  title={Hierarchical travel demand estimation using multiple data sources: A forward and backward propagation algorithmic framework on a layered computational graph},
  author={Wu, Xin and Guo, Jifu and Xian, Kai and Zhou, Xuesong},
  journal={Transportation Research Part C: Emerging Technologies},
  volume={96},
  pages={321--346},
  year={2018},
  publisher={Elsevier}
}

@article{Ma2020,
  title = {Estimating multi-class dynamic origin-destination demand through a forward-backward algorithm on computational graphs},
  volume = {119},
  issn = {0968090X},
  url = {https://doi.org/10.1016/j.trc.2020.102747},
  doi = {10.1016/j.trc.2020.102747},
  number = {August},
  journal = {Transportation Research Part C: Emerging Technologies},
  author = {Ma, Wei and Pi, Xidong and Qian, Sean},
  year = {2020},
  note = {arXiv: 1903.04681
Publisher: Elsevier Ltd},
  pages = {102747},
}

@article{zhu2016optimal,
  title={Optimal heterogeneous sensor deployment strategy for dynamic origin--destination demand estimation},
  author={Zhu, Senlai and Zheng, Hong and Peeta, Srinivas and Guo, Yuntao and Cheng, Lin and Sun, Wei},
  journal={Transportation Research Record},
  volume={2567},
  number={1},
  pages={18--27},
  year={2016},
  publisher={SAGE Publications Sage CA: Los Angeles, CA}
}

@article{zhu2017integrating,
  title={Integrating optimal heterogeneous sensor deployment and operation strategies for dynamic origin-destination demand estimation},
  author={Zhu, Senlai and Guo, Yuntao and Chen, Jingxu and Li, Dawei and Cheng, Lin},
  journal={Sensors},
  volume={17},
  number={8},
  pages={1767},
  year={2017},
  publisher={MDPI}
}

@article{zhou2010information,
  title={An information-theoretic sensor location model for traffic origin-destination demand estimation applications},
  author={Zhou, Xuesong and List, George F},
  journal={Transportation Science},
  volume={44},
  number={2},
  pages={254--273},
  year={2010},
  publisher={INFORMS}
}

@article{salari2021modeling,
  title={Modeling the effect of sensor failure on the location of counting sensors for origin-destination (OD) estimation},
  author={Salari, Mostafa and Kattan, Lina and Lam, William HK and Esfeh, Mohammad Ansari and Fu, Hao},
  journal={Transportation Research Part C: Emerging Technologies},
  volume={132},
  pages={103367},
  year={2021},
  publisher={Elsevier}
}

@article{li2023submodularity,
  title={Submodularity of optimal sensor placement for traffic networks},
  author={Li, Ruolin and Mehr, Negar and Horowitz, Roberto},
  journal={Transportation research part B: methodological},
  volume={171},
  pages={29--43},
  year={2023},
  publisher={Elsevier}
}

@article{hu2024sensor,
  title={Sensor placement considering the observability of traffic dynamics: On the algebraic and graphical perspectives},
  author={Hu, Xinyue and Fan, Yueyue},
  journal={Transportation Research Part B: Methodological},
  volume={189},
  pages={103057},
  year={2024},
  publisher={Elsevier}
}

@article{leitloff2010vehicle,
  title={Vehicle detection in very high resolution satellite images of city areas},
  author={Leitloff, Jens and Hinz, Stefan and Stilla, Uwe},
  journal={IEEE transactions on Geoscience and remote sensing},
  volume={48},
  number={7},
  pages={2795--2806},
  year={2010},
  publisher={IEEE}
}

@article{chen2014vehicle,
  title={Vehicle detection in satellite images by hybrid deep convolutional neural networks},
  author={Chen, Xueyun and Xiang, Shiming and Liu, Cheng-Lin and Pan, Chun-Hong},
  journal={IEEE Geoscience and remote sensing letters},
  volume={11},
  number={10},
  pages={1797--1801},
  year={2014},
  publisher={IEEE}
}

@article{cao2016vehicle,
  title={Vehicle detection from highway satellite images via transfer learning},
  author={Cao, Liujuan and Wang, Cheng and Li, Jonathan},
  journal={Information sciences},
  volume={366},
  pages={177--187},
  year={2016},
  publisher={Elsevier}
}

@article{dai2020road,
  title={Road extraction from high-resolution satellite images based on multiple descriptors},
  author={Dai, Jiguang and Zhu, Tingting and Wang, Yang and Ma, Rongchen and Fang, Xinxin},
  journal={IEEE Journal of Selected Topics in Applied Earth Observations and Remote Sensing},
  volume={13},
  pages={227--240},
  year={2020},
  publisher={IEEE}
}

@article{mei2021coanet,
  title={CoANet: Connectivity attention network for road extraction from satellite imagery},
  author={Mei, Jie and Li, Rou-Jing and Gao, Wang and Cheng, Ming-Ming},
  journal={IEEE Transactions on Image Processing},
  volume={30},
  pages={8540--8552},
  year={2021},
  publisher={IEEE}
}

@article{toledo2012estimation,
  title={Estimation of dynamic origin--destination matrices using linear assignment matrix approximations},
  author={Toledo, Tomer and Kolechkina, Tanya},
  journal={IEEE Transactions on Intelligent Transportation Systems},
  volume={14},
  number={2},
  pages={618--626},
  year={2012},
  publisher={IEEE}
}

@article{lu2013dynamic,
  title={Dynamic origin--destination demand flow estimation under congested traffic conditions},
  author={Lu, Chung-Cheng and Zhou, Xuesong and Zhang, Kuilin},
  journal={Transportation Research Part C: Emerging Technologies},
  volume={34},
  pages={16--37},
  year={2013},
  publisher={Elsevier}
}

@article{kumarage2023hybrid,
  title={A hybrid modelling framework for the estimation of dynamic origin--destination flows},
  author={Kumarage, Sakitha and Yildirimoglu, Mehmet and Zheng, Zuduo},
  journal={Transportation Research Part B: Methodological},
  volume={176},
  pages={102804},
  year={2023},
  publisher={Elsevier}
}

@article{sun2024stochastic,
  title={Stochastic OD demand estimation using stochastic programming},
  author={Sun, Ran and Fan, Yueyue},
  journal={Transportation Research Part B: Methodological},
  volume={183},
  pages={102943},
  year={2024},
  publisher={Elsevier}
}

@article{ou2019learn,
  title={Learn, assign, and search: real-time estimation of dynamic origin-destination flows using machine learning algorithms},
  author={Ou, Jishun and Lu, Jiawei and Xia, Jingxin and An, Chengchuan and Lu, Zhenbo},
  journal={IEEE Access},
  volume={7},
  pages={26967--26983},
  year={2019},
  publisher={IEEE}
}

@article{lu2015kalman,
  title={A Kalman filter approach to dynamic OD flow estimation for urban road networks using multi-sensor data},
  author={Lu, Zhenbo and Rao, Wenming and Wu, Yao-Jan and Guo, Li and Xia, Jingxin},
  journal={Journal of Advanced Transportation},
  volume={49},
  number={2},
  pages={210--227},
  year={2015},
  publisher={Wiley Online Library}
}

@article{huo2023simulation,
  title={Simulation-based dynamic origin--destination matrix estimation on freeways: A Bayesian optimization approach},
  author={Huo, Jinbiao and Liu, Chengqi and Chen, Jingxu and Meng, Qiang and Wang, Jian and Liu, Zhiyuan},
  journal={Transportation Research Part E: Logistics and Transportation Review},
  volume={173},
  pages={103108},
  year={2023},
  publisher={Elsevier}
}

@article{Cao02092024,
author = {Yumin Cao, Jiarong Yao, Keshuang Tang and Qi Kang},
title = {Dynamic origin–destination flow estimation for urban road network solely using probe vehicle trajectory data},
journal = {Journal of Intelligent Transportation Systems},
volume = {28},
number = {5},
pages = {756--773},
year = {2024},
publisher = {Taylor \& Francis},
doi = {10.1080/15472450.2023.2209910}
}

@article{tang2021dynamic,
  title={Dynamic origin-destination flow estimation using automatic vehicle identification data: A 3D convolutional neural network approach},
  author={Tang, Keshuang and Cao, Yumin and Chen, Can and Yao, Jiarong and Tan, Chaopeng and Sun, Jian},
  journal={Computer-Aided Civil and Infrastructure Engineering},
  volume={36},
  number={1},
  pages={30--46},
  year={2021},
  publisher={Wiley Online Library}
}

@article{zhou2006dynamic,
  title={Dynamic origin-destination demand estimation using automatic vehicle identification data},
  author={Zhou, Xuesong and Mahmassani, Hani S},
  journal={IEEE Transactions on intelligent transportation systems},
  volume={7},
  number={1},
  pages={105--114},
  year={2006},
  publisher={IEEE}
}

@article{wang2013estimating,
  title={Estimating dynamic origin-destination data and travel demand using cell phone network data},
  author={Wang, Ming-Heng and Schrock, Steven D and Vander Broek, Nate and Mulinazzi, Thomas},
  journal={International Journal of Intelligent Transportation Systems Research},
  volume={11},
  pages={76--86},
  year={2013},
  publisher={Springer}
}

@article{lu2024two,
  title={A two-stage stochastic programming approach for dynamic OD estimation using LBSN data},
  author={Lu, Qing-Long and Qurashi, Moeid and Antoniou, Constantinos},
  journal={Transportation Research Part C: Emerging Technologies},
  volume={158},
  pages={104460},
  year={2024},
  publisher={Elsevier}
}

@article{dixon2002real,
  title={Real-time OD estimation using automatic vehicle identification and traffic count data},
  author={Dixon, Michael P and Rilett, Laurence R},
  journal={Computer-Aided Civil and Infrastructure Engineering},
  volume={17},
  number={1},
  pages={7--21},
  year={2002},
  publisher={Wiley Online Library}
}

@article{cao2021day,
  title={Day-to-day dynamic origin--destination flow estimation using connected vehicle trajectories and automatic vehicle identification data},
  author={Cao, Yumin and Tang, Keshuang and Sun, Jian and Ji, Yangbeibei},
  journal={Transportation Research Part C: Emerging Technologies},
  volume={129},
  pages={103241},
  year={2021},
  publisher={Elsevier}
}

@article{ros2022practical,
  title={A practical approach to assignment-free dynamic origin--destination matrix estimation problem},
  author={Ros-Roca, Xavier and Montero, L{\'\i}dia and Barcel{\'o}, Jaume and N{\"o}kel, Klaus and Gentile, Guido},
  journal={Transportation Research Part C: Emerging Technologies},
  volume={134},
  pages={103477},
  year={2022},
  publisher={Elsevier}
}

@INPROCEEDINGS{Frederix2010_density,
  author={Frederix, Rodric and Viti, Francesco and Tampère, Chris M.J.},
  booktitle={13th International IEEE Conference on Intelligent Transportation Systems}, 
  title={A density-based dynamic OD estimation method that reproduces within-day congestion dynamics}, 
  year={2010},
  volume={},
  number={},
  pages={694-699},
  doi={10.1109/ITSC.2010.5625220}
}

@article{mo2020_LPR,
author = {Mo, Baichuan and Li, Ruimin and Dai, Jingchen},
title = {Estimating dynamic origin–destination demand: A hybrid framework using license plate recognition data},
journal = {Computer-Aided Civil and Infrastructure Engineering},
volume = {35},
number = {7},
pages = {734-752},
doi = {https://doi.org/10.1111/mice.12526},
url = {https://onlinelibrary.wiley.com/doi/abs/10.1111/mice.12526},
eprint = {https://onlinelibrary.wiley.com/doi/pdf/10.1111/mice.12526},
year = {2020}
}

@article{CARRESE201783,
title = {Dynamic demand estimation and prediction for traffic urban networks adopting new data sources},
journal = {Transportation Research Part C: Emerging Technologies},
volume = {81},
pages = {83-98},
year = {2017},
issn = {0968-090X},
doi = {https://doi.org/10.1016/j.trc.2017.05.013},
url = {https://www.sciencedirect.com/science/article/pii/S0968090X17301432},
author = {Stefano Carrese and Ernesto Cipriani and Livia Mannini and Marialisa Nigro}
}

@article{kim2024computational,
  title={Computational graph-based mathematical programming reformulation for integrated demand and supply models},
  author={Kim, Taehooie and Lu, Jiawei and Pendyala, Ram M and Zhou, Xuesong Simon},
  journal={Transportation Research Part C: Emerging Technologies},
  volume={164},
  pages={104671},
  year={2024},
  publisher={Elsevier}
}

@article{lu2023physics,
  title={Physics-informed neural networks for integrated traffic state and queue profile estimation: A differentiable programming approach on layered computational graphs},
  author={Lu, Jiawei and Li, Chongnan and Wu, Xin Bruce and Zhou, Xuesong Simon},
  journal={Transportation Research Part C: Emerging Technologies},
  volume={153},
  pages={104224},
  year={2023},
  publisher={Elsevier}
}

@article{guarda_estimating_2024,
  title = {Estimating network flow and travel behavior using day-to-day system-level data: {A} computational graph approach},
  volume = {158},
  copyright = {All rights reserved},
  issn = {0968090X},
  shorttitle = {Estimating network flow and travel behavior using day-to-day system-level data},
  url = {https://linkinghub.elsevier.com/retrieve/pii/S0968090X23003996},
  doi = {10.1016/j.trc.2023.104409},
  urldate = {2023-12-23},
  journal = {Transportation Research Part C: Emerging Technologies},
  author = {Guarda, Pablo and Battifarano, Matthew and Qian, Sean},
  month = jan,
  year = {2024},
  pages = {104409},
}

@misc{guarda_traffic_2024,
  title = {Traffic estimation in unobserved network locations using data-driven macroscopic models},
  copyright = {All rights reserved},
  url = {http://arxiv.org/abs/2401.17095},
  urldate = {2024-02-19},
  publisher = {arXiv},
  author = {Guarda, Pablo and Qian, Sean},
  month = jan,
  year = {2024},
  note = {arXiv:2401.17095 [cs]},
}

@article{liu2024modeling,
  title={Modeling Multimodal Curbside Usage in Dynamic Networks},
  author={Liu, Jiachao and Qian, Sean},
  journal={Transportation Science},
  year={2024},
  publisher={INFORMS}
}

@article{liu2023end,
  title={End-to-end learning of user equilibrium with implicit neural networks},
  author={Liu, Zhichen and Yin, Yafeng and Bai, Fan and Grimm, Donald K},
  journal={Transportation Research Part C: Emerging Technologies},
  volume={150},
  pages={104085},
  year={2023},
  publisher={Elsevier}
}

@article{liu2025end,
  title={End-to-End Learning of User Equilibrium: Expressivity, Generalization, and Optimization},
  author={Liu, Zhichen and Yin, Yafeng},
  journal={Transportation Science},
  year={2025},
  publisher={INFORMS}
}

@article{du2025modeling,
  title={Modeling Metro Passenger Routing Choices with a Fully Differentiable End-to-End Simulation-Based Optimization (SBO) Approach},
  author={Du, Kejun and Lee, Enoch and Ma, Qiru and Su, Zhiya and Zhang, Shuyang and Lo, Hong K},
  journal={Transportation Science},
  year={2025},
  publisher={INFORMS}
}

@article{sheehan2023city,
  title={City scale traffic monitoring using worldview satellite imagery and deep learning: a case study of Barcelona},
  author={Sheehan, Annalisa and Beddows, Andrew and Green, David C and Beevers, Sean},
  journal={Remote Sensing},
  volume={15},
  number={24},
  pages={5709},
  year={2023},
  publisher={MDPI}
}

@article{mccord2003estimating,
  title={Estimating annual average daily traffic from satellite imagery and air photos: Empirical results},
  author={McCord, Mark R and Yang, Yongliang and Jiang, Zhuojun and Coifman, Benjamin and Goel, Prem K},
  journal={Transportation Research Record},
  volume={1855},
  number={1},
  pages={136--142},
  year={2003},
  publisher={SAGE Publications Sage CA: Los Angeles, CA}
}

@article{larsen2009traffic,
  title={Traffic monitoring using very high resolution satellite imagery},
  author={Larsen, Siri {\O}yen and Koren, Hans and Solberg, Rune},
  journal={Photogrammetric Engineering \& Remote Sensing},
  volume={75},
  number={7},
  pages={859--869},
  year={2009},
  publisher={American Society for Photogrammetry and Remote Sensing}
}

@article{larsen2013automatic,
  title={Automatic system for operational traffic monitoring using very-high-resolution satellite imagery},
  author={Larsen, Siri {\O}yen and Salberg, Arnt-B{\o}rre and Eikvil, Line},
  journal={International Journal of Remote Sensing},
  volume={34},
  number={13},
  pages={4850--4870},
  year={2013},
  publisher={Taylor \& Francis}
}

@inproceedings{kaack2019truck,
  title={Truck traffic monitoring with satellite images},
  author={Kaack, Lynn H and Chen, George H and Morgan, M Granger},
  booktitle={Proceedings of the 2nd ACM SIGCAS Conference on Computing and Sustainable Societies},
  pages={155--164},
  year={2019}
}

@article{balamuralidhar2021multeye,
  title={MultEYE: Monitoring system for real-time vehicle detection, tracking and speed estimation from UAV imagery on edge-computing platforms},
  author={Balamuralidhar, Navaneeth and Tilon, Sofia and Nex, Francesco},
  journal={Remote sensing},
  volume={13},
  number={4},
  pages={573},
  year={2021},
  publisher={MDPI}
}

@article{butilua2022urban,
  title={Urban traffic monitoring and analysis using unmanned aerial vehicles (UAVs): A systematic literature review},
  author={Butil{\u{a}}, Eugen Valentin and Boboc, R{\u{a}}zvan Gabriel},
  journal={Remote Sensing},
  volume={14},
  number={3},
  pages={620},
  year={2022},
  publisher={MDPI}
}

@article{bai2025dynamic,
  title={A Dynamic Unmanned Aerial Vehicle Routing Framework for Urban Traffic Monitoring},
  author={Bai, Yumeng and Feng, Yiheng},
  journal={arXiv preprint arXiv:2501.09249},
  year={2025}
}

@inproceedings{kopsiaftis2015vehicle,
  title={Vehicle detection and traffic density monitoring from very high resolution satellite video data},
  author={Kopsiaftis, George and Karantzalos, Konstantinos},
  booktitle={2015 IEEE international geoscience and remote sensing symposium (IGARSS)},
  pages={1881--1884},
  year={2015},
  organization={IEEE}
}

@inproceedings{sakai2019traffic,
  title={Traffic density estimation method from small satellite imagery: Towards frequent remote sensing of car traffic},
  author={Sakai, Kengo and Seo, Toru and Fuse, Takashi},
  booktitle={2019 IEEE Intelligent Transportation Systems Conference (ITSC)},
  pages={1776--1781},
  year={2019},
  organization={IEEE}
}

@article{ganji_traffic_2022,
  title = {Traffic volume prediction using aerial imagery and sparse data from road counts},
  volume = {141},
  issn = {0968090X},
  url = {https://linkinghub.elsevier.com/retrieve/pii/S0968090X22001735},
  doi = {10.1016/j.trc.2022.103739},
  language = {en},
  urldate = {2023-11-03},
  journal = {Transportation Research Part C: Emerging Technologies},
  author = {Ganji, Arman and Zhang, Mingqian and Hatzopoulou, Marianne},
  month = aug,
  year = {2022},
  pages = {103739},
}

@article{chen2021spatial,
  title={Spatial temporal analysis of traffic patterns during the COVID-19 epidemic by vehicle detection using planet remote-sensing satellite images},
  author={Chen, Yulu and Qin, Rongjun and Zhang, Guixiang and Albanwan, Hessah},
  journal={Remote Sensing},
  volume={13},
  number={2},
  pages={208},
  year={2021},
  publisher={MDPI}
}

@inproceedings{he2021inferring,
  title={Inferring high-resolution traffic accident risk maps based on satellite imagery and GPS trajectories},
  author={He, Songtao and Sadeghi, Mohammad Amin and Chawla, Sanjay and Alizadeh, Mohammad and Balakrishnan, Hari and Madden, Samuel},
  booktitle={Proceedings of the IEEE/CVF International Conference on Computer Vision},
  pages={11977--11985},
  year={2021}
}

@article{yu2024sky,
  title={From sky to road: incorporating the satellite imagery into analysis of freight truck-related crash factors},
  author={Yu, Chengcheng and Hua, Wei and Yang, Chao and Fang, Shen and Li, Yuanhe and Yuan, Quan},
  journal={Accident Analysis \& Prevention},
  volume={200},
  pages={107491},
  year={2024},
  publisher={Elsevier}
}

@article{horsler2023predicting,
  title={Predicting Regional Road Transport Emissions From Satellite Imagery},
  author={Horsler, Adam and Baker, Jake and others},
  journal={arXiv preprint arXiv:2312.10551},
  year={2023}
}

@inproceedings{nachmany_detecting_2019,
  title = {Detecting roads from satellite imagery in the developing world},
  booktitle = {Proceedings of the {IEEE}/{CVF} {Conference} on {Computer} {Vision} and {Pattern} {Recognition} {Workshops}},
  author = {Nachmany, Yoni and Alemohammad, Hamed},
  year = {2019},
  pages = {83--89},
}

@misc{azaveaelement_84_robert_cheetham_raster_nodate,
  title = {Raster {Vision}: {An} open source library and framework for deep learning on satellite and aerial imagery (2017-2023).},
  url = {https://github.com/azavea/raster-vision},
  author = {{Azavea/Element 84, Robert Cheetham}},
  doi = {10.5281/zenodo.8018177},
}

@inproceedings{liu_ssd_2016,
  title = {Ssd: {Single} shot multibox detector},
  isbn = {3-319-46447-7},
  booktitle = {Computer {Vision}–{ECCV} 2016: 14th {European} {Conference}, {Amsterdam}, {The} {Netherlands}, {October} 11–14, 2016, {Proceedings}, {Part} {I} 14},
  publisher = {Springer},
  author = {Liu, Wei and Anguelov, Dragomir and Erhan, Dumitru and Szegedy, Christian and Reed, Scott and Fu, Cheng-Yang and Berg, Alexander C},
  year = {2016},
  pages = {21--37},
}

@article{simonyan_very_2014,
  title = {Very deep convolutional networks for large-scale image recognition},
  journal = {arXiv preprint arXiv:1409.1556},
  author = {Simonyan, Karen and Zisserman, Andrew},
  year = {2014},
}

@article{ren_faster_2017,
  title = {Faster {R}-{CNN}: {Towards} {Real}-{Time} {Object} {Detection} with {Region} {Proposal} {Networks}},
  volume = {39},
  copyright = {https://ieeexplore.ieee.org/Xplorehelp/downloads/license-information/IEEE.html},
  issn = {0162-8828, 2160-9292},
  shorttitle = {Faster {R}-{CNN}},
  url = {http://ieeexplore.ieee.org/document/7485869/},
  doi = {10.1109/TPAMI.2016.2577031},
  language = {en},
  number = {6},
  urldate = {2024-04-29},
  journal = {IEEE Transactions on Pattern Analysis and Machine Intelligence},
  author = {Ren, Shaoqing and He, Kaiming and Girshick, Ross and Sun, Jian},
  month = jun,
  year = {2017},
  pages = {1137--1149},
}

@inproceedings{he_deep_2016,
  title = {Deep residual learning for image recognition},
  booktitle = {Proceedings of the {IEEE} conference on computer vision and pattern recognition},
  author = {He, Kaiming and Zhang, Xiangyu and Ren, Shaoqing and Sun, Jian},
  year = {2016},
  pages = {770--778},
}

@article{van_etten_you_2018,
  title = {You only look twice: {Rapid} multi-scale object detection in satellite imagery},
  journal = {arXiv preprint arXiv:1805.09512},
  author = {Van Etten, Adam},
  year = {2018},
}

@article{lam_xview_2018,
  title = {xview: {Objects} in context in overhead imagery},
  journal = {arXiv preprint arXiv:1802.07856},
  author = {Lam, Darius and Kuzma, Richard and McGee, Kevin and Dooley, Samuel and Laielli, Michael and Klaric, Matthew and Bulatov, Yaroslav and McCord, Brendan},
  year = {2018},
}

@article{pi2019general,
  title={A general formulation for multi-modal dynamic traffic assignment considering multi-class vehicles, public transit and parking},
  author={Pi, Xidong and Ma, Wei and Qian, Zhen Sean},
  journal={Transportation Research Part C: Emerging Technologies},
  volume={104},
  pages={369--389},
  year={2019},
  publisher={Elsevier}
}

@techreport{pi2018regional,
  title={Regional traffic planning and operation using mesoscopic car-truck traffic simulation},
  author={Pi, X and Ma, W and Qian, S},
  year={2018},
  institution={Technical report for Traffic21 Institute}
}

@article{qian2013hybrid,
  title={A hybrid route choice model for dynamic traffic assignment},
  author={Qian, Zhen and Zhang, H Michael},
  journal={Networks and Spatial Economics},
  volume={13},
  pages={183--203},
  year={2013},
  publisher={Springer}
}

@article{zhang2020path,
  title={Path-based system optimal dynamic traffic assignment: A subgradient approach},
  author={Zhang, Pinchao and Qian, Sean},
  journal={Transportation Research Part B: Methodological},
  volume={134},
  pages={41--63},
  year={2020},
  publisher={Elsevier}
}

@inproceedings{liu2024sat,
  author={Liu, Jiachao and Guarda, Pablo and Niinuma, Koichiro and Qian, Sean},
  booktitle={2024 IEEE 27th International Conference on Intelligent Transportation Systems (ITSC)}, 
  title={Enhancing Multi-Class Mesoscopic Network Modeling with High-Resolution Satellite Imagery}, 
  year={2024},
  volume={},
  number={},
  pages={733-740},
  doi={10.1109/ITSC58415.2024.10919613}}

@article{gagliardi2023satellite,
  title={Satellite remote sensing and non-destructive testing methods for transport infrastructure monitoring: Advances, challenges and perspectives},
  author={Gagliardi, Valerio and Tosti, Fabio and Bianchini Ciampoli, Luca and Battagliere, Maria Libera and D’Amato, Luigi and Alani, Amir M and Benedetto, Andrea},
  journal={Remote Sensing},
  volume={15},
  number={2},
  pages={418},
  year={2023},
  publisher={MDPI}
}

@article{reksten_estimating_2021,
  title = {Estimating {Traffic} in {Urban} {Areas} from {Very}-{High} {Resolution} {Aerial} {Images}},
  volume = {42},
  issn = {0143-1161, 1366-5901},
  url = {https://www.tandfonline.com/doi/full/10.1080/01431161.2020.1815891},
  doi = {10.1080/01431161.2020.1815891},
  language = {en},
  number = {3},
  urldate = {2023-11-06},
  journal = {International Journal of Remote Sensing},
  author = {Reksten, Jarle Hamar and Salberg, Arnt-Børre},
  month = feb,
  year = {2021},
  pages = {865--883},
}

@article{wang_transportation_2022,
    title = {Transportation {Origin}-{Destination} {Demand} {Estimation} with {Quasi}-{Sparsity}},
    issn = {0041-1655, 1526-5447},
    url = {http://pubsonline.informs.org/doi/10.1287/trsc.2022.1178},
    doi = {10.1287/trsc.2022.1178},
    language = {en},
    urldate = {2023-01-24},
    journal = {Transportation Science},
    author = {Wang, Jingxing and Lu, Shu and Liu, Hongsheng and Ban, Xuegang (Jeff)},
    month = oct,
    year = {2022},
    pages = {trsc.2022.1178},
}

\section*{Appendix}
\begin{table}[H]
\scriptsize
    \centering
    \begin{tabular}{ll}
       \toprule
        Notation & Description \\
        \midrule
        $G$ & road network\\
        $N$ & node set\\
        $A$ & link set\\
        $A_p$ & link set allowed for on-street parking\\
        $T_d$ & assignment time horizon\\
        $R$ & origin node set\\
        $S$ & destination node set\\
        $q^{rs}_{i,t}$ & demand of vehicle class $i$ departing at time $t$ from $r$ and heading to $s$\\
        $\mathcal{C}$ & car class\\
        $\mathcal{T}$ & truck class\\
        $P^{rs}_i$ & path set for vehicle class $i$ between $rs$\\
        $f^{rs}_{k,i,t}$ & flow on path $k$ for vehicle class $i$ departing at time $t$ between OD pair $rs$\\
        $c^{rs}_{k,i,t}$ & travel cost of path $k$ for vehicle class $i$ departing at time $t$ between OD pair $rs$\\
        $p^{rs}_{k,i,t}$ & route choice proportion of choosing path $k$ for vehicle class $i$ departing at time $t$ between OD pair $rs$\\
        $h^{t}_{a,i}$ & travel time of link $a$ for vehicle class $i$ at time $t$\\
        $\Psi$ & route choice function\\
        $\mathcal{A}^m_{a,i}$ & arrival curve of link $a$ for moving vehicles of class $i$\\
        $\mathcal{A}^p_{a,i}$ & arrival curve of link $a$ for parking vehicles of class $i$\\
        $\mathcal{D}^m_{a,i}$ & departure curve of link $a$ for moving vehicles of class $i$\\
        $\mathcal{D}^p_{a,i}$ & departure curve of link $a$ for parking vehicles of class $i$\\
        $x^t_{\text{arr},a,i}$ & total arrival flow on link $a$ for vehicle class $i$ at time $t$\\
        $x^{t,m}_{\text{arr},a,i}$ & arrival flow on link $a$ for moving vehicle of class $i$ at time $t$\\
        $x^{t,p}_{\text{arr},a,i}$ & arrival flow on link $a$ for parking vehicle of class $i$ at time $t$\\
        $\bar{\Lambda}$ & link dynamic function\\
        $R_{a,i}(t)$ & remaining vehicle number of class $i$ on link $a$ at time $t$\\
        $l_a$ & length of link $a$\\
        $k^t_{a,i}$ & density of link $a$ for vehicle class $i$ at time $t$\\
        $\rho^{rs,m}_{\text{arr},k,a,i}(t_1, t_2)$ & the proportion of $k$-th path flow departing at time $t_1$ that arrives on link $a$ at time $t_2$ recorded in cumulative curves for moving vehicles\\
        $\rho^{rs,p}_{\text{arr},k,a,i}(t_1, t_2)$ & the proportion of $k$-th path flow departing at time $t_1$ that arrives on link $a$ at time $t_2$ recorded in cumulative curves for parking vehicles\\
        $\rho^{rs,m}_{\text{dep},k,a,i}(t_1, t_2)$ & the proportion of $k$-th path flow departing at time $t_1$ that departs from link $a$ at time $t_2$ recorded in cumulative curves for moving vehicles\\
        $\rho^{rs,p}_{\text{dep},k,a,i}(t_1, t_2)$ & the proportion of $k$-th path flow departing at time $t_1$ that departs from link $a$ at time $t_2$ recorded in cumulative curves for parking vehicles\\
        $A_{a,i}(\bar{t})$ & total cumulative arrival flow before timestamp $\bar{t}$ of link $a$ for class $i$\\
        $D_{a,i}(\bar{t})$ & total cumulative departure flow before timestamp $\bar{t}$ of link $a$ for class $i$\\
        $\Lambda$ & dynamic network loading function\\
        $\bm{f}_i$ & vectorized path flow of class $i$\\
        $\bm{k}_i$ & vectorized link density of class $i$\\
        $\bm{q}_i$ & vectorized demand of class $i$\\
        $\bm{l}$ & vectorized link length of class $i$\\
        $\bm{\rho}^m_{a,i}$ & vectorized DAR matrix for arrival moving vehicles of class $i$\\
        $\bm{\rho}^p_{a,i}$ & vectorized DAR matrix for arrival parking vehicles of class $i$\\
        $\bm{\rho}^m_{d,i}$ & vectorized DAR matrix for departure moving vehicles of class $i$\\
        $\bm{\rho}^p_{d,i}$ & vectorized DAR matrix for departure parking vehicles of class $i$\\
        $\bm{x}_i$ & vectorized link flow of class $i$\\
        $\bm{h}_i$ & vectorized link travel time of class $i$\\
        $\bm{p}_i$ & vectorized route choice proportion of class $i$\\
        $\bm{c}_i$ & vectorized travel cost of class $i$\\
        $\bm{L}_i$ & aggregation matrix for link flow observation of class $i$\\
        $\bm{M}_i$ & aggregation matrix for travel time observation of class $i$\\
        $\bm{I}_i$ & aggregation matrix for link density observation of class $i$\\
        $\bm{H}_i$ & aggregation matrix for summing total cumulative arrival and departure flow\\
        $\bm{x}^o$ & observed link flow\\
        $\bm{h}^o$ & observed link travel time\\
        $\bm{k}^o$ & observed link density\\
        \bottomrule
    \end{tabular}
    \caption{Notations}
    \label{tab:notation}
\end{table}

\begin{table}[H]
\small
    \begin{tabular}{llllllll}
    \toprule
        ID & length & car free-flow speed & truck free-flow speed & car capacity & truck capacity & car jam density & truck jam density \\
        \midrule
        1 & 1 & 50 & 40 & 6000 & 3600 & 540 & 240\\
        2 & 0.5 & 50 & 40 & 2000 & 1200 & 90 & 40\\
        3 & 1 & 50 & 40 & 6000 & 3600 & 540 & 240\\
        4 & 0.5 & 50 & 40 & 2000 & 1200 & 180 & 80\\
        5 & 1 & 50 & 40 & 6000 & 3600 & 540 & 240\\
        6 & 0.5 & 50 & 40 & 2000 & 1200 & 90 & 40\\
        7 & 1 & 50 & 40 & 4000 & 2400 & 360 & 160\\
        8 & 0.5 & 50 & 40 & 2000 & 1200 & 90 & 40\\
        9 & 1 & 50 & 40 & 6000 & 3600 & 540 & 240\\
        16 & 2 & 30 & 20 & 2000 & 1200 & 360 & 160\\
        17 & 0.5 & 30 & 20 & 2000 & 1200 & 90 & 40\\
        18 & 2 & 30 & 20 & 2000 & 1200 & 360 & 160\\
        \bottomrule
    \end{tabular}
    \caption{Summary of link properties of small network}
    \label{tab:nie_network}
\end{table}

\begin{table}[H]
\small
    \centering
    \begin{tabular}{ll}
       \toprule
        Name & Value \\
        \midrule
        Study period & 11:00-13:00\\
        Simulation time resolution & 5s\\
        Time interval for condition estimation & 15min \\
        Number of time intervals & 8 \\
        Number of links & 3548 \\
        Number of nodes & 1515 \\
        Number of origins/destinations & 126\\
        Number of OD pairs & 15876\\
        \bottomrule
    \end{tabular}
    \caption{Pittsburgh network setting}
    \label{tab:pgh}
\end{table}
\newpage
\begin{figure}[H]
    \centering
    \includegraphics[width=0.7\linewidth]{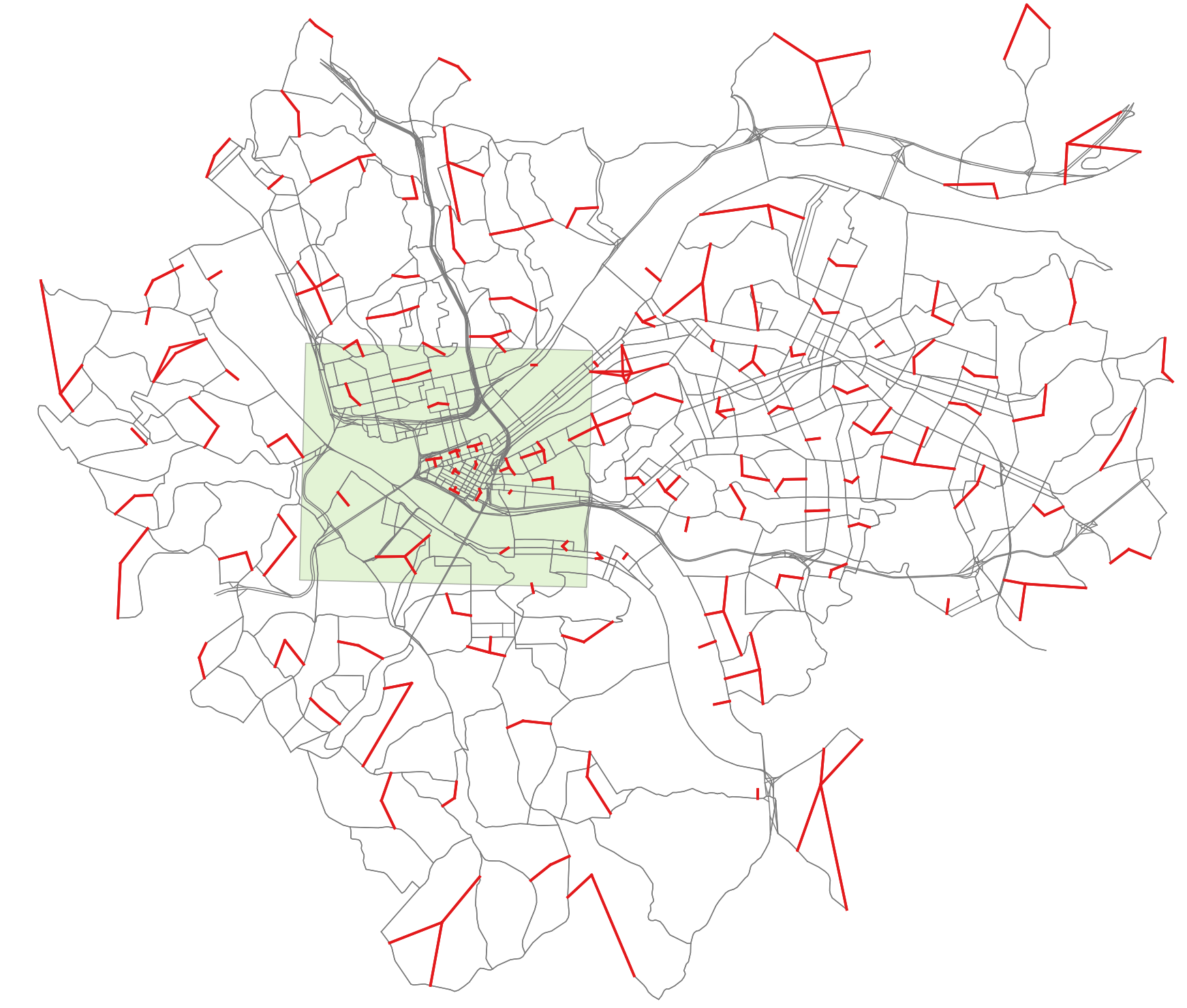}
    \caption{Coverage of two satellite images from Airbus}
    \label{fig:bbox}
\end{figure}

\newpage
\end{document}